\documentclass{article}

\usepackage[utf8]{inputenc}
\usepackage[T1]{fontenc}
\usepackage{amsmath}
\usepackage{amsfonts}
\usepackage{amssymb}
\usepackage{amsthm}
\usepackage{graphicx}
\usepackage{subcaption}
\usepackage[pdftex]{hyperref}
\usepackage{fullpage}
\usepackage{color}
\usepackage{comment}
\usepackage[sort&compress,numbers]{natbib}
\usepackage{todonotes}
\usepackage{booktabs}
\usepackage{algorithm}
\usepackage{algpseudocode}
\usepackage{enumitem}
\usepackage{tabularx}     
\usepackage{multirow}     
\usepackage{caption}      
\usepackage{etoolbox}     
\usepackage{tikz}
\usepackage{pgfplots}
\usetikzlibrary{positioning,calc}
\pgfplotsset{compat=1.18}

\theoremstyle{definition}
\newtheorem{definition}{Definition}[section]

\newtheorem{lemma}{Lemma}[section]

\title{Learnability Window in Gated Recurrent Neural Networks}

\author{
Lorenzo Livi\thanks{
OPIT -- Open Institute of Technology.
\href{mailto:lorenz.livi@gmail.com}{lorenz.livi@gmail.com}.
ORCID: \href{https://orcid.org/0000-0001-6384-4743}{0000-0001-6384-4743}.
\href{https://scholar.google.com/citations?user=hAL9amAAAAAJ&hl=en}{Google Scholar profile}.
}}
\date{\today}

\begin{document}
\maketitle

\begin{abstract}
We develop a statistical theory of temporal learnability in recurrent neural networks, quantifying the maximal temporal horizon $\mathcal{H}_N$ over which gradient-based learning can recover lag-dependent structure at finite sample size $N$. The theory is built on the effective learning rate envelope $f(\ell)$, a function that captures how gating mechanisms and adaptive optimizers jointly shape the coupling between state-space dynamics and parameter updates during Backpropagation Through Time.
Under heavy-tailed ($\alpha$-stable) fluctuations, where empirical averages concentrate at rate $N^{-1/\kappa_\alpha}$ with $\kappa_\alpha = \alpha/(\alpha-1)$, the interplay between envelope decay and statistical concentration yields explicit scaling laws for the growth of $\mathcal{H}_N$: logarithmic, polynomial, and exponential temporal learning regimes emerge according to the decay law of $f(\ell)$.
These results identify envelope decay as the key determinant of temporal learnability. Slower attenuation of $f(\ell)$ enlarges $\mathcal{H}_N$, while heavy-tailed fluctuations compress it by weakening statistical concentration.
Moreover, envelope geometry outweighs dataset size: slowing the envelope's decay enlarges $\mathcal{H}_N$ more than adding data, so more complex architectures that realize slower-decaying envelopes can be more data-efficient than simpler ones.
Experiments across multiple gated architectures and optimizers corroborate these structural predictions.
\end{abstract}

\section{Introduction}

Recurrent neural networks (RNNs) are fundamental models for processing
sequential data, yet their ability to learn long-range temporal dependencies
remains only partially understood.
Gated architectures such as the LSTM and GRU have dramatically improved
numerical stability and empirical performance, but it is still unclear which
temporal dependencies are statistically recoverable under finite data.
Existing analyses have studied dynamical stability, spectral properties of
Jacobian products, mean-field approximations, and forward signal propagation
through deep networks.
These contributions characterize important necessary conditions, e.g. when
gradients do not explode or vanish, or when input information survives
layer-to-layer, but they typically treat state dynamics and parameter
dynamics in isolation.
None provides a unified statistical criterion for when transported gradient
signals remain distinguishable from noise during training, which requires
analyzing the coupled evolution of states and optimizer-driven
parameter updates.

Our previous work showed that gating induces heterogeneous time scales that
shape both state evolution and gradient transport~\cite{livi2025timescale}.
Here we develop a statistical theory of finite-horizon learnability that analyzes this coupled system.
The central objects are the effective learning rates $\mu_{t,\ell}$, which quantify how Backpropagation Through Time (BPTT) reweights gradient signals across temporal lags for each neuron.
These quantities arise from the interaction between gate-induced transport of gradients through recurrent dynamics and the base learning rates imposed by the optimizer.
Their aggregate magnitude defines the envelope $f(\ell)=\|\mu_{t,\ell}\|_1$, which governs the effective strength of lagged gradient contributions to parameter updates during training. Although effective learning rates were originally~\cite{livi2025timescale} derived under SGD with a fixed global learning rate~$\mu$, we generalize them here to adaptive optimizers such as Adam, where each parameter receives its own time-varying rate.
The generalization replaces the global~$\mu$ with a neuron-specific adaptive base rate $\Lambda^{(q)}_{r,\ell}$, obtained by projecting the parameter-space preconditioner onto each neuron via a Rayleigh-quotient construction, while leaving the gate-induced transport factors unchanged.

The central question addressed in this paper is: given a finite number of training sequences, up to which temporal horizon can dependencies be statistically detected?
To answer this, we construct a matched statistic that aggregates gradient-derived contributions across neurons and time steps, and introduce the learnability window $\mathcal{H}_N$, defined as the largest lag for which this statistic remains distinguishable from its noise floor at sample size $N$.
Under heavy-tailed ($\alpha$-stable) fluctuations of the matched statistic, empirical averages concentrate at rate $N^{-1/\kappa_\alpha}$ with $\kappa_\alpha=\alpha/(\alpha-1)$.
We show that the interaction between this concentration rate and the decay law of the envelope $f(\ell)$ determines the scaling behavior of $\mathcal{H}_N$.
In particular, exponential, power-law, and logarithmic attenuation of $f(\ell)$ induce qualitatively distinct growth laws for the learnability window, providing a universal classification of temporal learning regimes.\footnote{%
We adopt power-law as the canonical term for envelope decay of the
form $f(\ell)\asymp c\,\ell^{-\beta}$. The same relation is commonly called
polynomial when referring to the asymptotic scaling regime or the
growth of derived quantities such as $N(\ell)$ or $\mathcal{H}_N$, and is
also known as algebraic decay in parts of the literature; we treat
these as synonyms for power-law throughout the paper.}

The two factors entering this classification (envelope decay and noise concentration) are mathematically separable but mechanistically coupled.
Both are products of the same training trajectory: gating dynamics and optimizer adaptation jointly shape $f(\ell)$, while the same trajectory generates the matched-statistic fluctuations whose tail index $\alpha$ governs concentration through $\kappa_\alpha$.
Consistent with this shared origin, we observe empirically that regimes with slow envelope decay tend to co-occur with persistent heavy-tailed fluctuations.

Envelope decay and heavy-tailed concentration are thus the two ingredients of temporal learnability, and the contributions below formalize each side.

\paragraph{Contributions.}
\begin{itemize}[leftmargin=*,itemsep=2pt]

\item \emph{A statistical-detectability framing of temporal learnability.}
We cast temporal learnability as a finite-sample statistical question rather than as a property of local Jacobian stability.
The learnability window $\mathcal{H}_N$ is defined as the largest lag at which a matched gradient statistic remains distinguishable from its noise floor at sample size $N$.
This places long-range learning in a finite-data regime where detectability is the operative constraint.
The growth of $\mathcal{H}_N$ with $N$ falls into three canonical regimes: logarithmic, polynomial, and exponential.
The realized regime is set by the asymptotic decay class of the envelope $f(\ell)$.

\item \emph{A master sample-complexity relation, in which envelope geometry outweighs dataset size.}
We establish the fundamental scaling
$N(\ell)\asymp f(\ell)^{-\kappa_\alpha}$, with concentration exponent $\kappa_\alpha$ amplifying envelope decay into sample-complexity growth.
Because heavy-tailed averages concentrate only as $N^{1/\kappa_\alpha}$, adding data extends the learnability window slowly, whereas slowing the envelope's decay changes its growth class outright.
A direct consequence is that a more complex architecture that realizes a slower-decaying envelope can be more data-efficient than a simpler one.
This relation also underlies the three-regime classification and quantifies the data cost of detecting lag-dependent structure at each horizon.

\item \emph{Heavy-tailed fluctuations universally compress temporal learnability.}
The exponent $\kappa_\alpha$ enters every regime multiplicatively, so
smaller $\alpha$ tightens $\mathcal{H}_N$ across every envelope decay
class.
Envelope geometry and fluctuation statistics thus act as joint,
antagonistic determinants of the learnability window. Heterogeneous
time-scale structure slows envelope decay and extends $\mathcal{H}_N$,
whereas heavy-tailed fluctuations counteract this effect by weakening
statistical concentration.

\end{itemize}
The theoretical predictions are tested empirically across several
gated recurrent architectures, including LSTM, GRU, and gated baseline variants, and across both adaptive and simple SGD-like optimizers.
This validation probes whether the envelope decay class and the
heavy-tailed fluctuation regime predicted by the theory are actually
realized by the trained dynamics, and whether the resulting
learnability window aligns with the canonical scaling regimes.

Figure~\ref{fig:scaling_laws} provides a schematic representation of the scaling laws derived in this paper.
\begin{figure}[t]
\centering
\begin{tikzpicture}
\begin{semilogyaxis}[
    name=panA,
    width=0.40\textwidth, height=4.8cm,
    xlabel={Lag $\ell$}, ylabel={$f(\ell)$},
    xmin=1, xmax=200, ymin=2e-3, ymax=2,
    grid=both, grid style={gray!15},
    legend cell align={left}, legend pos=south west,
    legend style={font=\scriptsize, draw=none, fill=none, row sep=-1pt},
    title={\textbf{(a)} Envelope $f(\ell)$},
    title style={font=\small},
    label style={font=\small}, tick label style={font=\footnotesize},
    samples=200, domain=1:200,
]
\addplot[thick, red!75!black, solid]   {exp(-x/30)};         \addlegendentry{$e^{-\ell/\tau}$}
\addplot[thick, blue!65!black, dashed] {(1+x)^(-0.8)};       \addlegendentry{$\ell^{-\beta}$}
\addplot[thick, green!50!black, densely dotted, line width=1.3pt]
                                        {1/ln(2.71828+x)};   \addlegendentry{$(\log\ell)^{-\vartheta}$}
\end{semilogyaxis}

\begin{semilogyaxis}[
    at={(panA.east)}, anchor=west, xshift=1.4cm,
    name=panB,
    width=0.40\textwidth, height=4.8cm,
    xlabel={Lag $\ell$}, ylabel={$N(\ell)$},
    xmin=1, xmax=200, ymin=1, ymax=1e9,
    grid=both, grid style={gray!15},
    title={\textbf{(b)} Sample complexity $N(\ell)$},
    title style={font=\small},
    label style={font=\small}, tick label style={font=\footnotesize},
    samples=200, domain=1:200,
]
\addplot[thick, red!75!black, solid]   {exp(3*x/30)};
\addplot[thick, blue!65!black, dashed] {(1+x)^(2.4)};
\addplot[thick, green!50!black, densely dotted, line width=1.3pt]
                                        {(ln(2.71828+x))^3};
\addplot[dashed, gray!70, thick, domain=1:200] {1e6};
\node[anchor=west, font=\scriptsize, gray!70!black]
  at (axis cs:3,1.8e6) {$N_{\rm budget}$};
\end{semilogyaxis}


\begin{loglogaxis}[
    at={($(panA.south west)!0.5!(panB.south east)+(0,-2.2cm)$)},
    anchor=north,
    name=panC,
    width=0.55\textwidth, height=5cm,
    xlabel={Sample budget $N$}, ylabel={$\mathcal{H}_N$},
    xmin=1, xmax=1e6, ymin=1, ymax=1e4,
    grid=both, grid style={gray!15},
    legend cell align={left}, legend pos=north west,
    legend style={font=\scriptsize, draw=none, fill=none, row sep=-1pt},
    title={\textbf{(c)} Learnability window $\mathcal{H}_N$},
    title style={font=\small},
    label style={font=\small}, tick label style={font=\footnotesize},
    samples=200, domain=1:1e6,
]
\addplot[thick, red!75!black, solid]   {1+ln(x)};
\addlegendentry{$\mathcal{H}_N\!\sim\!\log N$}
\addplot[thick, blue!65!black, dashed] {x^(1/2.4)};
\addlegendentry{$\mathcal{H}_N\!\sim\!N^{1/(\beta\kappa_\alpha)}$}
\addplot[thick, green!50!black, densely dotted, line width=1.3pt]
                                        {exp(x^(1/3))};
\addlegendentry{$\mathcal{H}_N\!\sim\!\exp(N^{1/(\vartheta\kappa_\alpha)})$}
\end{loglogaxis}
\end{tikzpicture}

\caption{Schematic illustration of the three envelope-decay
regimes and their consequences for finite-sample learnability.
(a)~Envelope $f(\ell)$ for the three canonical decay regimes.
(b)~Heavy-tailed $\alpha$-stable concentration maps the envelope into
a sample-complexity requirement $N(\ell)\asymp f(\ell)^{-\kappa_\alpha}$.
(c)~Inverting $N(\ell)$ at a finite training budget yields the
learnability window $\mathcal{H}_N$ as a function of $N$.}
\label{fig:scaling_laws}
\end{figure}

\paragraph{Paper organization.}
The remainder of the paper is organized as follows.
Section~\ref{sec:related_work} reviews related literature.
Section~\ref{sec:bptt} recalls the BPTT framework under both SGD and adaptive optimizers.
Section~\ref{sec:generalized_elr} introduces the generalized effective learning rate formulation used throughout the paper.
Architecture-specific Jacobian derivations and transport factors for all recurrent models are provided in Appendix~\ref{app:RNNs_jacobians_and_elr}.
Section~\ref{sec:learnability_theory} develops the theoretical framework of the learnability window.
Section~\ref{sec:experiments} presents empirical validation.
Section~\ref{sec:discussion} concludes with implications and future directions.
Additional technical derivations and supporting analyses are provided in the
appendices (Appendices~\ref{app:first_order_expansion}--\ref{app:code}).

\section{Related work}
\label{sec:related_work}

The introduction of gating mechanisms in LSTM~\cite{hochreiter1997long} and GRU~\cite{cho2014encoderdecoder} was pivotal for controlling temporal credit assignment in recurrent neural networks.  
Foundational and later theoretical works linked the forget gate to interpretable exponential decay and controllable memory retention~\cite{gers2000learning,tallec2018can}, while more recent studies place recurrent dynamics in continuous-time or ODE-based perspectives~\cite{chang2019antisymmetric}.  
These contributions illuminate how gates shape state dynamics, but do not analyze how gating multiplicative structures affect parameter updates and the capacity for learning long-range dependencies.
A complementary line of work models gates as learned rescaling or time-warping mechanisms.  
Tallec and Ollivier~\cite{tallec2018can} formalized gating as adaptive time dilation, and recent geometry-based analyses~\cite{krishnamurthy2022theory} study how gating choices influence information flow.  
We build on these ideas by explicitly casting gate-derived Jacobian terms into effective learning rates and embedding them in Bayes-optimal detectability bounds.

The classical vanishing--exploding gradients problem motivated many stabilization techniques: clipping, spectral regularization, and orthogonality or unitary RNNs~\cite{pascanu2013difficulty,arjovsky2016unitary,wisdom2016full}.  
Hierarchical, dilated, or skip architectures shorten propagation paths~\cite{koutnik2014clockwork,chang2017dilated,chung2017hierarchical}, while continuous-time and implicit recurrent approaches offer alternative stability--expressivity tradeoffs~\cite{chang2019antisymmetric}.
Works on dynamical isometry control the conditioning of gradient transport~\cite{chen2018dynamical,pennington2017resurrecting} but do not address whether transported gradients still carry statistically reliable information over long lags, nor do they link the resulting propagation properties to sample-complexity requirements or a data-dependent learnability horizon.  
Recent work further argues that vanishing and exploding gradients are not the whole story: as memory increases, parameter perturbations can induce increasingly large output variations, making gradient-based optimization highly sensitive even without gradient explosion, while element-wise recurrences and careful parametrizations help mitigate this effect~\cite{zucchet2024vanishing}.  
In contrast, we analyze statistical detectability, showing that even when products of Jacobians remain numerically stable, learning can be ineffective if the envelope decays too rapidly.
Related difficulties have long been observed in tasks with long-delayed supervision~\cite{bengio1994learning}, where the learning signal must be propagated across long temporal gaps; however, existing analyses typically focus on optimization stability rather than statistical detectability across lags of the coupled dynamics.

A related but distinct line of work concerns memory capacity in reservoir computing and echo state networks, where the recurrent dynamics are kept fixed and only a readout layer is trained \cite{dambre2012information,verzelli2019echo}. In this setting, classical results characterize the ability of a randomly driven recurrent system to retain information about past inputs over time, typically quantified via linear or nonlinear memory capacity measures. Such analyses focus on expressivity and information retention properties of untrained dynamical systems, rather than on the optimization dynamics of recurrent weights under gradient-based learning. By contrast, the present work addresses a complementary question: how training dynamics themselves impose horizon-dependent limits on learnability through the interplay of effective learning rates and noise. In particular, our framework does not rely on fixed reservoirs, but instead characterizes how learnability degrades under heavy-tailed fluctuations observed during SGD-like optimization.
In fact, \c{S}im\c{s}ekli et al.~\cite{simsekli2019tail} show that mini-batch SGD noise often exhibits $\alpha$-stable, heavy-tailed behavior rather than Gaussian tails. 
Subsequent research has further examined the algorithmic and theoretical consequences of such heavy-tailed gradient statistics. 
H\"ubler et al.~\cite{hubler2025gradientclippingnormalizationheavy} and Sun et al.~\cite{sun2025heavytailed} connect gradient clipping to normalized SGD under heavy-tailed noise and derive improved convergence guarantees.
More broadly, heavy-tailed fluctuations have been identified as general dynamical signatures of SGD-trained deep networks, beyond the gradient-noise statistics themselves. Ly and Gong~\cite{ly2025optimization} show that a multifractal loss-landscape model reproduces anomalous diffusion and an extended edge of chaos and steers optimizers toward flatter minima, while Zhang and Tang~\cite{zhang2025heavytailed} show that heavy-tailed parameter updates self-organize from an exploration--relevance information balance into a stable power-law regime. Together, they indicate that scale-free fluctuations are an intrinsic, and plausibly functional, feature of SGD-like learning rather than an artifact of mini-batch sampling.

In the broader learning systems literature, trainability has been studied via mean-field theory, spectral initialization, curvature, and the geometry of parameter landscapes~\cite{schoenholz2017deep,saxe2013exact,pennington2017resurrecting,martens2020new,li2018visualizing,shwartz2017opening,zhang2022all,Bonnaire_2024,kingma2014adam}.
In recurrent and transformer settings, recent work further links trainability to
Jacobian spectra, Fisher information, and anisotropic learning
rates~\cite{barrett2021implicit,smith2018bayesian,livi2018fisher}.
A complementary line of work studies stochastic optimization itself as a dynamical system. Mandt et al.~\cite{mandt2017sgd} interpret constant-step SGD as approximate Bayesian inference with an Ornstein--Uhlenbeck-type stationary regime, while Yaida~\cite{yaida2019fluctuation} establishes fluctuation--dissipation relations governing stationary SGD statistics.
Zhou et al.~\cite{Zhou2020Towards} analyze the SGD--Adam generalization gap through a L\'evy-driven SDE perspective, arguing that Adam's coordinate-wise adaptation and exponential averaging can slow escape from sharp basins.
More recent work further analyzes the stationary distributions induced by stochastic optimization and the balance between noise and drift in high-dimensional systems~\cite{zyin2025noise}.
However, these approaches typically analyze state dynamics and parameter dynamics in isolation. They do not treat learning as a coupled dynamical system in which temporal transport through recurrent Jacobians interacts with stochastic optimization noise, nor do they characterize the resulting universal scaling regimes governing temporal learnability and the sample complexity of detecting long-range dependencies.

Finally, structured state-space models construct linear recurrences with learnable spectra and have shown strong long-range performance~\cite{gu2022efficiently,gu2021combining,gu2022parameterization,orvieto2023resurrecting,amoalonso-ssm2025}. We retain the focus on classical gated recurrent systems trained by SGD-like optimization; extending the envelope-based learnability analysis to structured state-space models is a natural future direction.

\section{Backpropagation through time}
\label{sec:bptt}

Training RNNs follows the same fundamental principle as feedforward models: parameter updates are obtained through SGD or variations thereof \cite{ruder2016overview}.
Given trainable parameters~$\theta$ and learning rate~$\mu$, the SGD update at iteration~$r$ is
\begin{equation}
\label{eq:sgd_update}
\theta_{r+1}
\;=\;
\theta_r
-\mu\,\nabla_\theta \mathcal{L}(\theta_r),
\qquad
\mathcal{L}
\;=\;
\sum_{t=1}^{T}\mathcal{E}_t,
\end{equation}
where $\mathcal{E}_t$ denotes the instantaneous loss at time~$t$ within a sequence of length~$T$.
In practice, gradients are averaged over a mini-batch of independent sequences; since this averaging does not affect the per-sequence BPTT structure analyzed below, we present the derivation for a single sequence without loss of generality.

In an adaptive optimizer, the update can be written as a diagonally preconditioned gradient descent step~\cite{kingma2014adam,ruder2016overview}
\begin{equation}
\label{eq:sgd_update_adaptive}
\theta_{r+1}
=
\theta_r
-
\Lambda_r\nabla_\theta \mathcal{L}(\theta_r),
\qquad
\Lambda_r = \mathrm{diag}(\lambda_{1,r},\dots, \lambda_{P,r})
\in\mathbb{R}^{P \times P},
\end{equation}
where $P = \dim\theta$ and $\lambda_{i,r} > 0$ are per-parameter adaptive learning rates determined by the optimizer state. For example, in Adam-type methods, these rates depend on running estimates of gradient moments.

In recurrent architectures, computing the gradient~$\nabla_\theta \mathcal{L}$ requires unrolling the network dynamics through time and accounting for how earlier states influence later losses---a process known as Backpropagation Through Time.

Let $h_t$ denote the recurrent state at time $t$.
The total gradient of the loss with respect to the parameters can be written as
\begin{equation}
\label{eq:bptt_grad_sum}
\nabla_\theta \mathcal{L}
\;=\;
\sum_{t=1}^{T}
\frac{\partial \mathcal{E}_t}{\partial \theta}
\;=\;
\sum_{t=1}^{T}
\frac{\partial \mathcal{E}_t}{\partial h_t}
\sum_{\ell=1}^{t}
\frac{\partial h_t}{\partial h_\ell}
\frac{\partial h_\ell}{\partial \theta}.
\end{equation}
The outer sum aggregates contributions from all time steps, while the inner chain rule expresses how parameter perturbations at earlier times influence the current loss through the recurrent dynamics.
To make this dependence explicit, define $J_j \;=\; \frac{\partial h_j}{\partial h_{j-1}}$ and $B_\ell(\theta) \;=\; \frac{\partial h_\ell}{\partial \theta}$.
Here $J_j$ is the state Jacobian, which quantifies how the state evolves in response to infinitesimal perturbations of the previous state, and $B_\ell(\theta)$ is the parameter-state Jacobian, measuring the instantaneous sensitivity of the state to the parameters at step~$\ell$.
Substituting these definitions into Eq.~\eqref{eq:bptt_grad_sum}, the gradient contribution of a specific loss term $\mathcal{E}_t$ becomes
\begin{equation}
\label{eq:bptt_one_step}
\frac{\partial \mathcal{E}_t}{\partial \theta}
\;=\;
\delta_t^\top
\sum_{\ell=1}^{t}
\mathcal{M}_{t,\ell}\,B_\ell(\theta),
\qquad
\mathcal{M}_{t,\ell}
\;=\;
\prod_{j=\ell+1}^{t} J_j,
\end{equation}
where $\delta_t = \partial \mathcal{E}_t / \partial h_t$ denotes the local loss gradient at time~$t$.  
The matrix product $\mathcal{M}_{t,\ell}$, often referred to as the state-transition Jacobian product, transports this signal backward through the sequence, modulating both its magnitude and direction as it interacts with the intermediate state Jacobians~$J_j$.

In Eqs.~\eqref{eq:bptt_grad_sum}--\eqref{eq:bptt_one_step}, $\ell$ is an absolute source-time index. From the following section onward, $\ell$ denotes a temporal lag relative to the reference time~$t$.
To avoid introducing duplicate notation, we retain the symbols $\mathcal{M}_{t,\ell}$ and $B_\ell$ under this re-indexing: when $\ell$ denotes a lag, $\mathcal{M}_{t,\ell}$ refers to the source-time transport $\mathcal{M}_{t,t-\ell}$ in Eq.~\eqref{eq:bptt_one_step}, while $B_\ell(\theta)$ refers to $B_{t-\ell}(\theta)=\partial h_{t-\ell}/\partial\theta$.
Thus, in the lag convention, $\mathcal{M}_{t,\ell}$ transports over the $\ell$ steps from time $t-\ell$ to time~$t$. This is only a re-indexing of the same BPTT quantities.

\section{Generalized effective learning rates under adaptive optimization}
\label{sec:generalized_elr}

This section extends the effective learning rate framework to adaptive optimizers.
Appendix~\ref{app:RNNs_jacobians_and_elr} derives the neuronwise effective learning
rates obtained from the first-order expansion of recurrent Jacobian products
for all architectures considered in this paper.
These derivations show that the lag-dependent learning rates induced by BPTT can be expressed as the product of two
components: a scalar optimizer step size and a transport factor determined
by the recurrent dynamics.

To streamline notation, we collect the diagonal contributions of the
Jacobian-product expansion into the transport factor
\begin{equation}
\label{eq:transport_factor}
\Gamma^{(q)}_{t,\ell}
=
\gamma^{(0,q)}_{t,\ell} + \gamma^{(1,q)}_{t,\ell},
\end{equation}
corresponding to the sum of the zeroth- and first-order transport contributions.

Under plain SGD, the optimizer applies a uniform base learning rate $\mu$ to all parameters.  
The neuronwise effective learning rates therefore factorize as
\begin{equation}
\label{eq:effective_learning_rate_SGD}
\mu^{(q)}_{t,\ell}
=
\mu\,\Gamma^{(q)}_{t,\ell},
\end{equation}
Hence, $\Gamma^{(q)}_{t,\ell}$ entirely determines the decay geometry of the effective learning rates across temporal lags up to a scaling factor $\mu$.

Modern recurrent networks, however, are typically trained using
adaptive optimizers such as Adam or RMSprop, which apply a diagonal
preconditioner to the gradient.  
In this setting, the global base rate $\mu$ is replaced by a
neuron-specific adaptive base rate $\Lambda^{(q)}_{r,\ell}$ derived
from the optimizer preconditioner.  
The resulting \emph{generalized} effective learning rates take the form
\begin{equation}
\label{eq:GELR}
\mu^{(q)}_{t,\ell}
=
\Lambda^{(q)}_{r,\ell}\Gamma^{(q)}_{t,\ell}.
\end{equation}

The adaptive base rate $\Lambda^{(q)}_{r,\ell}$ is obtained by
projecting the diagonal optimizer preconditioner onto the
parameter-space direction associated with neuron $q$ at lag $\ell$.
This projection is given by the Rayleigh quotient
\begin{equation}
\label{eq:rayleigh}
\Lambda^{(q)}_{r,\ell}
=
\frac{
B_\ell^{(q)}\,\Lambda_r\,(B_\ell^{(q)})^\top
}{
B_\ell^{(q)}(B_\ell^{(q)})^\top
} > 0,
\end{equation}
where $B_\ell^{(q)}$ denotes the parameter-state sensitivity of neuron
$q$ and $\Lambda_r$ is the diagonal optimizer preconditioner.
A justification and the full derivation of this projection are
provided in Appendix~\ref{app:adaptive_base_rate}, where we show that
the Rayleigh quotient is the unique scalar that best approximates the
preconditioned neuron-sensitivity direction in the least-squares
sense, is invariant under arbitrary normalization of the sensitivity
vector, and reduces to the global learning rate $\mu$ under plain SGD.

\section{Learnability under heavy-tailed fluctuations}
\label{sec:learnability_theory}

In this section, we quantify finite-horizon learnability.
Starting from the per-lag, per-neuron generalized effective learning rates $\mu_{t,\ell}^{(q)}$, we construct a matched statistic that casts lag-$\ell$ credit assignment as a binary detection problem.
Under a symmetric $\alpha$-stable ($\mathcal{S}\alpha\mathcal{S}$) \cite{nolan2020stable} location model for the fluctuations of this statistic, standardization reduces the problem to testing two shifted stable laws whose separation is the dimensionless quantity $d_N(\ell)$.
We characterize detectability through the Bayes error of this two-point experiment, obtain a critical separation threshold, and translate it into sample-complexity requirements and a learnability window $\mathcal{H}_N$.
Finally, we derive scaling laws for $\mathcal{H}_N$ under logarithmic, power-law, and exponential decay of the envelope $f(\ell)$, highlighting the role of the tail index $\alpha$.\footnote{Throughout the learnability analysis, $\ell\ge1$ denotes a
temporal lag under the re-indexing introduced at the end of
Section~\ref{sec:bptt}. We suppress the explicit dependence on the
reference time~$t$ in quantities such as $f(\ell)$ and $m_q(\ell)$.
In the theoretical development, $t$ is fixed as the endpoint of the
transport. In the empirical evaluation, we average over all valid
reference times and diagnostic sequences, leaving only the dependence
on the lag~$\ell$.}

\subsection{The envelope}
\label{sec:envelope_GELR}

The precise form of the effective learning rates depends on the optimizer (Section~\ref{sec:generalized_elr}). However, the learnability analysis depends on the effective learning rates $\mu^{(q)}_{t,\ell}$ only through their aggregate magnitude across neurons. We therefore introduce the envelope
\begin{equation}
\label{eq:envelope_def}
f(\ell)
\;=\;
\|\mu_{t,\ell}\|_1
\;=\;
\sum_{q=1}^{H} |\mu^{(q)}_{t,\ell}|.
\end{equation}
This function describes how the state--parameter coupling strength varies with lag $\ell$. More precisely, it summarizes the total strength with which lag-$\ell$ gradient contributions are amplified or attenuated by the combined action of recurrent transport and optimizer scaling.

The envelope admits the following decomposition
\begin{equation}
\label{eq:envelope_decomp}
f(\ell)
=
\underbrace{
\mu\sum_{q=1}^{H}|\Gamma^{(q)}_{t,\ell}|
}_{f_{\mathrm{gates}}(\ell)}
+
\underbrace{
\sum_{q=1}^{H}
\Delta\Lambda^{(q)}_{r,\ell}\,|\Gamma^{(q)}_{t,\ell}|
}_{f_{\mathrm{adapt}}(\ell)},
\end{equation}
where
\[
\Delta\Lambda^{(q)}_{r,\ell}
=
\Lambda^{(q)}_{r,\ell}-\mu .
\]

The first term represents the contribution to the envelope arising purely from gate-controlled gradient transport, while the second isolates the contribution of the adaptive optimizer.
Because $\Lambda^{(q)}_{r,\ell}>0$, the decomposition is purely algebraic and introduces no approximation. Under SGD we have
$\Delta\Lambda^{(q)}_{r,\ell}=0$, so that $f(\ell)=f_{\mathrm{gates}}(\ell)$.
Under adaptive optimization, the neuron-specific base rates $\Lambda^{(q)}_{r,\ell}$ reweight the transport terms $|\Gamma^{(q)}_{t,\ell}|$, which can modify the decay geometry of the envelope.

The envelope relies on the transport factors $\Gamma^{(q)}_{t,\ell}$ obtained from the first-order diagonal expansion of the Jacobian product $\mathcal{M}_{t,\ell}$ (Appendix~\ref{app:first_order_expansion}).
These factors admit a transparent structure: at zeroth order they reduce to products of gating values, which are the time scales operating in state space; the first-order terms add gate derivatives mixed with weight matrices (see Appendix~\ref{app:RNNs_jacobians_and_elr} for architecture-specific derivations).
The numerical analysis in~\cite{livi2025timescale} shows that, in many operationally significant cases, truncating the matrix product to first order preserves numerical accuracy.

Alternatively, one could define the envelope directly from the full transport matrix as
\[
f_{\mathcal{M}}(\ell)=\sum_{q=1}^{H}|\Lambda^{(q)}[\mathcal{M}_{t,\ell}]_{qq}|,
\]
bypassing the first-order expansion.  This would yield $H$ numbers with no structural decomposition into gating and mixing contributions, making it impossible to assign a time scale to individual neurons.
The first-order expansion provides this interpretability without losing numerical quality: as verified in Appendix~\ref{app:envelope_validation}, the envelope $f(\ell)$ in Eq.~\ref{eq:envelope_def} preserves the decay profile of $f_{\mathcal{M}}(\ell)$ (Spearman $\rho\geq0.972$, Pearson $r\geq0.823$, mean~$0.95$).

\subsection{Gradient contribution at lag $\ell$}
\label{sec:gradient_contribution_at_lagl}

We begin by isolating the portion of the BPTT gradient that reflects the
influence of past states.

Let $h_t$ denote the recurrent hidden state used by BPTT.
Recalling the decomposition in Eq.~\eqref{eq:bptt_one_step}, the lag-$\ell$ contribution to the parameter gradient can be written as
\begin{equation}
\label{eq:lagl_gradient_contribution}
g_{t,\ell}(\theta)
=
\delta_t^\top \mathcal{M}_{t,\ell}\,B_{\ell}(\theta)
\;\in\;
\mathbb{R}^{1\times P}.
\end{equation}
Then, decompose the parameter Jacobian into neuronwise rows,
\[
B_{\ell}(\theta)
=
\begin{bmatrix}
B^{(1)}_{\ell}\\
\vdots\\
B^{(H)}_{\ell}
\end{bmatrix},
\qquad
B^{(q)}_{\ell}\in\mathbb{R}^{1\times P},
\]

Following the first-order expansion of Jacobian products, we approximate $\mathcal M_{t,\ell}$ to first order and retain only the generalized effective learning rates $\mu^{(q)}_{t,\ell}$ \eqref{eq:GELR}. As demonstrated in Appendix~\ref{app:envelope_validation}, this operation preserves the decay profile of the envelope with high accuracy.
Thus, using $\mu^{(q)}_{t,\ell}$ yields the following first-order approximation of Eq.~\eqref{eq:lagl_gradient_contribution}:
\begin{equation}
\label{eq:gradient_contribution_first_order_approx}
g_{t,\ell}(\theta)
\approx
\sum_{q=1}^{H}
\mu^{(q)}_{t,\ell}\,
\delta_t^{(q)}\, B^{(q)}_{\ell}(\theta).
\end{equation}
As will be shown in the following subsections, this approximation allows us to precisely determine the contribution of the envelope~\eqref{eq:envelope_def} in the detection analysis, which leads to the definition of the learnability window.

\subsection{Matched statistic}
\label{sec:binary_detection_matched_statistic}

We formalize finite-horizon learnability by asking when information about a state at lag $\ell$ remains statistically present in the stochastic gradient at time $t$.
The lag-$\ell$ contribution to the BPTT gradient is, up to first-order approximation, the vector $g_{t,\ell}(\theta)\in\mathbb{R}^{1\times P}$ defined in Eq.~\eqref{eq:gradient_contribution_first_order_approx}.

To assess whether such information is recoverable in finite samples, we cast lag-$\ell$ learnability as a binary detection problem.
From the noisy gradient contribution $g_{t,\ell}(\theta)$, can we statistically distinguish the presence of a nonzero lag-$\ell$ signal from its absence?
In principle, this detection problem could be formulated directly in the $P$-dimensional parameter space.
However, working with a unidimensional statistic substantially simplifies both the statistical analysis and the interpretation of finite-sample effects.

We therefore compress the vector-valued quantity $g_{t,\ell}(\theta)$ into a
scalar by applying a fixed linear readout.
Let $w\in\mathbb{R}^P$ be a random unit-norm vector, drawn independently of the
data and fixed throughout training and across all model instances.
This projection defines a one-dimensional observable that can be interpreted as
a random but unbiased probe of the gradient geometry.

Applying this readout to $g_{t,\ell}(\theta)$ and using the first-order
approximation Eq.~\eqref{eq:gradient_contribution_first_order_approx}, we obtain the scalar
neuronwise alignment variables
\begin{equation}
\label{eq:neuronwise_alignment}
\zeta^{(q)}_{t,\ell}
=\delta_t^{(q)}\,
\langle B^{(q)}_{\ell}(\theta),\,w\rangle ,
\end{equation}
so that the lag-$\ell$ contribution reduces to the weighted sum
$\sum_{q=1}^{H}\mu^{(q)}_{t,\ell}\,\zeta^{(q)}_{t,\ell}$.

Because the same realization of $w$ is used for all models and training runs, differences in detectability across architectures reflect differences in the matched statistics, rather than artifacts of the projection itself.
Moreover, since $w$ is drawn uniformly from the unit sphere, its distribution is rotationally invariant.
As a result, the projection $\langle B^{(q)}_{\ell}(\theta),\,w\rangle$ provides an unbiased random probe of the typical magnitude and relative alignment of the neuronwise parameter-sensitivity directions, without privileging any fixed parameter direction.
Using the same realization of $w$ across models and training runs therefore
yields a consistent comparative diagnostic of norm and alignment structure,
without requiring that a single scalar projection exactly preserve the
signal-to-noise ratio of the full gradient.
Appendix~\ref{app:alpha_chain_validation} validates this choice
empirically and develops the multi-projection extension that
underlies the aggregation procedure used in the experiments.

The scalar variables $\zeta^{(q)}_{t,\ell}$ quantify how the transported gradient components associated with neuron~$q$ align, through the parameter Jacobian, with the fixed readout direction $w$.
The effective learning rates $\mu^{(q)}_{t,\ell}$ weight these alignment terms and thus govern how strongly each lag influences parameter updates.

We now define a matched statistic that allows us to characterize when past states remain recoverable during training.
Define the expected alignment $m_q(\ell)=\mathbb{E}[\zeta^{(q)}_{t,\ell}]$ and construct a matched statistic that aggregates lag-$\ell$ evidence across neurons:
\begin{equation}
\label{eq:matched_statistic}
S_{t,\ell}
=\sum_{q=1}^{H}
\mu^{(q)}_{t,\ell}\,
\mathrm{sgn}\!\big(m_q(\ell)\big)\,
\zeta^{(q)}_{t,\ell}.
\end{equation}
The factor $\operatorname{sgn}(m_q(\ell))$ reorients the $q$-th
coordinate according to the sign of its population mean alignment,
removing cancellations arising from the sign pattern of $m_q(\ell)$.\footnote{Indeed,
\[
\mathbb{E}\!\left[\operatorname{sgn}(m_q(\ell))\,
\zeta^{(q)}_{t,\ell}\right]
=
|m_q(\ell)|
\;\ge 0
\]
since $m_q(\ell)=\mathbb{E}[\zeta^{(q)}_{t,\ell}]$.
Note that the full summand in~\eqref{eq:matched_statistic}
also involves $\mu^{(q)}_{t,\ell}$, which can be signed;
the reorientation eliminates cancellations due to
$m_q(\ell)$ but not those due to $\mu^{(q)}_{t,\ell}$.}
This sign-oriented construction should be interpreted as the
\emph{theoretical} matched statistic. In the empirical pipeline
below, the population signs $\operatorname{sgn}(m_q(\ell))$ are
not available, so we instead work with the raw signed sum and
assess detectability through the magnitude of its empirical
mean shift.
A parallel theory could also be formulated for the raw signed statistic,
whose expectation would still factorize into the same envelope term
multiplied by a signed alignment coefficient. We introduce the
sign-oriented version as an oracle benchmark because the reorientation
removes cancellations due solely to heterogeneous sign patterns in the
unitwise expected alignments, thereby isolating a cleaner notion of
lag-$\ell$ signal strength.

Taking expectations over the randomness of the data yields
\begin{equation}
\label{eq:expectation_matched_statistic}
\mathbb{E}[S_{t,\ell}]
=\sum_{q=1}^{H}\mu^{(q)}_{t,\ell}\,|m_q(\ell)|
=\overline{m}_{\mu}(\ell)\,f(\ell),
\end{equation}
where
\begin{equation}
\label{eq:m_bar_mu_def}
\overline{m}_{\mu}(\ell)
=\frac{\sum_{q=1}^{H}\mu^{(q)}_{t,\ell}\,|m_q(\ell)|}
{\sum_{q=1}^{H}\big|\mu^{(q)}_{t,\ell}\big|}
\end{equation}
is an envelope-normalized alignment coefficient aggregating the neuronwise magnitudes $|m_q(\ell)|$ with signed weights $\mu^{(q)}_{t,\ell}$.

This factorization separates two complementary aspects:
(i)~$\overline{m}_{\mu}(\ell)$ captures the average informational alignment between
the gradient signal and the parameter-sensitivity directions at lag~$\ell$, and
(ii)~$f(\ell)$ quantifies the aggregate gain determining how strongly
lagged gradient information contributes to parameter updates after
both recurrent transport and optimizer scaling.
Together, they determine the expected strength of the lagged contribution to the
overall gradient, that is, how detectable past dependencies remain after
being propagated through the coupled state--parameter dynamics of training.

For finite-sample detection, the independent statistical units are full
training sequences rather than individual time positions within a sequence.
We therefore assign one normalized lag-specific observation to each sequence by
averaging $S_{t,\ell}$ over the admissible positions
$t=\ell+1,\dots,T$. This removes the trivial dependence on the number
$T-\ell$ of valid lag positions and yields a sequence-level statistic that is
comparable across lags. For the $n$th sequence, define
\begin{equation}
\label{eq:theoretical_sequence_average}
\bar{S}^{(n)}_{\ell}
=
\frac{1}{T-\ell}\sum_{t=\ell+1}^{T} S^{(n)}_{t,\ell}.
\end{equation}
Averaging these sequence-level statistics over $N$ independent training
sequences yields
\begin{equation}
\label{eq:theoretical_cross_sequence_average}
\widehat{S}_N(\ell)=\tfrac{1}{N}\sum_{n=1}^{N}\bar{S}^{(n)}_{\ell},
\end{equation}
the finite-sample matched statistic associated with the theoretical construction above.

\subsection{Finite-sample analysis}
\label{sec:sample_complexity}

Here, we derive the finite-sample requirements for detecting a
lag-$\ell$ dependency from noisy gradient information.
We write
\begin{equation}
\label{eq:kappa_alpha_def}
\kappa_\alpha=\frac{\alpha}{\alpha-1},
\qquad \alpha>1,
\end{equation}
for the concentration exponent associated with averages of
$\alpha$-stable fluctuations.

\subsubsection{Statistical model}
\label{sec:stat_model}

Empirical studies of SGD indicate that gradient fluctuations in deep networks
are well-described by a $\mathcal{S}\alpha\mathcal{S}$ distribution with
$1<\alpha\leq 2$, rather than just the Gaussian
case~\cite{simsekli2019tail,Savelii2024ClippingAdamHeavyTailed}.
$\mathcal{S}\alpha\mathcal{S}$ random variables form a family of heavy-tailed
laws indexed by the tail index $\alpha\!\in\!(0,2]$.
They are characterized by the characteristic function
$\phi_X(t)=\exp\!\big(-\sigma^\alpha |t|^\alpha\big)$, where $\sigma>0$ is a
scale parameter controlling dispersion.
Their probability densities generally lack closed form except in special cases
(Gaussian for $\alpha=2$ and Cauchy for $\alpha=1$),
and decay with power-law tails as $p(x)\sim |x|^{-(1+\alpha)}$.
For $\alpha<2$, these distributions have infinite variance; moreover, the
first moment is finite only when $\alpha>1$.
Throughout the detection framework developed below, we restrict $\alpha\in(1,2]$ to ensure that the mean exists and that empirical averages concentrate.

Motivated by this empirical evidence, and noting that the matched statistic aggregates gradient-derived contributions through linear operations, we model its fluctuations with a $\mathcal{S}\alpha\mathcal{S}$ location family (see Appendix~\ref{app:noise_floor} for a detailed justification). The $\mathcal{S}\alpha\mathcal{S}$ model is treated as an effective summary of the matched-statistic fluctuations, since heavy tails at this level may arise from interacting sources beyond gradient noise alone (e.g. heavy-tailed weight spectra~\citep{martin2021implicit} and parameter updates~\cite{zhang2025heavytailed}).

At lag~$\ell$, the finite-sample matched statistic $\widehat S_N(\ell)$ defined in Eq.~\eqref{eq:theoretical_cross_sequence_average} is modeled as
\begin{equation}
\label{eq:alpha_stable_model}
\widehat S_N(\ell)
\;\sim\;
\mathcal{S}\alpha\mathcal{S}\!\Big(
\theta_{\text{out}},\;
\sigma_\alpha(\ell)N^{-1/\kappa_\alpha}
\Big),
\qquad
\theta_{\text{out}}\!\in\!\left\{+\tfrac{1}{2}\Delta(\ell),\;-\tfrac{1}{2}\Delta(\ell)\right\},
\end{equation}
where $\theta_{\text{out}}$ is the location parameter, the mean separation $\Delta(\ell)=\overline{m}_{\mu}(\ell)f(\ell)$ quantifies the strength of the lag-$\ell$ signal, and $\sigma_\alpha(\ell)$ acts as a noise-scale proxy.

For analytical tractability, we assume bounded alignment magnitude and noise scale, $c_m \le |\overline{m}_{\mu}(\ell)| \le C_m$ and $c_{\sigma} \le \sigma_{\alpha}(\ell) \le C_{\sigma}$.
These conditions guarantee well-defined sample-complexity constants without affecting the asymptotic exponents.
The factor $N^{-1/\kappa_\alpha}$ encodes the slow concentration typical of $\alpha$-stable averages; Appendix~\ref{app:noise_floor}.

\subsubsection{Binary testing and Bayes-error threshold}
\label{sec:bayes_error_N_bound}

The detectability problem at lag~$\ell$ is equivalent to distinguishing
between the two location models in Eq.~\eqref{eq:alpha_stable_model}.
Let
\begin{equation}
\label{eq:stable_noise_scale}
s_N(\ell) = \sigma_\alpha(\ell)N^{-1/\kappa_\alpha}
\end{equation}
denote the noise scale of the averaged matched statistic.
Standardization by $s_N(\ell)$ is a one-to-one rescaling of the observation and therefore preserves all testing errors.
Under the two hypotheses, the standardized statistic is distributed as
\begin{equation}
\label{eq:standardized_binary_experiment}
  \frac{\widehat S_N(\ell)}{s_N(\ell)}
  \;\stackrel{d}{=}\;
  \pm
  \frac{\Delta(\ell)}{2s_N(\ell)}
  + Z_\alpha,
\end{equation}
where $Z_\alpha\sim\mathcal{S}\alpha\mathcal{S}(0,1)$ and $\stackrel{d}{=}$ means ``distributed as''.
Thus, detectability depends on the single dimensionless separation
\begin{equation}
\label{eq:normalized_separation}
  d_N(\ell)
  =
  \frac{|\Delta(\ell)|}{s_N(\ell)}
  =
  \frac{
    |\overline m_\mu(\ell)|\,f(\ell)\,N^{1/\kappa_\alpha}
  }{
    \sigma_\alpha(\ell)
  } .
\end{equation}
Equivalently, \(d_N(\ell)\) is the standardized signal separation:
it compares the lag-\(\ell\) mean shift \(|\Delta(\ell)|\) with the
finite-sample noise scale \(s_N(\ell)\). The Bayes-error argument below
supplies the critical threshold for this separation, and
Sec.~\ref{sec:experiments} implements the same dimensionless quantity
as an empirical signal-to-noise ratio (SNR) criterion.

We characterize detection by the equal-prior Bayes error of this two-point experiment under $0$--$1$ loss.
Under this loss, the Bayes classifier assigns an observation to the class with larger posterior probability, equivalently under equal priors to the model with larger likelihood~\citep{murphy2012,bishop2006}.

Let $p_\alpha$ denote the density of $Z_\alpha$. The two standardized hypotheses have densities $p_\alpha(x-d/2)$ and $p_\alpha(x+d/2)$, where $d\geq0$ is a generic separation.
An $\mathcal{S}\alpha\mathcal{S}$ density is also unimodal~\citep{nolan2020stable}.
Hence, the two shifted densities are equal at zero and ordered on either side of zero, so the Bayes rule is the sign test: decide for the positive-shift hypothesis when $x>0$ and for the negative-shift hypothesis when $x<0$, with ties at $x=0$ broken arbitrarily.
Under the positive-shift hypothesis, an error occurs when $d/2+Z_\alpha<0$, equivalently $Z_\alpha<-d/2$; under the negative-shift hypothesis, an error occurs when $-d/2+Z_\alpha>0$, equivalently $Z_\alpha>d/2$.

Therefore, the equal-prior Bayes error at separation $d$ is
\begin{equation}
\label{eq:stable_bayes_risk}
  R_\alpha(d)
  =
  \frac12\,\mathbb{P}\!\left(Z_\alpha < -\frac{d}{2}\right)
  +
  \frac12\,\mathbb{P}\!\left(Z_\alpha > \frac{d}{2}\right)
  =
  \mathbb{P}\!\left(Z_\alpha > \frac{d}{2}\right),
\end{equation}
where the final equality uses symmetry of $Z_\alpha$.
Since this is a survival probability, $R_\alpha(d)$ is nonincreasing in the standardized separation.
Thus, the theoretical detection error at lag~$\ell$ is exactly $R_\alpha(d_N(\ell))$.

For a target error level $\epsilon\in(0,1/2)$, define the critical standardized separation
\begin{equation}
\label{eq:d_alpha_epsilon}
  d_{\alpha,\epsilon}
  =
  \inf\{d>0:\,R_\alpha(d)\le\epsilon\}.
\end{equation}
Under the stable location model, Bayes-optimal detection at lag~$\ell$ has error at most~$\epsilon$ whenever
\begin{equation}
\label{eq:bayes_detectability_condition}
d_N(\ell)\ge d_{\alpha,\epsilon}.
\end{equation}
Using \eqref{eq:normalized_separation} and solving this condition for~$N$, gives the finite-sample requirement
\begin{equation}
\label{eq:min_N_of_ell_for_detection}
  N(\ell)
  =
  d_{\alpha,\epsilon}^{\,\kappa_\alpha}
  \left(
  \frac{\sigma_\alpha(\ell)}
  {|\overline m_\mu(\ell)|\,f(\ell)}
  \right)^{\!\kappa_\alpha}.
\end{equation}

This expression links the sample-complexity requirements at lag $\ell$ with the noise scale $\sigma_\alpha(\ell)$, the alignment magnitude $|\overline m_\mu(\ell)|$, the concentration exponent $\kappa_\alpha$, and the envelope $f(\ell)$.

\subsection{Learnability window}
\label{sec:learnability_window}

Starting from the Bayes-error detectability condition
in Eq.~\eqref{eq:bayes_detectability_condition}, together with the
normalized separation defined in Eq.~\eqref{eq:normalized_separation},
we solve for the minimal envelope mass required for reliable detection
at sample size~$N$. This yields the per-lag detectability threshold
\begin{equation}
\label{eq:theoretical_signal_threshold}
\varepsilon_{\mathrm{th}}(\ell)
=
\frac{
d_{\alpha,\epsilon}\,\sigma_\alpha(\ell)
}{
N^{1/\kappa_\alpha}\,|\overline m_\mu(\ell)|
}.
\end{equation}
Thus, a dependency at lag~$\ell$ is detectable at error level
$\epsilon$ whenever the envelope $f(\ell)$ exceeds
$\varepsilon_{\mathrm{th}}(\ell)$.
This threshold compactly summarizes the interplay of noise scale
$\sigma_\alpha(\ell)$, data size~$N$, alignment magnitude
$|\overline m_\mu(\ell)|$, and the critical separation $d_{\alpha,\epsilon}$.

We can now define the learnability window.

\begin{definition}[Learnability window]
\label{def:H_N_theoretical}
The \emph{learnability window} is
\begin{equation}
\label{eq:H_N_theoretical}
\mathcal{H}_N
=
\sup\big\{
\ell\ge1:\,
f(\ell)
\ge
\varepsilon_{\mathrm{th}}(\ell)
\big\}.
\end{equation}
\end{definition}
Intuitively, $\mathcal{H}_N$ is the largest lag for which the transported
gradient retains a recoverable signal.
Even if Jacobians are numerically stable, once $f(\ell)$ falls below
$\varepsilon_{\mathrm{th}}(\ell)$, heavy-tailed fluctuations dominate
and credit assignment becomes statistically infeasible.

\subsection{Scaling laws}
\label{sec:scaling_laws}

Here, we formalize two regularities that drive the scaling of $\mathcal{H}_N$.
Proofs appear in Appendix~\ref{app:lemma_proofs}.

\begin{lemma}[Envelope comparability]
\label{lem:mu_l1_monotone}
Fix $t$ and a neuron $q$.
Assume gate activations lie in $[0,1]$ and activation derivatives are
bounded in $[0,1]$.
\textbf{(i)}~The zeroth-order transport envelopes
$|\gamma^{(0,q)}_{t,\ell}|$ are nonincreasing in~$\ell$.
\textbf{(ii)}~If the first-order corrections satisfy the dominance
condition $|\gamma^{(1,q)}_{t,\ell}|\le c\,|\gamma^{(0,q)}_{t,\ell}|$
for some $c\in[0,1)$, then $|\Gamma^{(q)}_{t,\ell}|$ is comparable to
$|\gamma^{(0,q)}_{t,\ell}|$ up to the factor $(1\pm c)$.
\textbf{(iii)}~Under bounded adaptive base rates, the envelope
$f(\ell)=\sum_q|\mu^{(q)}_{t,\ell}|$ satisfies
\begin{equation}
\label{eq:envelope_comparability}
  \underline{\Lambda}(1-c)\,f_0(\ell)
  \;\le\;
  f(\ell)
  \;\le\;
  \overline{\Lambda}(1+c)\,f_0(\ell),
\end{equation}
where $f_0(\ell)=\sum_q|\gamma^{(0,q)}_{t,\ell}|$ is nonincreasing.
\end{lemma}

\begin{lemma}[Sample-complexity scaling and window bounds]
\label{lem:horizon_scaling_alpha}
Assume there exist constants $0<c_\sigma\le C_\sigma$ and
$0<c_m\le C_m$ such that, over the lags of interest,
$c_\sigma \le \sigma_\alpha(\ell) \le C_\sigma$ and
$c_m \le |\overline{m}_\mu(\ell)| \le C_m$,
as well as the envelope
comparability~\eqref{eq:envelope_comparability} of
Lemma~\ref{lem:mu_l1_monotone}.
Then, with $N(\ell)$ defined in Eq.~\eqref{eq:min_N_of_ell_for_detection}, there are
constants $0<c_\star\le C_\star$ for which
\begin{equation}
\label{eq:N_of_ell_sandwich_alpha}
c_\star\,f(\ell)^{-\kappa_\alpha}
\ \le\
N(\ell)
\ \le\
C_\star\,f(\ell)^{-\kappa_\alpha}.
\end{equation}
Consequently, with
$f_0^{\leftarrow}(y)=\sup\{\ell\ge1:\ f_0(\ell)\ge y\}$ denoting the
generalized inverse of the nonincreasing proxy~$f_0$,
\begin{equation}
\label{eq:H_window_sandwich_alpha}
f_0^{\leftarrow}\!\Big(\widetilde{C}_1\,N^{-1/\kappa_\alpha}\Big)
\ \le\
\mathcal{H}_N
\ \le\
f_0^{\leftarrow}\!\Big(\widetilde{C}_2\,N^{-1/\kappa_\alpha}\Big),
\end{equation}
where $\widetilde{C}_1$, $\widetilde{C}_2$ are positive constants that
absorb the comparability factors from
Eq.~\eqref{eq:envelope_comparability} and the boundedness conditions on
$\sigma_\alpha$ and $|\overline{m}_\mu|$.
\end{lemma}

The first lemma establishes that the zeroth-order gate-product
envelope $f_0(\ell)$ is nonincreasing and that the full envelope
$f(\ell)$ is comparable to it under a first-order dominance condition
(details in Appendix~\ref{app:lemma_proofs}).
The second lemma makes explicit that the per-lag sample complexity
grows as the inverse $\kappa_\alpha$-power of $f(\ell)$.
Together, they reduce the problem of characterizing $\mathcal{H}_N$
to inverting the decay profile of the monotone proxy~$f_0(\ell)$ at the
level~$N^{-1/\kappa_\alpha}$.
In other words, once $\sigma_\alpha(\ell)$ and
$|\overline{m}_\mu(\ell)|$ have bounded magnitude, the temporal reach of
learnability is controlled by the speed with which $f_0(\ell)$ decays
with~$\ell$ and by the heavy-tailed concentration
exponent~$\kappa_\alpha$.

In practice, gated RNNs may exhibit intricate envelope profiles, including mixtures of decay regimes or multi-phase behavior across different lag ranges. 
To develop analytic insight, we focus on three canonical asymptotic scaling profiles: logarithmic, power-law, and exponential decay. 
These regimes provide a structural classification of how $f(\ell)$, understood up to multiplicative constants via its monotone proxy $f_0(\ell)$, can attenuate in the large-lag limit and form the basis of the corresponding learnability scaling laws. 
Real networks need not follow a single regime uniformly across all lags; finite-dimensional implementations and bounded gates inevitably induce crossover behavior, typically resulting in exponential-like cutoffs at sufficiently large $\ell$. 
More complex envelope shapes can nevertheless be analyzed within the same framework by decomposing them into these canonical asymptotic components.

For each canonical asymptotic regime of $f(\ell)$, we obtain the corresponding large-$\ell$ scaling laws for the learnability window:
\begin{enumerate}[label=(\roman*), itemsep=2pt]
\item \textbf{Logarithmic decay:}
if $f(\ell)\!\asymp\!c/[\log(1+\ell)]^{\vartheta}$ with $\vartheta\!>\!0$, then
$N(\ell)\!\asymp\![\log(1+\ell)]^{\kappa_\alpha\vartheta}$ and $\log\!\left(1+\mathcal{H}_N\right) \;\asymp\; N^{1/(\kappa_\alpha\vartheta)}$.
Equivalently,
\begin{equation}
\label{eq:LW_scaling_law_log_env_decay}
\mathcal{H}_N = \exp\!\left(\Theta\left(N^{1/(\kappa_\alpha\vartheta)}\right)\right)-1.
\end{equation}

\item \textbf{Power-law decay:}
if $f(\ell)\!\asymp\!c\,\ell^{-\beta}$ with $\beta>0$, then $N(\ell)\!\asymp\!\ell^{\kappa_\alpha\beta}$ and
\begin{equation}
\label{eq:LW_scaling_law_poly_env_decay}
\mathcal{H}_N\!\asymp\!N^{1/(\kappa_\alpha\beta)}.
\end{equation}
\item \textbf{Exponential decay:}
if $f(\ell)\!\asymp\!c\,\lambda^{\ell}$ with $\lambda\!\in(0,1)$, then 
$N(\ell)\!\asymp\!\lambda^{-\kappa_\alpha\ell}$ and 
\begin{equation}
\label{eq:LW_scaling_law_exp_env_decay}
\mathcal{H}_N\!\asymp\!(\log N)/[\kappa_\alpha\log(1/\lambda)].
\end{equation}
\end{enumerate}

In these expressions, the symbol $\asymp$ denotes equality at the level of asymptotic order.\footnote{We write $f \asymp g$ to mean that $c_1\, g \le f \le c_2\, g$ for positive constants $c_1, c_2$ independent of $\ell$ and $N$.
Unlike $f \sim g$ (which requires $f/g \to 1$), the symbol $\asymp$ preserves the correct scaling exponent while absorbing multiplicative constants, including the envelope prefactor $c$ and the architecture- and threshold-dependent constants ($c_\sigma, C_\sigma, c_m, C_m, d_{\alpha,\epsilon}$) that appear in the sandwich bounds of Appendix~\ref{app:scaling_derivations}.
For instance, $f(\ell)\asymp c\,\ell^{-\beta}$ gives $N(\ell)\asymp c^{-\kappa_\alpha}\ell^{\kappa_\alpha\beta}$, and the factor $c^{-\kappa_\alpha}$ is absorbed by $\asymp$.
In the logarithmic-envelope case, however, inversion places these constants inside an exponential. We therefore state the asymptotic order through $\log(1+\mathcal H_N) \asymp N^{1/(\kappa_\alpha\vartheta)}$, or equivalently through a $\Theta$ term in the exponent, rather than using a multiplicative $\asymp$ relation for $\mathcal{H}_N$ itself.
When $\kappa_\alpha\vartheta>1$, this is stretched-exponential growth in~$N$.}
These scaling laws characterize how envelope decay and heavy-tailed fluctuations jointly determine the large-lag behavior of temporal learnability.
Exponential decay corresponds to rapid forgetting and only logarithmic growth of the horizon; power-law decay yields polynomial horizon growth; and logarithmic decay represents a boundary regime with rapidly expanding $\mathcal{H}_N$.
In all cases, smaller $\alpha$ compresses $\mathcal{H}_N$ by increasing $\kappa_\alpha$ and slowing statistical concentration.
Full derivations and the role of constant factors are provided in
Appendix~\ref{app:scaling_derivations}.

\subsection{Theoretical implications}
\label{sec:learnability_theoretical_implications}

Combining the sample-complexity bound \eqref{eq:min_N_of_ell_for_detection} with the learnability window analysis yields several theoretical insights into how the coupled architecture--optimizer dynamics, noise statistics, and finite data interact to determine temporal learnability.

\paragraph{(i) Envelopes set sample complexity under heavy-tailed noise.}
The detectability of a lag-$\ell$ dependency is governed by the standardized separation~\eqref{eq:normalized_separation}.
This expression exposes the key leverage imbalance: the envelope $f(\ell)$ enters the detectable signal linearly, whereas the number of sequences enters only through the concentration factor $N^{1/\kappa_\alpha}$.
For a fixed envelope class, increasing $N$ improves detectability only sub-linearly, equivalently by lowering the required envelope magnitude at rate $N^{-1/\kappa_\alpha}$.
By contrast, changing the decay class changes how rapidly $f(\ell)$ falls with lag, and therefore changes how far in time the same threshold can be pushed.
Thus, additional data move the operating point within a fixed decay regime, whereas envelope geometry determines the regime itself.
This is made explicit by the bound~\eqref{eq:N_of_ell_sandwich_alpha}, which yields the \emph{master proportionality}
\begin{equation}
\label{eq:master_prop_maintext}
N(\ell)\;\asymp\; f(\ell)^{-\kappa_\alpha}.
\end{equation}

Concrete scaling laws follow once the master proportionality is combined with a specific decay regime for $f(\ell)$; see Appendix~\ref{app:scaling_derivations} for details.

\paragraph{(ii) The tail index $\alpha$ governs statistical efficiency.}
The concentration exponent $\kappa_\alpha$ \eqref{eq:kappa_alpha_def} quantifies how rapidly noise averages out when statistics are aggregated across independent sequences.
For Gaussian fluctuations ($\alpha=2$), one recovers the familiar concentration
rate $N^{-1/2}$; for $\alpha<2$, the slower rate
$N^{-1/\kappa_\alpha}=N^{-(1-1/\alpha)}$ substantially weakens the effective SNR.
Thus, isolating the concentration effect by holding the envelope $f(\ell)$ and the lag-dependent noise scale fixed, smaller values of $\alpha$ increase $\kappa_\alpha$, compress the learnability window $\mathcal{H}_N$, and increase the number of samples $N$ required for reliable detection.

\paragraph{(iii) Architectural and training implications.}
The effective scaling regime is jointly determined by the envelope $f(\ell)$ and noise statistics, rather than defined a priori by architectural and optimizer-specific details.
Architectures that permit slowly decaying envelopes are, in principle, better positioned to support long-range learnability under adverse noise statistics.
However, the training dynamics and the realized noise regime jointly determine which decay profile is effectively stabilized in practice.
Consequently, optimization choices that alter fluctuation structure or stabilize training can improve learnability either by modifying the realized envelope, by changing the effective noise scale, or by shifting the effective tail behavior of the aggregated statistic.

\paragraph{(iv) Vanishing learnability as a statistical obstruction.}
In the limiting case where the effective lag signal vanishes identically at all
nonzero lags, the learnability window collapses: $\mathcal H_N=0$ for all
finite $N$.
Equivalently, $\Delta(\ell)=0$ and hence the standardized separation
$d_N(\ell)$ in Eq.~\eqref{eq:normalized_separation} is zero for every
sample size.
The two location models are then identical, the Bayes error remains
$1/2$, and no decision rule can distinguish the two hypotheses better
than chance.
This is not an optimization failure or a data limitation, but an
\emph{statistical obstruction}: the transported gradients carry no
statistically usable signal at nonzero lags.

\paragraph{(v) Vanishing gradients.}

The classical gradient pathologies are extreme cases of envelope decay, not separate phenomena. Strong, uniform contraction of the recurrent Jacobians, the standard condition for vanishing gradients, forces the envelope into the fast-exponential class, in which $f(\ell)$ decays quickly with the lag. By the master relation, the number of samples needed to detect a lag then grows exponentially with that lag, so the learnability window can grow only logarithmically with the dataset size, no matter how much data is added. This is the \emph{data-complexity wall}: vanishing gradients do not merely shrink the mean gradient, they render long-range dependencies statistically undetectable at any realistic sample size, and heavier-tailed fluctuations only tighten the wall. Adaptive optimizers can rescale the envelope but cannot change this exponential rate. Appendix~\ref{app:vanishing_exploding} makes these statements precise.

\section{Empirical validation}
\label{sec:experiments}

The purpose of this section is not to compare architectures in terms of predictive performance, but to empirically probe the mechanisms identified by the theory: how envelope decay, noise statistics, gating structure, and optimizer adaptation jointly determine finite-horizon learnability.

Here, learnability refers to the statistical detectability of
lagged gradient contributions during training. Detectability is a
necessary (though not sufficient) condition for successful learning of
long-range dependencies: if the gradient signal originating from a past state is not statistically distinguishable from noise, consistent information propagation cannot occur even when gradients remain numerically stable.

Consequently, the experiments are designed to test the structural predictions of the theory.
In particular, we aim to:
(i) measure the empirical decay profiles of the envelope $f(\ell)$ and assess how they align with the canonical decay regimes considered by the theory;
(ii) quantify how the resulting empirical learnability windows
$\widehat{\mathcal H}_N$ vary with gating structure, optimizer dynamics,
and sample size, and examine whether their scaling behavior is
consistent with the regime implied by the observed envelope decay;
and
(iii) investigate how the neuronwise time-scale spectra $\{\tau_q\}_q$
and the estimated matched statistic fluctuations jointly relate to the
empirically realized learnability regimes.

\subsection{Experimental setup}
\label{sec:exp_setup}

\paragraph{Task and data.}
We consider a synthetic regression task defined on input--output pairs
$\{(x_t, y_t)\}_{t=1}^T$, where the inputs $x_t \in \mathbb{R}^{16}$ are
i.i.d.\ standard Gaussian vectors, $x_t\sim\mathcal{N}(0, I_{16})$,
and the target $y_t \in \mathbb{R}$ is scalar.
All experiments use sequences of length $T = 1536$.

The target is generated as a weighted sum of delayed inputs projected
onto a fixed direction,
\begin{equation}
  y_t = \sum_{k=1}^6 c_k\, u^\top x_{t-\ell_k} + \varepsilon_t,
  \label{eq:exp_task}
\end{equation}
where $u \in \mathbb{R}^{16}$ is a unit vector drawn uniformly on the
sphere and held fixed across all sequences,
the task lags are
$\{\ell_k\} = \{64,\,128,\,256,\,384,\,512,\,768\}$,
the mixing coefficients are
$\{c_k\} = \{0.60,\,0.50,\,0.40,\,0.30,\,0.22,\,0.16\}$,
and $\varepsilon_t$ is independent Gaussian noise with standard deviation
$\sigma_{\mathrm{noise}} = 0.4$.
The random unit $u$ reduces the $16$-dimensional input to a scalar
target without privileging any coordinate direction, forcing the model
to recover the relevant projection from data; fixing $u$ across
sequences keeps the regression target well-defined and consistent.
For positions $t<\max_k\ell_k$ the sum in~\eqref{eq:exp_task}
truncates naturally to those lag terms with $\ell_k\le t$, so the
first $\max_k\ell_k=768$ time steps of each sequence carry a
progressively growing number of active components before the full
six-lag structure is reached.
These delayed components induce informative dependencies spanning short,
intermediate, and long temporal scales, with the longest lag matching
the diagnostic range upper bound.
This makes the task a controlled reference for the learnability theory:
i.i.d.\ inputs remove input-side temporal autocorrelation that could
confound the envelope, and six well-separated task lags probe a broad
range of time scales within a single analytically tractable regression;
because the structural scaling predictions depend only on the trained
envelope $f(\ell)$ and not on task-specific features, a heterogeneous
benchmark suite would introduce task-specific confounds rather than
sharpen those predictions.

To probe the full decay profile of the learned envelope beyond the
specific task lags, all diagnostic quantities are evaluated on a
separate lag grid $\ell \in [4, 768]$ consisting of $192$ uniformly
spaced values.
This grid covers both sub-task and supra-task temporal scales.

\paragraph{Architectures.}
We consider five recurrent architectures described in
Appendix~\ref{app:RNNs_jacobians_and_elr}, spanning increasing gating
expressivity and potential time-scale heterogeneity:
ConstGate, SharedGate, DiagGate, GRU, and LSTM.
All models use hidden dimension $H=256$, layer normalization on the
hidden state, and a linear readout $y_t = w^\top h_t$ trained with
mean-squared error loss.
Recurrent weight matrices are initialized orthogonally and
biases are initialized to zero unless otherwise stated.
Gate pre-activations in learned-gate models are initialized near zero,
yielding initial gate values close to $\tfrac12$ and well-conditioned
initial time scales, consistent with the time-scale initialization
heuristic of~\cite{livi2025timescale}.
In particular, the LSTM forget gate is initialized near
$\sigma(b_f)\approx 0.5$ in the main experiments.
The GRU update-gate bias is set so that $\sigma(b_z) \approx 0.05$,
biasing the gate toward retaining the previous hidden state at
initialization.
For brevity, in figures and legends we denote ConstGate,
SharedGate, DiagGate, LSTM, and GRU by
\emph{const}, \emph{shared}, \emph{diag}, \emph{lstm}, and
\emph{gru}, respectively.

\paragraph{Training protocol.}
All models are trained using mini-batch
AdamW~\cite{loshchilov2019decoupled} with base learning rate
$\mu = 2\times10^{-4}$, weight decay $\lambda = 10^{-4}$, batch size
$B = 384$, and gradient clipping at $\ell_2$ norm $1.0$.
Training uses $8000$ independent sequences of length $T = 1536$
for $1500$ epochs, with model selection based on the best
validation loss.
The validation set consists of $1024$ additional sequences generated
from the same data-generating process and kept separate from both the
training and diagnostic sets; all diagnostics are computed using the
selected best-validation checkpoint.
No learning-rate schedule is applied, and all architectures share
identical optimization hyperparameters.
Adaptive optimizers such as AdamW are known to preserve heavy-tailed
gradient fluctuations even in the presence of gradient
clipping~\cite{Savelii2024ClippingAdamHeavyTailed}.
We verified that disabling gradient clipping yields qualitatively
similar envelope and noise statistics (results not shown), indicating
that clipping affects numerical stability but not the scaling regimes
of interest.

All experiments are repeated across five independent random seeds,
each controlling the parameter initialization and the data-generating
process. For each seed, the learnability pipeline is executed with
$K=50$ independent projection directions $w$; results are reported as cross-seed averages unless otherwise stated.
Detailed envelope-fit tables and additional experiments with other optimizers are reported in Appendix~\ref{app:fitting_details_and_optimizer_regimes}.

\paragraph{Diagnostic evaluation.}
After training, parameters are frozen and all diagnostics are computed on an
independent set of $12000$ fresh sequences generated from the same
data-generating process, with no overlap with the training set.
This protocol is not a simplification: the learnability window characterizes the statistical detectability of lag-dependent structure at a given parameter configuration, not along a training trajectory.
The scaling laws describe the structural regime (exponential, polynomial, or logarithmic) that the trained network has settled into, and evaluating them requires only a fixed snapshot of the parameters, gate activations, and optimizer state.
Computing the matched statistic and the envelope on an independent diagnostic set ensures that these estimates reflect the transport and noise structure of the trained model.

\paragraph{Empirical envelope.}
For each trained model, the transport factors
$\Gamma^{(q)}_{t,\ell}$ are computed using the architecture-specific expressions derived in
Appendix~\ref{app:RNNs_jacobians_and_elr}.
For AdamW, the adaptive base rates $\Lambda^{(q)}_{r,\ell}$ are
obtained from the Rayleigh
projection~\eqref{eq:rayleigh}; for plain SGD they reduce to the constant
$\Lambda^{(q)}_{r,\ell}=\mu$.
For each diagnostic lag $\ell$ and unit $q$, we summarize the effective
learning-rate magnitude by averaging $|\mu_{t,\ell}^{(q)}|$ over admissible time
positions and independent diagnostic sequences. Writing
$\langle \cdot \rangle_{t,n}$ for this joint average, the empirical envelope is
defined as
\begin{equation}
\label{eq:empirical_envelope_definition}
\hat f(\ell)
=
\sum_{q}
\Big\langle |\mu_{t,\ell}^{(q)}| \Big\rangle_{t,n},
\end{equation}
which corresponds to the empirical $\ell_1$ aggregation of
neuronwise effective learning rates at lag~$\ell$.

\paragraph{Heavy-tailed statistics.}
For each diagnostic lag $\ell$, we compute the empirical matched statistic
introduced in Sec.~\ref{sec:binary_detection_matched_statistic} on the
diagnostic set.
From these statistics, we estimate the parameters of an
$\alpha$-stable law using the empirical characteristic function (ECF)
regression estimator of~\cite{koutrouvelis1980regression},
obtaining a lag-dependent tail index $\hat\alpha(\ell)$
and scale parameter $\hat\sigma_\alpha(\ell)$.
As a robustness check, we also estimate $\alpha$ using the
McCulloch (MCC) quantile-based method~\cite{mcculloch1986simple},
bootstrapped over $500$ resamples.
For each lag, $\hat\alpha(\ell)$ is estimated over the diagnostic
sequences: each sequence provides one matched-statistic scalar after
averaging across the $K=50$ projection directions, and the ECF and
MCC estimators are fit to the resulting set of sequence-level
scalars. $\hat\alpha(\ell)$ is therefore the tail index of the
projection-averaged matched statistic at lag $\ell$.
Projection-wise tail-index distributions are reported separately in
Appendix~\ref{app:projection_alpha_diagnostic} as a diagnostic of
directional variability and are not pooled into the per-(seed, lag)
estimates reported here.
Although we estimate a tail index at every diagnostic lag, the
closed-form learnability laws in Sec.~\ref{sec:learnability_theory}
require one concentration exponent
$\kappa_\alpha=\alpha/(\alpha-1)$.
We therefore reduce the lag-wise estimates to a single effective tail
index for each model, taking $\hat\alpha$ to be the median of
$\hat\alpha(\ell)$ over diagnostic lags. It is this single
$\hat\alpha$ that enters the empirical SNR
in Eq.~\eqref{eq:empirical_SNR} (and hence the empirical learnability
window in Eq.~\eqref{eq:H_N_empirical}) through the concentration
exponent; the noise scale itself remains lag-dependent through
$\hat\sigma_\alpha(\ell)$.

\paragraph{Projection aggregation.}
The matched statistic in Sec.~\ref{sec:binary_detection_matched_statistic} is defined for a projection direction $w\in\mathbb{R}^P$.
In the empirical pipeline, we reduce projection-specific variance by using $K=50$ independent directions $w_1,\dots,w_K$ drawn uniformly on the unit sphere.
For each sequence and lag, the matched statistics are averaged across these $K$ directions before tail-index estimation, producing one sequence-level scalar per diagnostic sequence.
Thus, the tail-index estimators are fit to one sequence-level value per diagnostic sequence per lag, not to $K$ times as many projection-level values.
The envelope $f(\ell)$ is defined before this projection step (Eq.~\eqref{eq:empirical_envelope_definition}) and is therefore unchanged by the number of projection directions.
Appendix~\ref{app:projection_validation} validates that the resulting one-dimensional matched statistic is a stable proxy for the lag-specific transport object across tested seeds and projection directions, and Appendix~\ref{app:multi_projection} gives the corresponding multi-projection scaling formulas.

\paragraph{Empirical learnability window and empirical matched statistic.}
The theoretical learnability window in Eq.~\eqref{eq:H_N_theoretical}
is equivalently characterized by the Bayes-error separation condition
\(d_N(\ell)\ge d_{\alpha,\epsilon}\), where \(d_N(\ell)\) is the
standardized separation in Eq.~\eqref{eq:normalized_separation}.
Thus, the empirical task is to estimate this same standardized
signal-to-noise quantity from diagnostic data.

The population separation depends on the unsigned signal amplitude
\(|\Delta(\ell)|\), the stable noise scale \(\sigma_\alpha(\ell)\), and
the concentration factor \(N^{1/\kappa_\alpha}\).
In practice, the oracle sign factors
\(\operatorname{sgn}(m_q(\ell))\) in Eq.~\eqref{eq:matched_statistic}
are not available, so we use the raw signed empirical matched statistic
\begin{equation}
\label{eq:empirical_raw_matched_statistic}
\widetilde{S}_{t,\ell}
\;=\;
\sum_{q=1}^{H}\mu^{(q)}_{t,\ell}\,\zeta^{(q)}_{t,\ell}.
\end{equation}
Applying the same two-stage aggregation used in
Eqs.~\eqref{eq:theoretical_sequence_average}--\eqref{eq:theoretical_cross_sequence_average}
gives
\begin{equation}
\label{eq:empirical_cross_sequence_average}
\widetilde{S}_N(\ell)
\;=\;
\frac{1}{N(T-\ell)}
\sum_{n=1}^{N}\sum_{t=\ell+1}^{T}
\widetilde{S}^{(n)}_{t,\ell}.
\end{equation}
Because the raw statistic may have either sign, while Bayes
detectability depends only on the separation between the two opposite
location shifts in Eq.~\eqref{eq:alpha_stable_model}, we estimate the
unsigned signal amplitude by
\[
\widehat{\Delta}(\ell)
=
\bigl|\widetilde{S}_N(\ell)\bigr|.
\]

Writing
\(\widehat{\kappa}_\alpha=\hat\alpha/(\hat\alpha-1)\), the empirical
standardized separation, or empirical SNR, is then
\begin{equation}
\label{eq:empirical_SNR}
\widehat{\mathrm{SNR}}(\ell,N)
\;=\;
\frac{\widehat{\Delta}(\ell)\;
      N^{1/\widehat{\kappa}_\alpha}}
     {\hat\sigma_\alpha(\ell)}.
\end{equation}
This is the plug-in counterpart of \(d_N(\ell)\): the population
signal, tail index, and stable scale are replaced by their empirical
estimates.

Since the theoretical constant \(d_{\alpha,\epsilon}\) is defined
through the idealized stable-location Bayes error, we set the empirical
SNR threshold \(\epsilon_{\mathrm{snr}}\) directly.
Here \(\epsilon_{\mathrm{snr}}\) is an empirical separation threshold,
not the Bayes error level \(\epsilon\) used to define
\(d_{\alpha,\epsilon}\).
The empirical learnability window is therefore
\begin{equation}
\label{eq:H_N_empirical}
\widehat{\mathcal{H}}_N
\;=\;
\max\bigl\{\ell :
\widehat{\mathrm{SNR}}(\ell,N) > \epsilon_{\mathrm{snr}}\bigr\}.
\end{equation}
Equation~\eqref{eq:H_N_empirical} is thus the direct empirical analogue of Eq.~\eqref{eq:H_N_theoretical}.
The use of the raw signed statistic \eqref{eq:empirical_raw_matched_statistic} makes this estimate conservative
relative to the oracle matched statistic, because sign cancellations can reduce \(\widehat{\Delta}(\ell)\).

For interpretability, we report the threshold through its inverse,
the \emph{noise tolerance} $1/\epsilon_{\mathrm{snr}}$, which represents the
maximum tolerated inverse SNR in the empirical detection criterion.
Thus, $1/\epsilon_{\mathrm{snr}} = 0.1$ means detection requires
$\widehat{\mathrm{SNR}}(\ell,N) > 10$, while $1/\epsilon_{\mathrm{snr}} = 0.05$
requires $\widehat{\mathrm{SNR}}(\ell,N) > 20$. Smaller
$1/\epsilon_{\mathrm{snr}}$ therefore imposes a stricter detectability
requirement. Changing $1/\epsilon_{\mathrm{snr}}$ adjusts the operating point of
the empirical learnability window, while preserving the same scaling
form in $N$, signal amplitude, and noise scale. All experiments use
$1/\epsilon_{\mathrm{snr}} = 0.05$ ($\epsilon_{\mathrm{snr}} = 20$).
The learnability window is evaluated on a grid of $54$ sample sizes
$N \in \{25, 50, \ldots, 25600\}$, spaced densely at small $N$
(steps of $25$) and logarithmically at large $N$.

\paragraph{Time-scale extraction.}
The effective learning rates $\mu_{t,\ell}^{(q)}$ attenuate the gradient contribution of neuron~$q$ as a function of lag.
To summarize each neuron's temporal reach with a single scalar, we fit the exponential model
\[
|\mu_{t,\ell}^{(q)}|
\;\approx\;
C_q \exp(-\ell / \tau_q)
\]
to the diagnostic measurements.
The exponential form is motivated by the dominant zeroth-order structure of the transport factors, which are products of gating values across the lag window (Appendix~\ref{app:RNNs_jacobians_and_elr}); for architectures with multiple interacting gates and adaptive optimizers, $\tau_q$ should be understood as an effective time scale rather than an exact decay constant.
What matters for the learnability analysis is the collective distribution $\{\tau_q\}_q$: narrow spectra indicate nearly synchronized dynamics, while broad spectra reflect heterogeneous mixtures of time scales that extend the model's temporal reach.
The shape of this spectrum also determines the large-lag behaviour of the envelope $\hat f(\ell)$.

\subsection{Results}
\label{sec:exp_results}

\begin{figure}[tp!]
  \centering

  \begin{subfigure}[b]{0.32\textwidth}
    \centering
    \includegraphics[width=\textwidth]{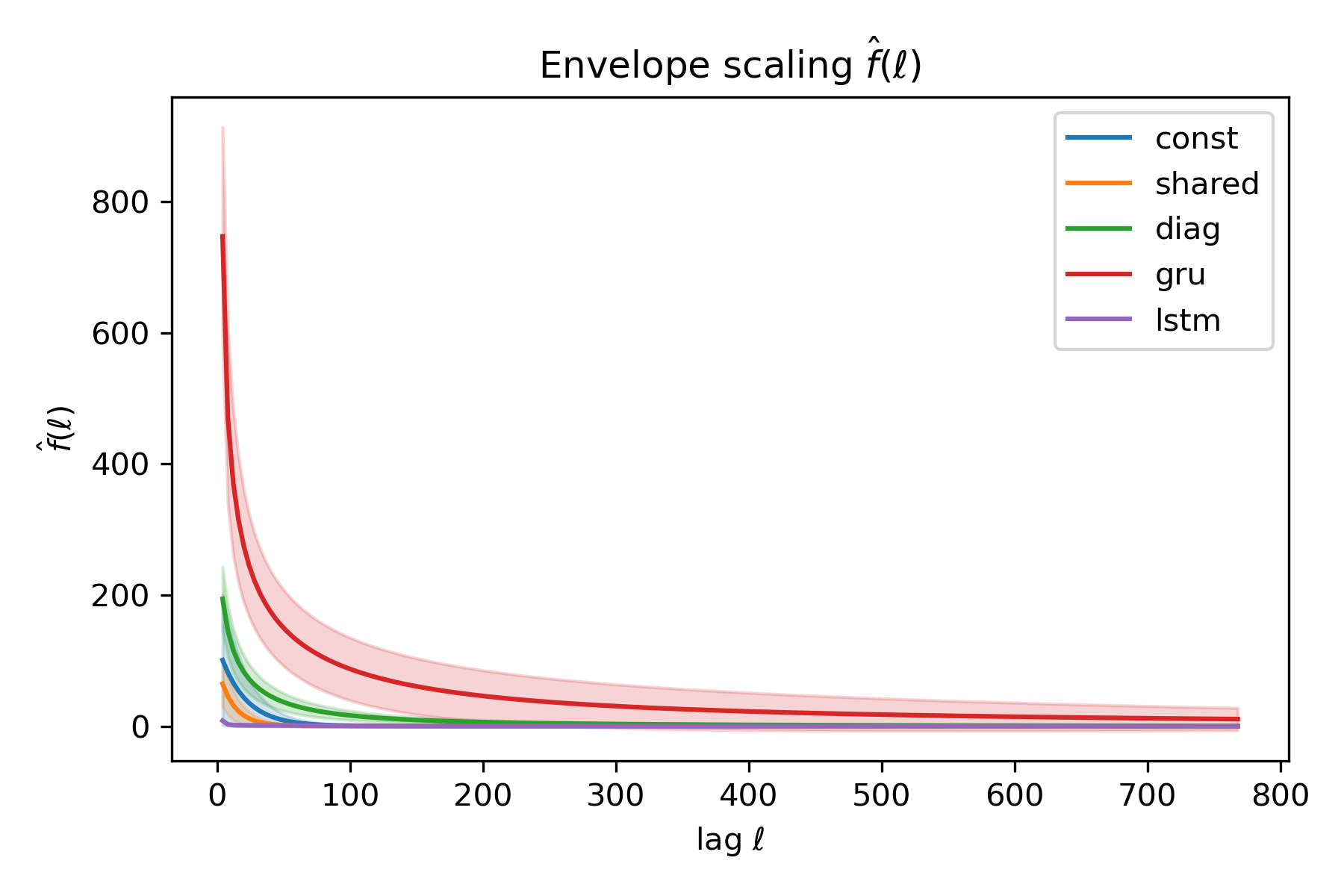}
    \caption{Linear scale.}
    \label{fig:envelope_linear}
  \end{subfigure}
  \hfill
  \begin{subfigure}[b]{0.32\textwidth}
    \centering
    \includegraphics[width=\textwidth]{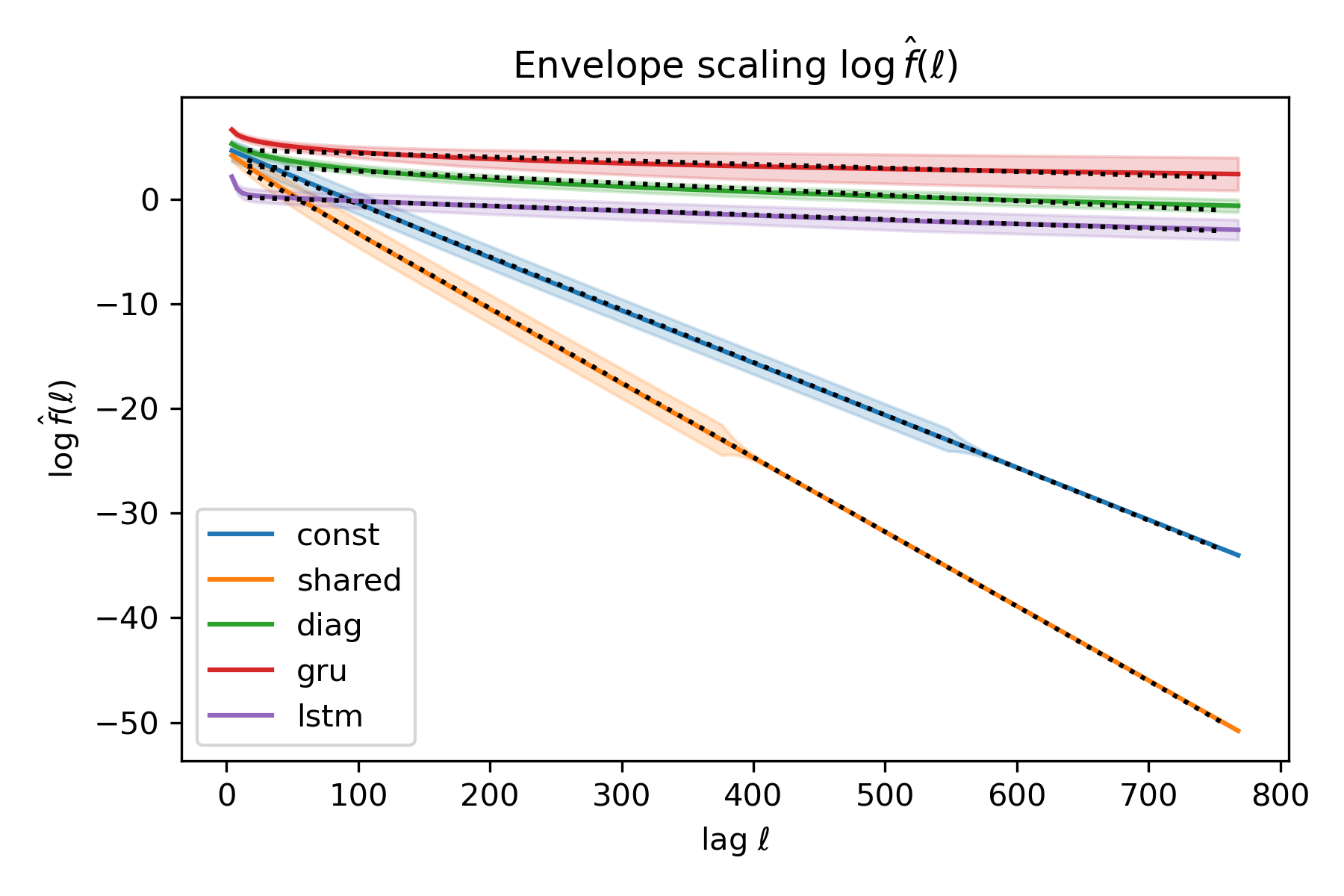}
    \caption{Semi-log scale.}
    \label{fig:envelope_semilog}
  \end{subfigure}
  \hfill
  \begin{subfigure}[b]{0.32\textwidth}
    \centering
    \includegraphics[width=\textwidth]{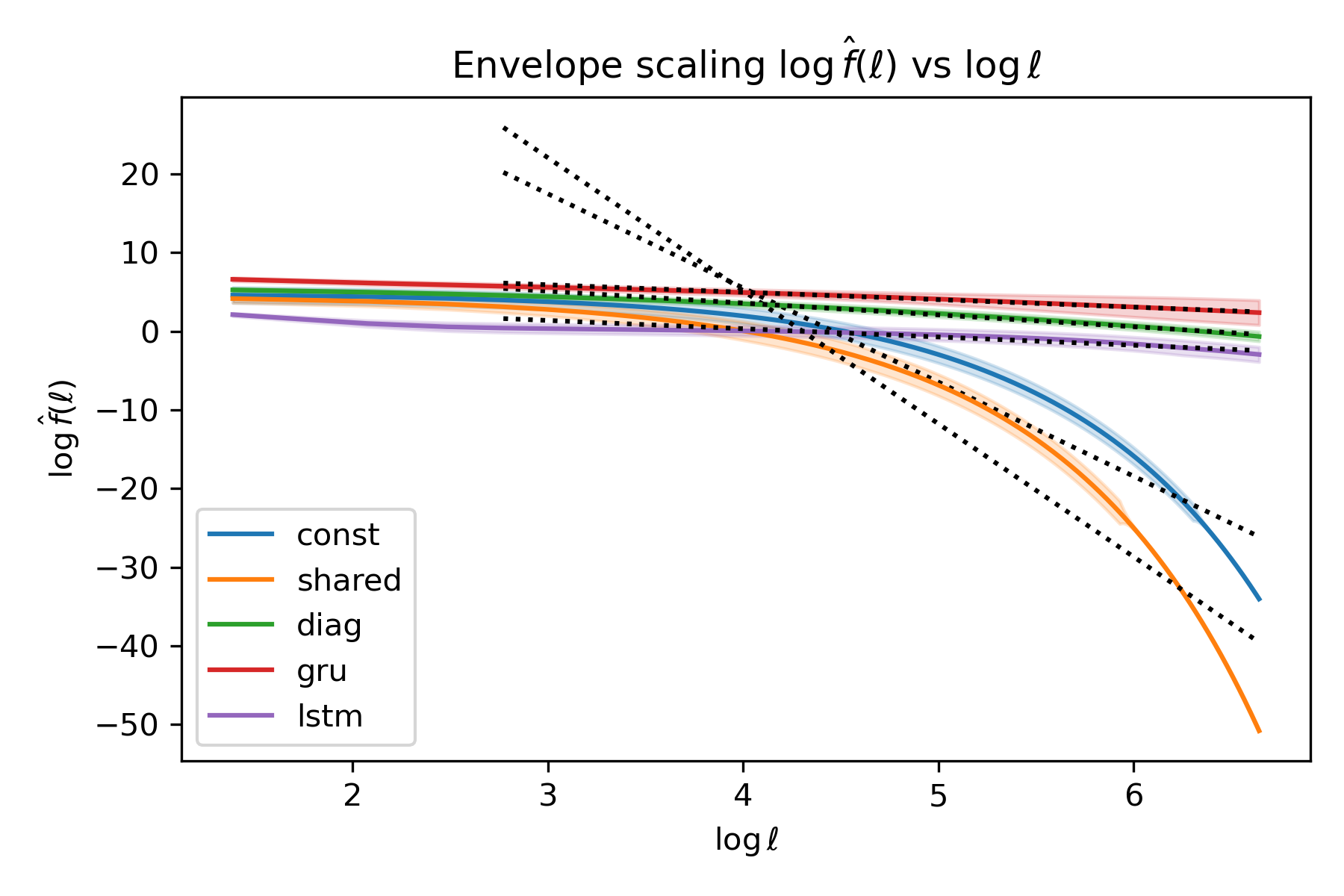}
    \caption{Log--log scale.}
    \label{fig:envelope_loglog}
  \end{subfigure}

\caption{Envelopes of the effective learning rates $\hat f(\ell)$
for ConstGate, SharedGate, DiagGate, GRU, and LSTM models, averaged
across training seeds (shaded regions: cross-seed standard
deviation).
\textbf{(a)} Linear scale.
\textbf{(b)} Semi-logarithmic representation highlights the rapid
exponential decay of ConstGate and SharedGate (dotted lines: linear
fits) and the much slower attenuation of DiagGate, GRU, and LSTM.
\textbf{(c)} Log--log representation reveals approximate polynomial
scaling over intermediate lags for DiagGate and GRU (dotted lines:
power-law fits), with an eventual exponential cut-off implied by
finite-state dynamics; the LSTM envelope sits closer to the
exponential boundary in this experimental configuration.}
  \label{fig:envelope_scalings}
\end{figure}

\paragraph{Envelope scaling.}
Figure~\ref{fig:envelope_scalings} shows the estimated envelopes
$\hat f(\ell)$ for ConstGate, SharedGate, DiagGate, GRU, and LSTM,
averaged over the training seeds.
The seed-averaged envelopes display an overall decay with increasing
lag, but with markedly different attenuation regimes.
We emphasize that the experimental configuration used here was not
tuned to drive the architectures into any particular operating regime;
the realized envelope shapes reported below are therefore one specific
instance within the broader theoretical classification, and other
configurations (different optimizers, learning rates, or initialization
schemes) would generally land at different points within the same theoretical classification.

ConstGate and SharedGate display rapid envelope decay.
On a semi-logarithmic scale (b), both curves are well approximated by straight lines, consistent with exponential attenuation.
Exponential fits of the form $\hat f(\ell)\approx c\,\exp(-\lambda \ell)$ yield decay rates $\lambda_{\text{const}}\approx 0.050$ and $\lambda_{\text{shared}}\approx 0.071$, with coefficients of determination $r^2 \approx 1$ in both cases.
Power-law fits perform substantially worse ($r^2 \approx 0.85$ for both), confirming that these architectures lie firmly in the exponential regime.
Their envelopes decay by 16--32 orders of magnitude across the diagnostic range, indicating a very short effective temporal reach.

DiagGate and GRU exhibit qualitatively slower attenuation.
On a log--log scale (c), both envelopes display approximately linear trends over the diagnostic range.
A power-law fit $\hat f(\ell)\approx c\,\ell^{-\beta}$ yields $\beta_{\text{diag}}\approx 1.51$ ($r^2 \approx 0.98$) and $\beta_{\text{gru}}\approx 0.95$ ($r^2 \approx 0.99$), in both cases dominating the corresponding exponential fits ($\lambda_{\text{diag}}\approx 0.006$, $r^2\approx 0.93$; $\lambda_{\text{gru}}\approx 0.004$, $r^2\approx 0.90$).
We interpret DiagGate and GRU as operating in a power-law-like regime with a stretched cutoff: empirically consistent with polynomial scaling across the explored window, with an inevitable exponential cut-off at sufficiently large lags due to finite-state dynamics.

LSTM occupies an intermediate position between the constrained
exponential regime and the gated power-law-with-cutoff regime.
The simple exponential fit ($\lambda_{\text{lstm}}\approx 0.004$, $r^2 \approx 0.99$) edges out the simple power-law fit ($\beta_{\text{lstm}}\approx 1.05$, $r^2\approx 0.89$) in the present configuration, but the fitted rate is much smaller than for the exponential baselines ($\lambda_{\text{const}}\approx 0.050$ and $\lambda_{\text{shared}}\approx 0.071$).
Thus LSTM is not behaving like a short-memory exponential model; over
the explored horizon it can be read either as a very slowly decaying
exponential or as a strongly tempered power-law.
The cleanest structural description is therefore a tempered power-law
$\hat f(\ell) \approx c\,\ell^{-\beta}\exp(-(\ell/\ell_c)^k)$ with
$k\approx 0.7$ and $\ell_c\approx 300$, placing LSTM between the gated
regime ($k\approx 0.25$ for DiagGate and GRU) and the pure exponential
regime ($k=1$ for ConstGate and SharedGate).
This realized operating point is contingent on the specific
experimental setup employed here, including the LSTM forget-gate
bias initialization.
Different choices of these hyperparameters may shift the LSTM
envelope to different points within the same theoretical
classification.

The interpretation of intermediate power-law-like regimes for DiagGate
and GRU is further supported by the decay of the complementary
cumulative distribution function (CCDF) of the neuronwise effective
time scales, reported below. In these architectures, the tail of the
CCDF exhibits a slow decay compatible with a broad distribution of
time scales, followed by a clear attenuation at large lags. This
pattern is consistent with approximate power-law-like behavior over
intermediate lags, combined with an inevitable exponential cut-off
induced by finite-dimensional state dynamics. Detailed envelope-fit tables are shown in Appendix~\ref{app:fitting_details_and_optimizer_regimes}.

Collectively, these results reveal a hierarchy of temporal regimes:
homogeneous scalar gating produces fast exponential decay; diagonal
gating induces slow, approximately power-law attenuation; and
multi-gate architectures realize broad mixtures of time scales whose
realized envelope position along the cutoff continuum depends on
gating expressivity, optimizer, and initialization jointly
(GRU lands deep in the gated power-law regime in this configuration;
LSTM sits closer to the exponential boundary).
Importantly, the corresponding state Jacobians remain numerically
well-conditioned in all cases (results not shown), indicating that
Jacobian stability alone does not determine finite-horizon
learnability.
Rather, it is the geometry of the envelope that governs how strongly
lagged gradient contributions influence parameter updates across
temporal lags.
All models exhibit consistent training loss reduction and stable
optimization trajectories (learning curves not shown), indicating
that they extract non-trivial signal from the task under the shared
training protocol; we do not claim that any of the architectures
achieves an optimal fit, only that learning proceeds in a stable
manner sufficient for the diagnostics reported here.
However, the objective of this study is not to fine-tune architectures
for maximal predictive accuracy, but to characterize the structural
conditions under which long-range dependencies remain statistically
detectable.

\begin{figure}[tp!]
  \centering

  \begin{subfigure}[b]{0.49\textwidth}
    \centering
    \includegraphics[width=\textwidth]{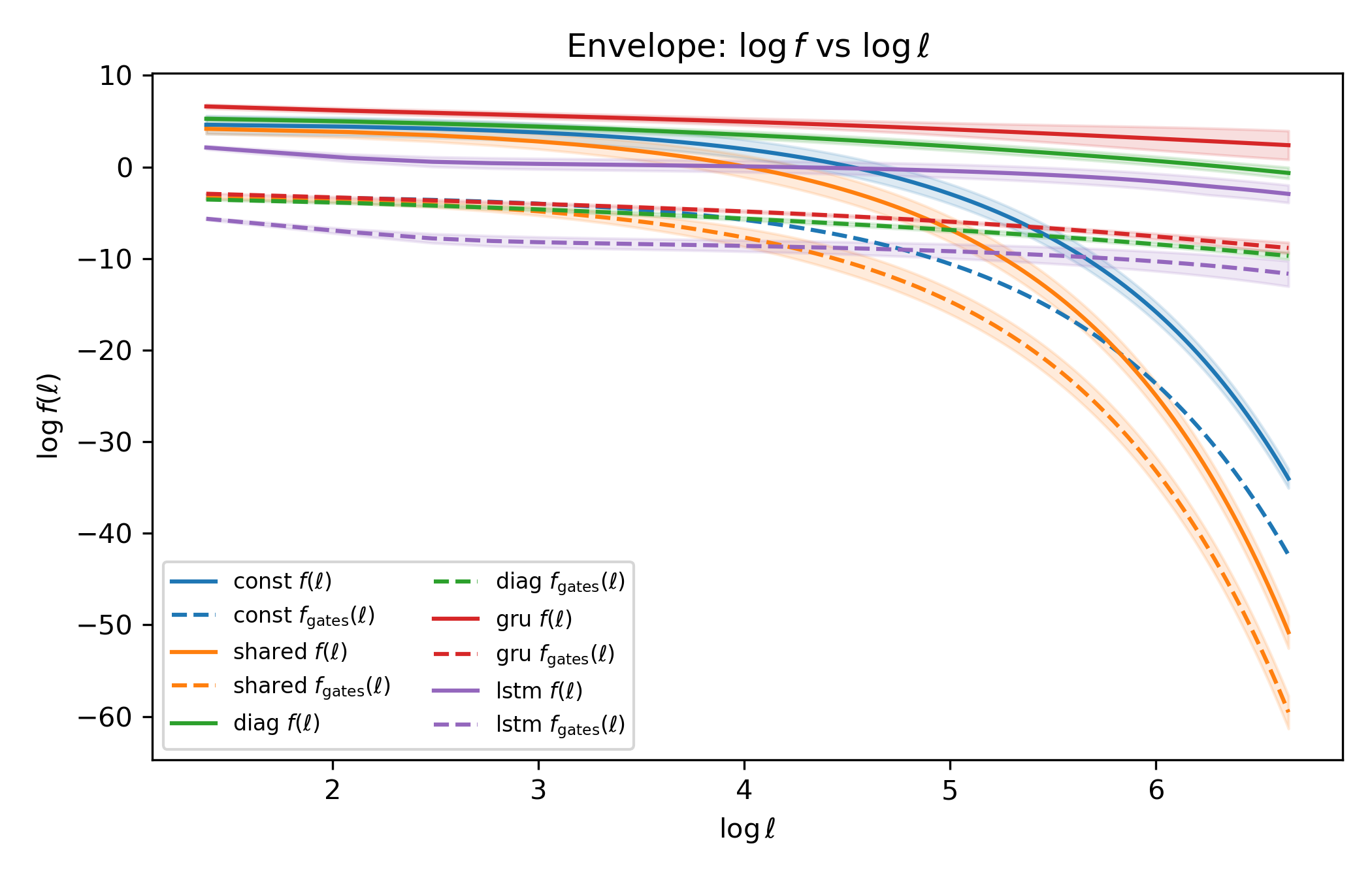}
    \caption{Full envelope $\hat f(\ell)$ vs gates-only contribution
$f_{\mathrm{gates}}(\ell)$, log--log scale.}
    \label{fig:decomp_loglog}
  \end{subfigure}
  \hfill
  \begin{subfigure}[b]{0.49\textwidth}
    \centering
    \includegraphics[width=\textwidth]{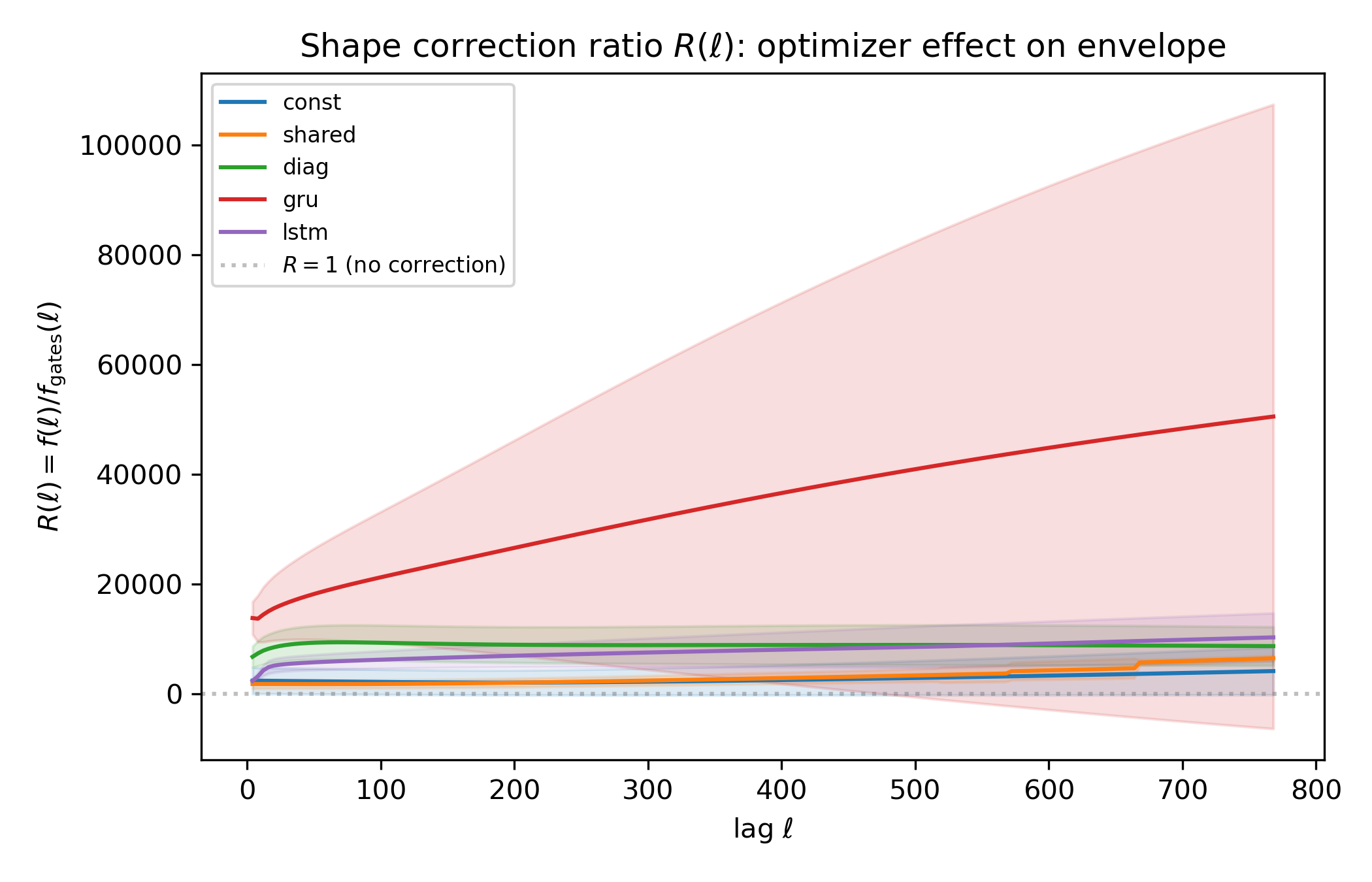}
    \caption{Shape-correction ratio
$R(\ell) = \hat f(\ell)/f_{\mathrm{gates}}(\ell)$.}
    \label{fig:decomp_ratio}
  \end{subfigure}

  \caption{Empirical decomposition of the envelope into the gates-only
contribution and the AdamW-induced amplification, per
Eq.~\eqref{eq:envelope_decomp}, averaged across training seeds (shaded
regions: cross-seed standard deviation).
\textbf{(a)} Solid curves show the full envelope $\hat f(\ell)$;
dashed curves show $f_{\mathrm{gates}}(\ell)$.
For DiagGate the two curves are roughly parallel on the log--log
scale, indicating a lag-independent multiplicative amplification.
For ConstGate and SharedGate the gap widens at large lags,
reflecting shape correction near the noise floor.
\textbf{(b)} The lag-dependent ratio $R(\ell)$ is approximately flat
for ConstGate, SharedGate, and DiagGate; rises moderately for LSTM
(from $\sim\!2.4\times 10^3$ at small lag to $\sim\!8.6\times 10^3$
at $\ell=512$); and grows dramatically with lag for GRU (from
$\sim\!1.4\times 10^4$ at small lag to $\sim\!5\times 10^4$ at
$\ell=768$), revealing that AdamW selectively amplifies long-lag
transport in GRU.}
  \label{fig:envelope_decomposition}
\end{figure}

The envelope decomposition introduced in
Eq.~\eqref{eq:envelope_decomp} provides additional empirical insight
into the optimizer's role in shaping $\hat f(\ell)$.
Figure~\ref{fig:envelope_decomposition} contrasts the gates-only
contribution $f_{\mathrm{gates}}(\ell)$ with the full envelope
$\hat f(\ell)$ (a) and reports the lag-dependent shape-correction
ratio $R(\ell) = \hat f(\ell)/f_{\mathrm{gates}}(\ell)$ (b).
For DiagGate, the AdamW preconditioning amplifies $f_{\mathrm{gates}}$ by a roughly lag-independent factor across the diagnostic range (visible as a near-parallel shift on the log--log scale), so the optimizer scales without substantially reshaping the gate-induced profile.
For ConstGate and SharedGate, $R(\ell)$ is also approximately flat
across the diagnostic range; the deviation between $\hat f(\ell)$ and
$f_{\mathrm{gates}}(\ell)$ visible on the log--log scale is
concentrated near the noise floor at large lags, where AdamW's
preconditioning provides a shape correction.
For LSTM the ratio rises moderately from $\sim\!2.4\times 10^3$ at
small lag to $\sim\!8.6\times 10^3$ at $\ell=512$, indicating partial
lag-dependence on top of a substantial multiplicative amplification.
For GRU the rise is dramatic, from $\sim\!1.4\times 10^4$ at small
lag to $\sim\!5\times 10^4$ at $\ell=768$, revealing that the
optimizer \emph{selectively amplifies} long-lag transport.
This decomposition makes explicit the architecture--optimizer
interaction underlying the regime classification: gating sets the
shape of $f_{\mathrm{gates}}$, while the projected adaptive base
rates $\Lambda^{(q)}_{r,\ell}$ provide either a near-uniform
multiplicative amplification (DiagGate, ConstGate, SharedGate),
a moderate lag-dependent amplification (LSTM), or a strongly
lag-dependent amplification (GRU) on top of it.

\begin{figure}[tp!]
    \centering
    \begin{subfigure}[b]{0.49\textwidth}
        \centering
        \includegraphics[width=\textwidth]{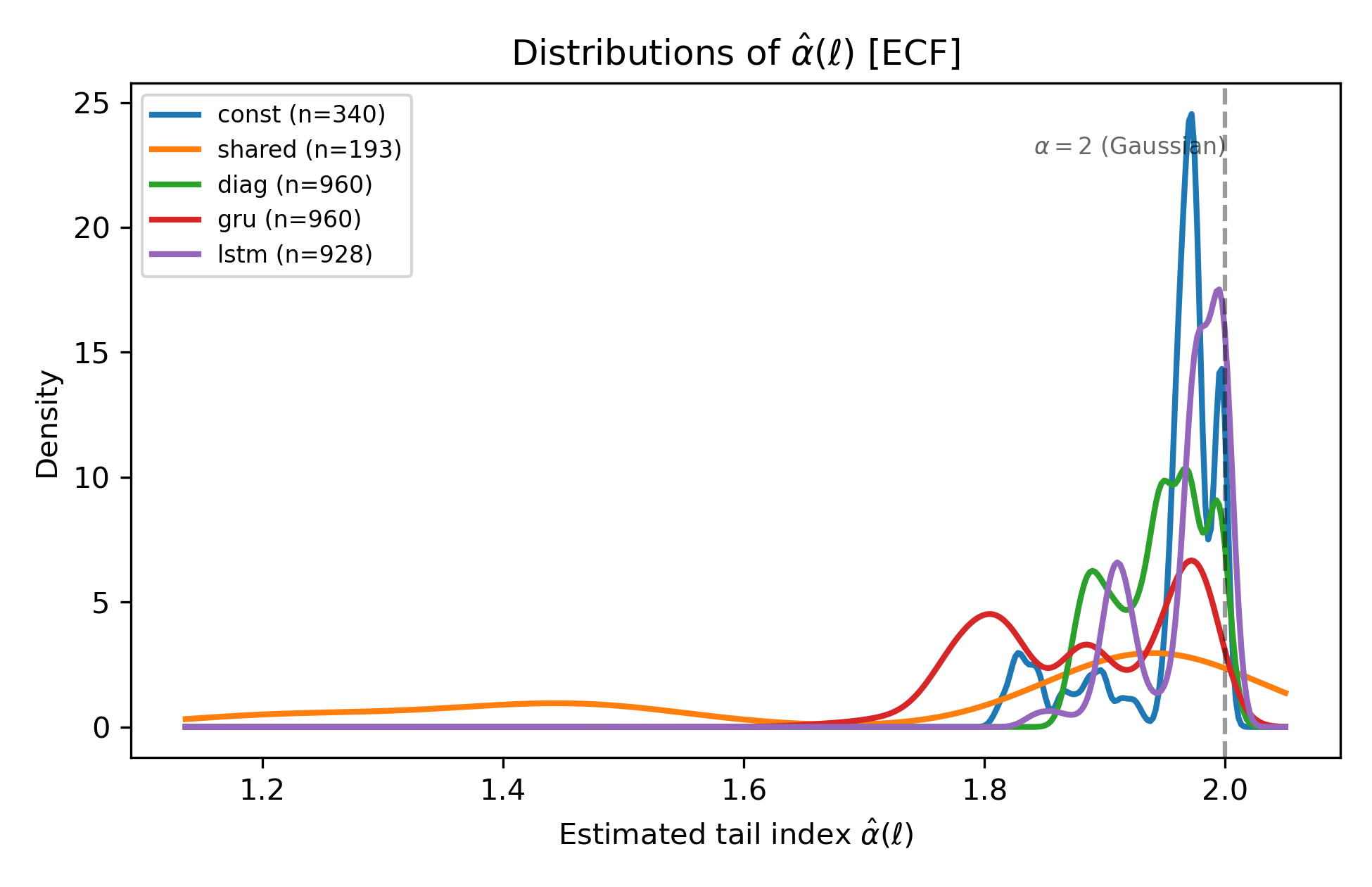}
        \caption{ECF estimator.}
        \label{fig:alpha_hat_ecf}
    \end{subfigure}
    \hfill
    \begin{subfigure}[b]{0.49\textwidth}
        \centering
        \includegraphics[width=\textwidth]{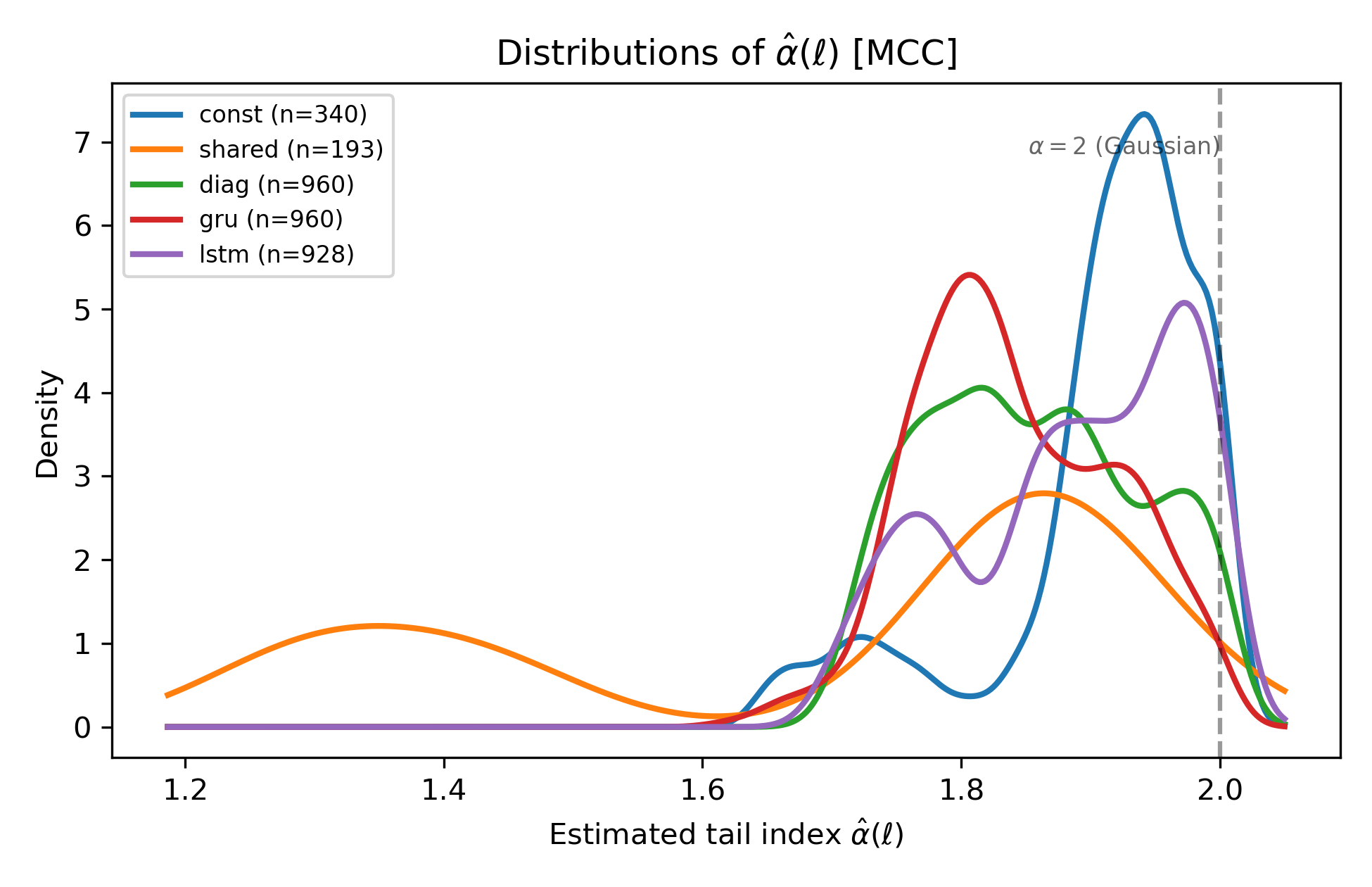}
        \caption{MCC estimator.}
        \label{fig:alpha_hat_mcc}
    \end{subfigure}
    \par\medskip
    \begin{subfigure}[b]{0.75\textwidth}
        \centering
        \includegraphics[width=\textwidth]{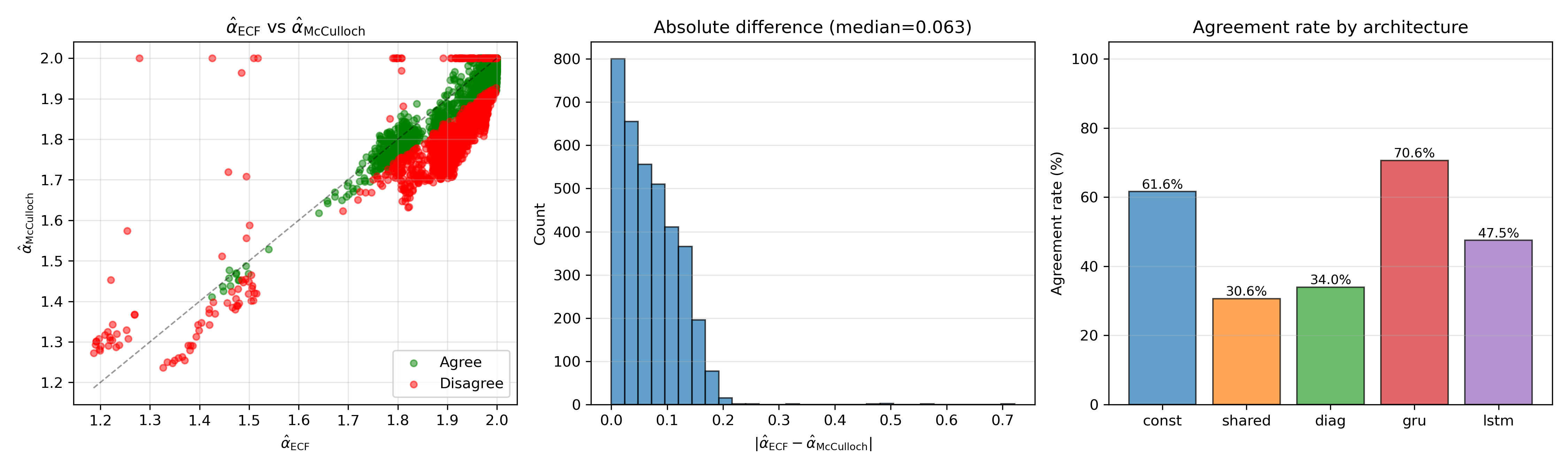}
        \caption{Agreement.}
        \label{fig:alpha_hat_agreement}
    \end{subfigure}
    \caption{
    Tail-index diagnostics for the sequence-level matched statistic.
    \textbf{(a)} ECF estimates of $\hat\alpha(\ell)$.
    \textbf{(b)} Bootstrapped MCC estimates.
    \textbf{(c)} Estimator agreement, defined per-$(\text{seed},\ell)$
    as the indicator that the ECF point estimate lies inside the MCC
    bootstrap $95\%$ CI; the panel reports the
    resulting per-architecture agreement rate together with the
    $\hat\alpha_{\mathrm{ECF}}$ vs $\hat\alpha_{\mathrm{MCC}}$ scatter.
    MCC generally shifts the gated-model estimates downward,
    indicating heavier tails and wider uncertainty near the Gaussian
    boundary, but it preserves the same qualitative ordering across
    architectures.
    }
    \label{fig:alpha_hat}
\end{figure}

\begin{figure}[tp!]
    \centering
    \begin{subfigure}[b]{0.49\textwidth}
        \centering
        \includegraphics[width=\textwidth]{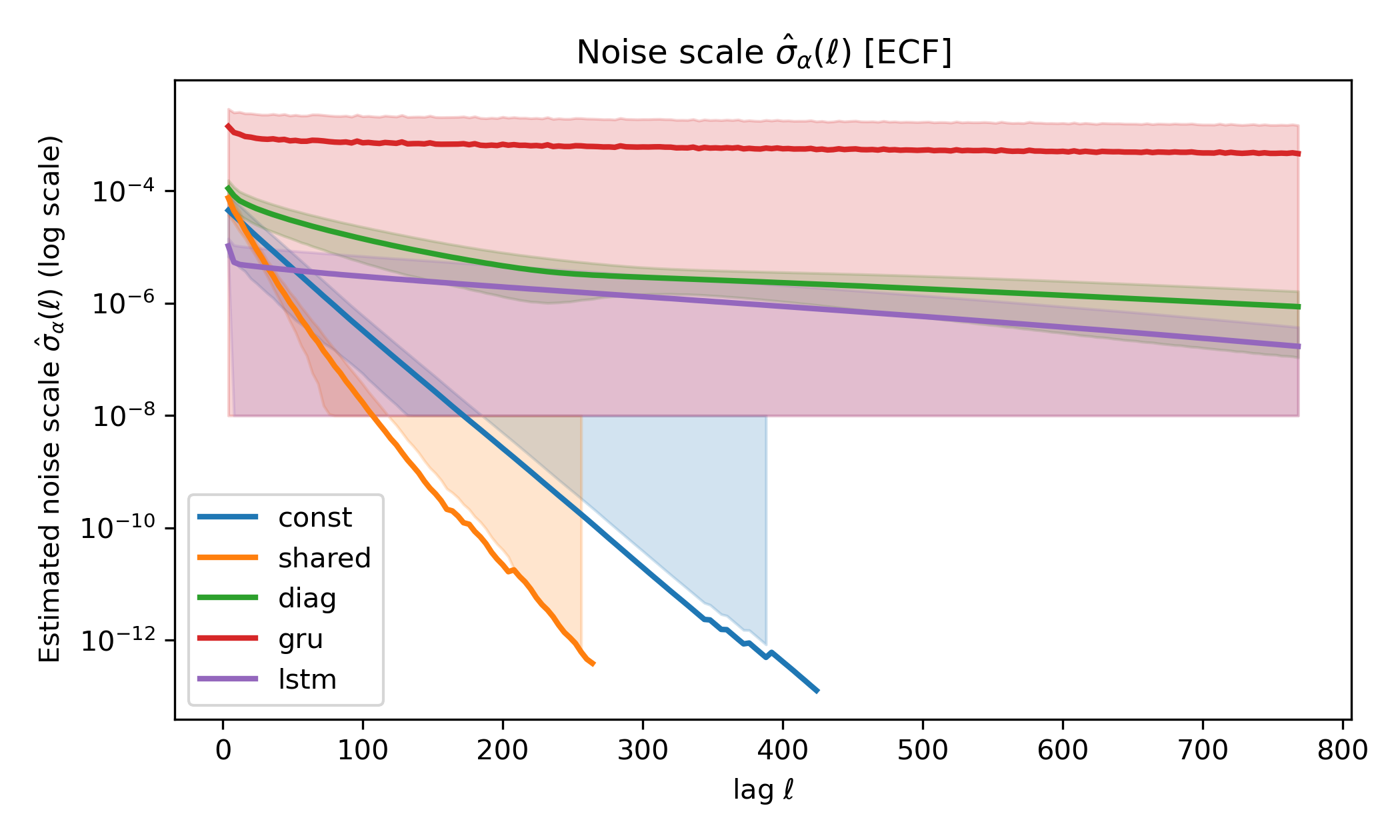}
        \caption{ECF estimator.}
        \label{fig:sigma_alpha_ecf}
    \end{subfigure}
    \hfill
    \begin{subfigure}[b]{0.49\textwidth}
        \centering
        \includegraphics[width=\textwidth]{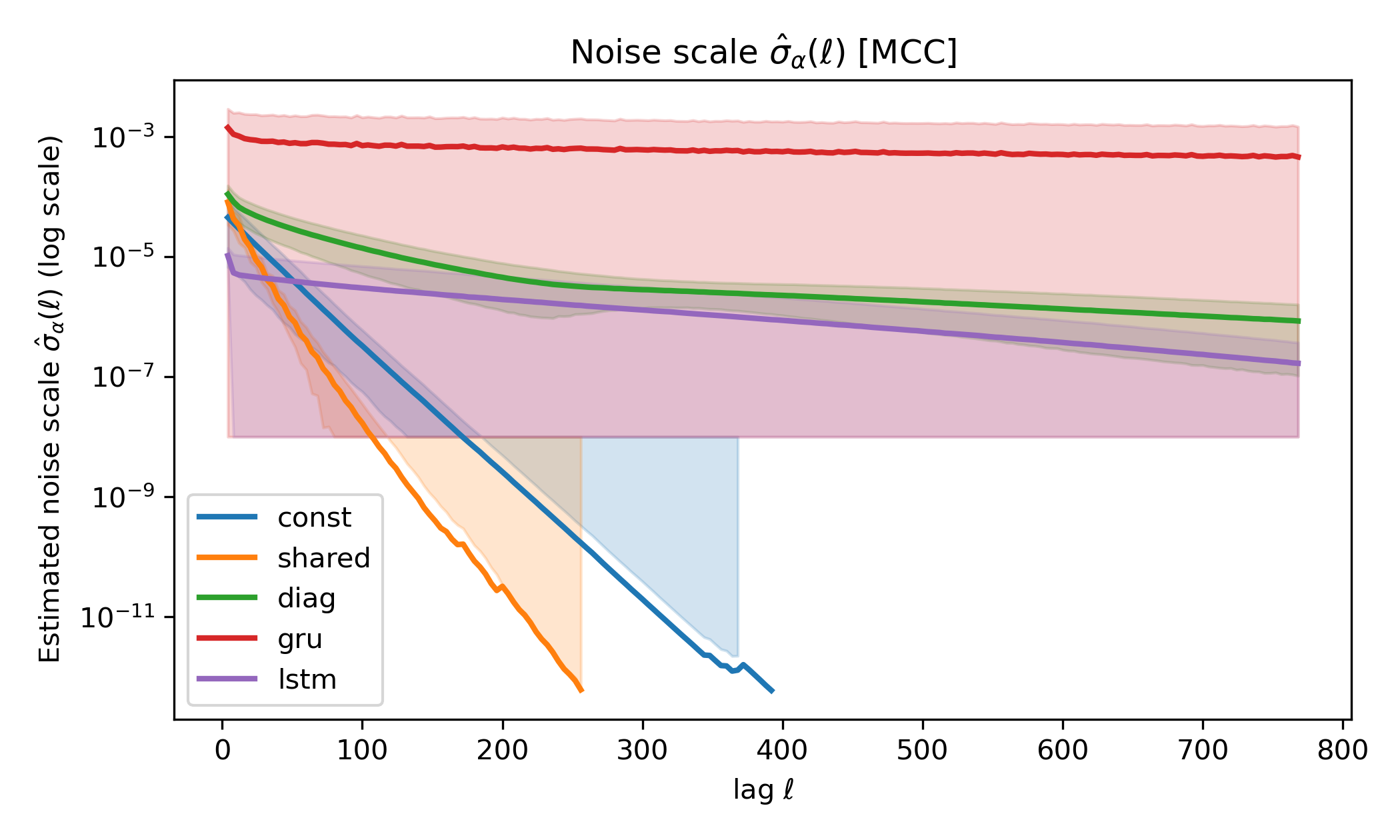}
        \caption{MCC estimator.}
        \label{fig:sigma_alpha_mcc}
    \end{subfigure}
    \caption{
    Estimated lag-dependent noise scale $\hat\sigma_\alpha(\ell)$ for the
    matched statistic under the two tail-index estimators.
    ConstGate and SharedGate rapidly reach the numerical floor.
    DiagGate retains a measurable but decaying scale, GRU maintains the
    largest persistent fluctuations, and LSTM has a much smaller scale
    despite reaching the full diagnostic learnability range.
    }
    \label{fig:sigma_alpha}
\end{figure}

\paragraph{Noise statistics and selection of scaling regimes.}
The learnability window depends not only on the envelope
$\hat f(\ell)$, but also on the fluctuations of the matched statistic
around its mean.
We therefore estimate, for each lag, both the stable tail index
$\hat\alpha(\ell)$ and the corresponding scale
$\hat\sigma_\alpha(\ell)$.
Figure~\ref{fig:alpha_hat} compares the lag-wise tail-index estimates
obtained from the ECF estimator and from the bootstrapped MCC quantile
estimator. The MCC estimator carries its own bootstrap (over $500$
resamples) and therefore returns a $95\%$ confidence interval (CI) $[\hat\alpha_{\mathrm{MCC}}^{\mathrm{lo}}(\ell),\,
\hat\alpha_{\mathrm{MCC}}^{\mathrm{hi}}(\ell)]$ for every lag.
Panel~\subref{fig:alpha_hat_agreement} reports the estimator-agreement
diagnostic, defined per-$(\text{seed},\ell)$ as the indicator that the
ECF point estimate falls inside the MCC bootstrap CI,
$\hat\alpha_{\mathrm{MCC}}^{\mathrm{lo}}(\ell)\le
\hat\alpha_{\mathrm{ECF}}(\ell)\le\hat\alpha_{\mathrm{MCC}}^{\mathrm{hi}}(\ell)$;
this is also reported as a per-architecture agreement rate.
A complementary, ECF-side bootstrap that attaches within-sample CIs to $\hat\alpha_{\mathrm{ECF}}(\ell)$ itself is introduced in the within-sample diagnostics paragraph below.
Figure~\ref{fig:sigma_alpha} reports the corresponding lag-dependent noise scales for both estimators.
The tail index controls the concentration exponent, whereas $\hat\sigma_\alpha(\ell)$ sets the lag-dependent noise floor in the detectability threshold.

The ECF estimates show no lower-endpoint pinning.
ConstGate remains close to the Gaussian boundary on the lags where the
matched statistic is estimable, while SharedGate exhibits stronger
floor effects and seed-to-seed variation in the tail estimates.
Among the gated models, DiagGate shows a mild departure from the
Gaussian boundary, GRU is the most consistently heavy-tailed
($\hat\alpha_{\mathrm{ECF}}$ ranges down to about $1.64$), and LSTM
lies close to the boundary for some seeds while still showing
non-Gaussian tails for others.
The MCC estimator preserves the same qualitative ordering, but
is more conservative near the Gaussian boundary: for DiagGate, GRU,
and LSTM it lowers the average tail-index estimate by roughly
$0.09$, $0.04$, and $0.08$, respectively.
Thus the robustness check does not change the regime classification,
but it indicates that ECF estimates very close to $\alpha=2$ should not
be read as evidence of exactly Gaussian fluctuations.

We strengthen this picture with two within-sample diagnostics applied
directly to the ECF estimator, complementing the MCC-derived agreement
diagnostic discussed above
(see Appendices~\ref{app:bootstrap_ci} and~\ref{app:projection_alpha_diagnostic}
for full details).
First, we attach a $95\%$ bootstrap CI to each per-$(\text{seed},\ell)$
ECF estimate by resampling, with replacement, the projection-averaged
sequence-level scalars over the diagnostic sequences.
The CIs are tight: the median CI width across
$(\text{seed},\ell)$ pairs is between $0.017$ (LSTM, narrowest) and
$0.04$ (GRU, widest); for every architecture, at least one
$(\text{seed},\ell)$ pair has a CI strictly below $\alpha=2$, so the
heavy-tail departures from Gaussian are statistically resolved.
The clearest moderate-heavy-tail signals are GRU's minimum
$\hat\alpha\approx 1.64$ ($95\%$ CI $[1.61,\,1.67]$) and LSTM's
minimum $\hat\alpha\approx 1.77$ ($95\%$ CI $[1.75,\,1.80]$), both
well separated from the Gaussian boundary at the resampling level.
Second, we estimate $\hat\alpha$ separately for each of the $K=50$
projection directions used to define the matched statistic; the
resulting per-projection distributions are tightly concentrated for
each $(\text{seed},\ell)$ (interquartile range
$0.017$ -- $0.058$ across architectures), which is consistent with
approximately isotropic directional behavior.
For the gated models, the projection-averaged tail-index estimates are
as heavy as, or heavier than, the typical per-direction estimates,
which is consistent with shared within-sequence dependence across
projection directions and confirms that $K$-projection averaging
provides variance reduction without forcing the aggregate toward the
Gaussian boundary.
The projection-averaged matched statistic therefore retains the joint
tail weight required by the framework.

The scale estimates separate the architectures even more clearly.
ConstGate and SharedGate cross the numerical floor rapidly, consistent
with their fast exponential envelopes.
DiagGate retains a measurable but decaying noise scale across the full
diagnostic range.
GRU has by far the largest and most persistent matched-statistic
fluctuations: under ECF, the cross-seed mean scale decreases only from
about $1.4\times 10^{-3}$ at $\ell=4$ to
$4.6\times 10^{-4}$ at $\ell=768$.
LSTM shows the opposite combination: its tail index remains close to
the Gaussian boundary for much of the lag range, and its scale is much
smaller, decreasing from about $1.0\times 10^{-5}$ at $\ell=4$ to
$1.7\times 10^{-7}$ at $\ell=768$.
This small noise scale helps explain why the LSTM learnability window
reported below reaches the full diagnostic range despite a weaker
envelope tail than GRU.

These statistics clarify how scaling regimes are selected in finite data.
The architecture--optimizer pair defines the accessible envelope geometry, while the matched-statistic tail index and scale determine how costly it is to estimate the corresponding long-lag signal.
Slow envelope decay is beneficial only when the signal remains above the noise floor; conversely, a smaller noise scale can make an intermediate envelope empirically learnable over the explored range.


\begin{figure}[tp!]
    \centering

    \begin{subfigure}[b]{0.49\textwidth}
        \centering
        \includegraphics[width=\textwidth]{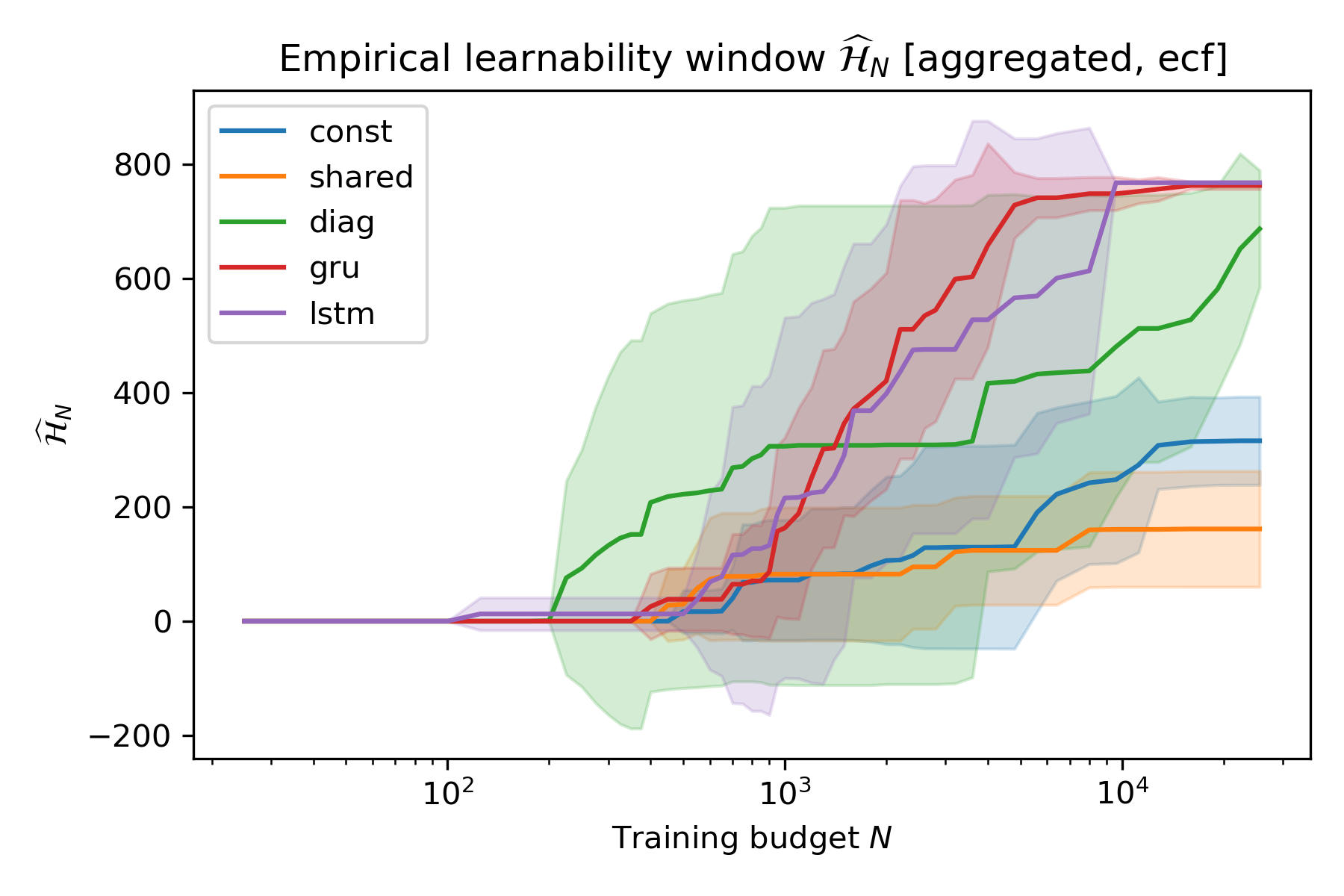}
        \caption{ECF estimator.}
        \label{fig:HN_vs_N_ecf}
    \end{subfigure}
    \hfill
    \begin{subfigure}[b]{0.49\textwidth}
        \centering
        \includegraphics[width=\textwidth]{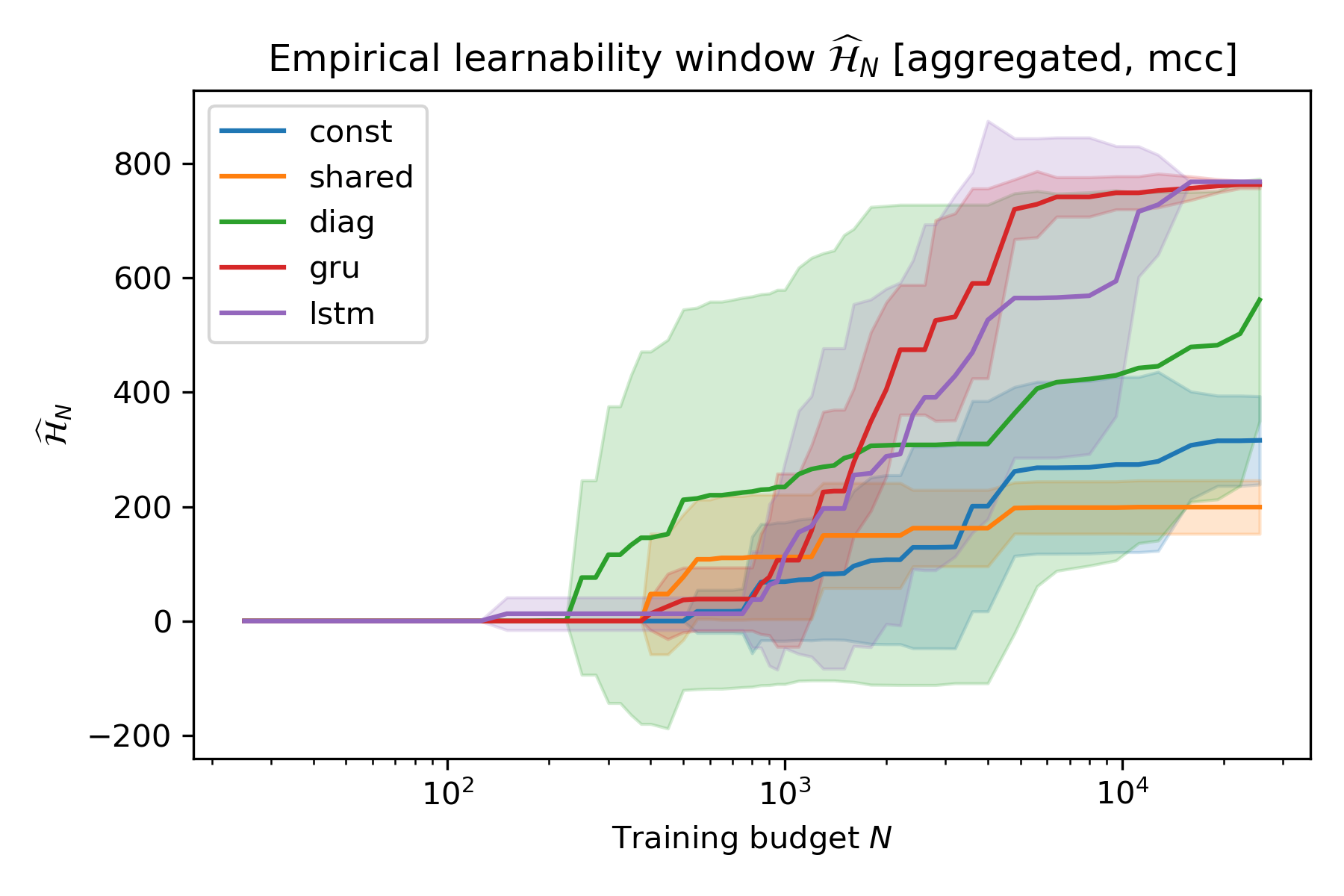}
        \caption{MCC estimator.}
        \label{fig:HN_vs_N_mcc}
    \end{subfigure}

    \caption{Empirical learnability windows $\widehat{\mathcal{H}}_N$
for ConstGate, SharedGate, DiagGate, GRU, and LSTM, as a function of
the number of independent training sequences $N$.
\textbf{(a)} Tail index $\alpha$ estimated with the Koutrouvelis ECF
regression estimator.
\textbf{(b)} Tail index $\alpha$ estimated with the bootstrapped
MCC quantile estimator.
Solid curves give the cross-seed mean over the training seeds,
and shaded regions denote the variability across seeds.
ConstGate and SharedGate (blue and orange) exhibit short and slowly
growing horizons, reflecting the rapid exponential decay of their
envelopes $\hat f(\ell)$.
DiagGate (green) shows a data-dependent expansion: once $N$ exceeds a
few hundred sequences, the window progressively reaches longer lags
with strong cross-seed heterogeneity.
GRU (red) and LSTM (purple) exhibit a delayed-transition behavior:
below a critical sample size the horizon is negligible, whereas above
this threshold the window rapidly extends to the maximal diagnostic
lag ($\widehat{\mathcal H}_N=768$).
The two estimators agree across architectures, with the most visible
discrepancy on the lower edge of the SharedGate shaded band, where
MCC yields a small nonzero window in cases where ECF remains at
the floor.}
    \label{fig:HN_vs_N}
\end{figure}

\paragraph{Empirical learnability windows.}
Combining the measured envelopes $\hat f(\ell)$ with the estimated
heavy-tailed noise parameters above, we evaluate the empirical
detectability condition for varying training budgets $N$ and construct
the corresponding learnability window $\widehat{\mathcal H}_N$ defined
in Eq.~\eqref{eq:H_N_empirical}.
Figure~\ref{fig:HN_vs_N} reports $\widehat{\mathcal H}_N$ as a function
of the number of independent training sequences, computed under both
tail-index estimators introduced above: the Koutrouvelis ECF estimator
(a) and the bootstrapped MCC quantile estimator (b).
Solid curves give the cross-seed mean over the training runs,
and shaded regions denote the cross-seed variability.

ConstGate and SharedGate exhibit short and slowly growing learnability
horizons. Under ECF, ConstGate reaches a moderate ceiling
($\widehat{\mathcal H}_N \approx 316$ at $N=25600$, with substantial
cross-seed spread), while SharedGate stays lower
($\widehat{\mathcal H}_N \approx 162$).
The MCC panel yields qualitatively identical curves and modestly
raises the lower edge of the SharedGate band, consistent with the
floor-sensitive nature of the ECF estimates in this rapidly decaying
regime.
In both architectures the windows remain effectively zero until
$N \gtrsim 800$, after which growth proceeds slowly with $N$.
This behavior is consistent with the exponential envelope regime:
when $\hat f(\ell)$ decays exponentially, the detectability threshold
grows rapidly with lag, so most increments in $N$ accommodate only
marginally longer detectable horizons.

DiagGate displays qualitatively different behavior.
For small $N$ the learnability window is negligible.
Once the budget exceeds a few hundred sequences,
$\widehat{\mathcal H}_N$ expands progressively toward the maximal
diagnostic lag, reaching $\widehat{\mathcal H}_N \approx 687$ at
$N=25600$.
The expansion is strongly heterogeneous across seeds: some trajectories
reach the full diagnostic range at moderate sample sizes, while others
open more slowly and remain below the full range within the explored
budget, producing the wide shaded band visible in both panels.
This data-dependent expansion is consistent with the observed
approximately polynomial envelope regime: slow attenuation allows
additional lags to cross the detectability threshold as $N$ increases.

GRU and LSTM exhibit an even more pronounced expansion of the window.
Both architectures display a delayed-transition pattern: below a
critical sample size, the window is essentially zero, whereas above this threshold the window rapidly extends to the maximal diagnostic lag.
At $N=25600$, LSTM consistently reaches the full diagnostic range
($\widehat{\mathcal H}_N = 768$), and GRU reaches a near-identical mean
of $\widehat{\mathcal H}_N \approx 763$ with similarly tight variability.
Both estimators agree on these values.
This behavior reflects the slow envelope decay over intermediate lags
for GRU, and the combination of an intermediate LSTM envelope with a
smaller matched-statistic noise scale: once the training budget exceeds
the noise-dominated regime, a large portion of the temporal range
becomes detectable within the explored learnability window.

The two tail-index estimators yield quantitatively similar conclusions across the architectures, with the most visible discrepancy in the SharedGate shaded band noted above.
This robustness check supports the qualitative pattern displayed in both panels.

These results are consistent with the structural prediction of the theory.
Exponential envelope decay leads to early saturation of the learnability window at a budget-limited ceiling, whereas slow, approximately power-law attenuation enables systematic expansion of the horizon with increasing data.
The decay class fixes this qualitative behavior, but the exact window level also depends on the architecture-specific constants entering the detectability bound (the alignment $|\overline m_\mu(\ell)|$, the noise scale $\sigma_\alpha(\ell)$, and the tail index), so comparable envelopes can yield different windows.
These constants affect only the bound prefactors, not the decay-class classification or the scaling exponents; sharpening them would tighten the inevitable gap between the predicted trends and the experimental results.

\begin{figure}[tp!]
    \centering
    \includegraphics[width=0.85\linewidth]{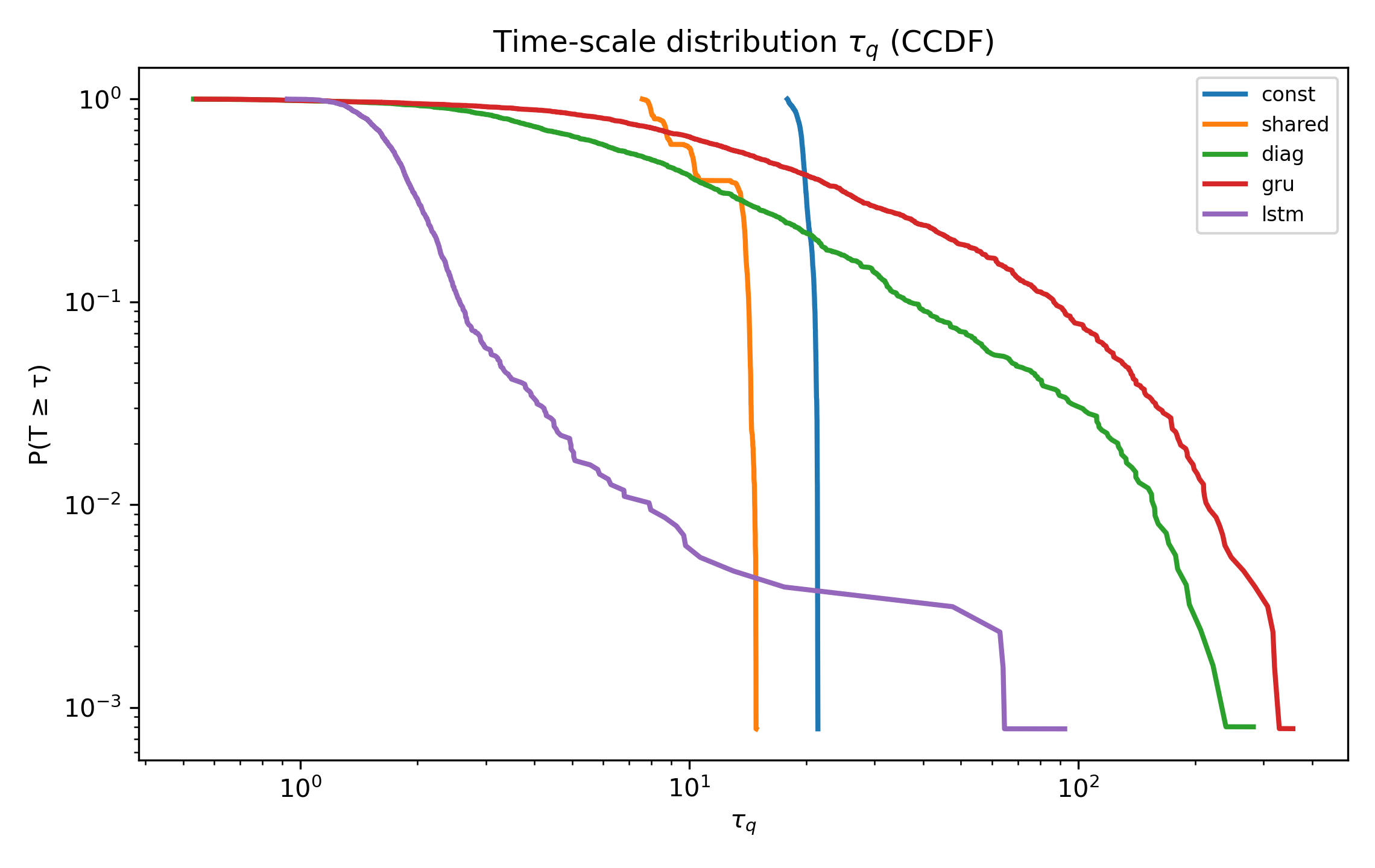}
    \caption{
    Complementary cumulative distribution functions (CCDFs) of the neuronwise
    effective time scales $\tau_q$, averaged across AdamW training runs.
    ConstGate and SharedGate show narrowly concentrated spectra.
    DiagGate and GRU exhibit broad heterogeneous spectra with slowly decaying
    intermediate tails and finite-size cut-offs.
    LSTM displays a sparse-tail structure: a narrow fast bulk together with a
    small long-time-scale tail.
    }
    \label{fig:tau_hist}
\end{figure}

\paragraph{Time-scale spectra.}
To connect the envelope geometry back to gating, we inspect the distribution of
neuronwise time scales $\{\tau_q\}_q$ inferred from the effective learning rates.
Rather than relying solely on density estimates, we summarize these spectra via
the CCDF, $\mathbb{P}(\tau_q \ge \tau)$, which directly quantifies the fraction of units
operating on time scales longer than~$\tau$.
Figure~\ref{fig:tau_hist} reports the empirical CCDFs for all architectures on
log--log axes, making both bulk concentration and sparse slow tails
visible.

ConstGate exhibits an almost degenerate spectrum, with all units clustered
near a single characteristic time scale.
SharedGate is similarly narrow, with only mild unit-to-unit variation.
These synchronized spectra match the rapid exponential envelopes reported
above: although the characteristic time scale is finite and nonzero, the lack
of a broad slow tail prevents sustained long-lag transport.

DiagGate and GRU are qualitatively different.
Both develop broad spectra with substantial mass over intermediate time scales
and a slowly decaying tail before the finite-dimensional cut-off.
This heterogeneity is most pronounced for GRU, whose CCDF remains elevated over
the widest range of $\tau$ values.
The broad intermediate tail is consistent with the approximately
power-law-like envelope attenuation observed in Fig.~\ref{fig:envelope_scalings}
and with the much larger learnability windows in Fig.~\ref{fig:HN_vs_N}.

LSTM occupies a distinct sparse-tail regime.
In the present experimental setup, most LSTM units concentrate in a fast
time-scale bulk near $\tau \approx 2$, while a small number of units populate a
sparse long-time-scale tail.
Thus the LSTM spectrum is not intermediate in bulk width: its bulk is narrow
and fast, but it is supplemented by a sparse slow tail.
This shape is qualitatively distinct from both the synchronized spectra of
ConstGate and SharedGate and the broadly heterogeneous spectra of DiagGate and
GRU.
It is nevertheless compatible with the envelope behaviour reported above:
LSTM lies closer to the exponential boundary than GRU, while the sparse tail
retains enough long-lag signal to reach the full diagnostic learnability range
at large sample sizes.

\section{Discussion and future directions}
\label{sec:discussion}

This work develops a theoretical framework linking gating dynamics, optimizer adaptation, heavy-tailed statistics, and the finite-horizon learnability of recurrent neural networks.
The central result is that the decay law of the envelope $f(\ell)$, together with the concentration exponent $\kappa_\alpha$, governs both sample-complexity requirements for detecting usable signals in gradient information and the growth of the corresponding learnability window.

Our empirical results support this theoretical picture while clarifying the roles of architectural inductive bias and optimizer adaptation.
Architectures with constrained gating tend to induce narrow and
nearly synchronized time-scale spectra, leading to fast exponential
envelope decay and learnability windows that expand slowly and remain
limited to a low ceiling relative to the diagnostic range.
Architectures capable of neuronwise or multi-gate modulation can admit
heterogeneous mixtures of time scales and realize slower attenuation
regimes, within which the learnability window expands systematically with available data.
Adaptive optimizers further shape this differentiation: in the AdamW
experiments, the projected adaptive base rates act as multiplicative
shape corrections on top of the gate-induced profile, ranging from
near-uniform amplification in some architectures to strong lag-dependent amplification in GRU.
This optimizer-induced shape correction is absent under SGD.
The decisive factor is therefore not nominal architectural complexity, but the dispersion of effective time scales that emerges from the coupled state-space and parameter-space dynamics.

\paragraph{Heavy-tailed fluctuations as a viability constraint.}
The scaling laws make precise the statistical cost of each decay regime: exponential envelope decay entails exponential sample complexity, while power-law decay entails only polynomial sample complexity.
For any finite training budget, exponential forgetting rapidly exhausts the available statistical resources, whereas polynomial forgetting keeps long-range dependencies within reach considering the same data budget.
Heavy-tailed fluctuations exacerbate this disparity by increasing~$\kappa_\alpha$ and slowing statistical concentration.
Therefore, heavy-tailed fluctuations act as a viability constraint, disfavoring rapid forgetting, and effectively selecting, among the dynamical regimes accessible to an architecture--optimizer pair, those that preserve signal over longer temporal scales.

We conjecture that this viability constraint induces a form of constrained self-organization: during training, recurrent networks tend to evolve toward regimes of slower envelope decay in order to maintain learnability with finite data.
This does not require explicit regularization; it follows from the fact that configurations with exponential forgetting become statistically unlearnable at long lags, creating implicit selection pressure toward slower envelope decay.

The taxonomy admits a third, still more favorable regime: a logarithmic envelope decay would yield near-exponential growth of the window with data, the least demanding regime in terms of sample complexity requirements.
We do not observe it in our experiments, which realize only the exponential and power-law regimes.
This does not establish that it is unattainable; whether deep learning models can sustain a logarithmic regime in realistic settings remains an open question for future research.

\paragraph{Broader implications.}
Temporal learnability is limited less by the size of the training set than by the envelope decay class, which the architecture--optimizer pair biases through the dispersion of the effective time scales distribution.
Under heavy-tailed fluctuations, this distinction becomes more evident: adding data yields diminishing returns within a fixed decay class, whereas changing the decay class can shift the system to significantly broader horizons outright.

More broadly, the framework is not tied to a specific architecture or optimizer.
We focus on gated recurrent networks because their dynamics admit a transparent interpretation in terms of tunable time scales, but similar limitations are expected in any learning system trained with SGD-like optimization, where signals propagate through long chains of Jacobians, including deep or sequence models where temporal depth can be traded for layer depth~\cite{sieber2024understanding}.

Finally, the present framework establishes which scaling regimes are statistically viable under finite data and heavy-tailed fluctuations, but does not explain how training dynamics navigate the space of accessible regimes.
A natural next step is the development of a stochastic model describing the evolution of the effective learning rates during training: a theory of how time-scale spectra emerge, broaden, or collapse as the architecture and the optimizer co-evolve.
Such a dynamical theory would complement the learnability analysis presented here with an account of long-range memory formation, addressing the open question of why and how some trained networks self-organize into broad time-scale mixtures, while others collapse to synchronized dynamics.

\clearpage
\bibliographystyle{abbrvnat}
\bibliography{bibliography.bib}

\clearpage
\appendix

\section{Matrix product expansion via the Fr\'echet derivative formulation}
\label{app:first_order_expansion}

This section summarizes the first-order expansion of a product of matrices with structured perturbations introduced in \cite{livi2025timescale}.

\begin{definition}[Fr\'echet differentiability {\cite{higham2008functions}}]
Let $f : \mathbb{C}^{n \times n} \to \mathbb{C}^{n \times n}$.
We say that $f$ is \emph{Fr\'echet differentiable} at $A \in \mathbb{C}^{n\times n}$ if there exists a bounded linear mapping
$L_f(A,\cdot)$ such that
\begin{equation}
\lim_{\|E\| \to 0} \frac{\|\,f(A+E) - f(A) - L_f(A,E)\,\|}{\|E\|} = 0.
\label{eq:frechet-definition}
\end{equation}
\end{definition}
If $f$ is Fr\'echet differentiable at $A$,
\begin{equation}
f(A+E) \;=\; f(A) \;+\; L_f(A, E) \;+\; o(\|E\|).
\label{eq:first-order-taylor}
\end{equation}

If $g$ and $h$ are Fr\'echet differentiable at $A$ and $f(X)=g(X)h(X)$, then
\begin{equation}
L_{gh}(A,E) \;=\; L_g(A,E)\,h(A) \;+\; g(A)\,L_h(A,E).
\label{eq:frechet-product-rule}
\end{equation}

Consider the perturbed matrix product
\begin{equation}
F(\varepsilon) = \prod_{j=1}^{n} \big(A_j + \varepsilon B_j\big),
\label{eq:product-def}
\end{equation}
where $A_j,B_j\in\mathbb{C}^{d\times d}$.
By recursive application of \eqref{eq:frechet-product-rule} one obtains
\begin{equation}
\label{eq:first-derivative}
L_{F_n}(0,E) \;=\; \sum_{i=1}^n 
\left( \prod_{j=1}^{i-1} A_j \right) \,
B_i \,
\left( \prod_{j=i+1}^n A_j \right),
\end{equation}
and the first-order expansion
\begin{equation}
\label{eq:first-order-expansion}
F(\varepsilon) \;=\;
\left( \prod_{j=1}^n A_j \right)
\;+\;
\varepsilon \sum_{m=1}^n
\left( \prod_{j=1}^{m-1} A_j \right) \, B_m \,
\left( \prod_{j=m+1}^n A_j \right)
\;+\; O(\varepsilon^2).
\end{equation}

\paragraph{Ordering in BPTT.}
In BPTT, the iterated Jacobian $\prod_{j=k+1}^{t} J_j$ is evaluated in
chain-rule order, with the latest factor on the left and the earliest
on the right. Identifying the abstract product index $j=1,\dots,n$
with the BPTT time index $j=k+1,\dots,t$ as in~\cite{livi2025timescale},
the first-order expansion of Eq.~\eqref{eq:first-order-expansion}
applies verbatim: the algebraic structure of the perturbation
insertions is unchanged, with each $B_m$ appearing between the factors
associated with earlier and later time steps.

\paragraph{Scope and limitations.}
The diagonal first-order expansion defines the per-neuron effective learning rates used throughout the main text.
The numerical accuracy of this expansion as an approximation of the full Jacobian product, including truncation error analysis and simulations across architectures, is validated in the original paper~\cite{livi2025timescale}, which also discusses its limitations.  Appendix~\ref{app:envelope_validation} provides complementary evidence that the resulting envelope preserves the decay profile required by the learnability theory.

\section{RNNs: One-step Jacobians and effective learning rates}
\label{app:RNNs_jacobians_and_elr}

This appendix provides the full architecture-specific derivations of
one-step Jacobians and per-neuron effective learning rates for the
recurrent models considered in this paper.
The derivations follow from the first-order expansion of Jacobian
products summarized in Appendix~\ref{app:first_order_expansion}.
For each architecture, we decompose the one-step Jacobian into a
zeroth-order transport operator and a recurrent mixing correction,
apply the first-order product rule, and extract the diagonal
components that define the neuron-wise transport factors.

The effective learning rates are presented here for plain SGD with a
fixed global step size~$\mu$, following the original formulation
in~\cite{livi2025timescale}.
This keeps the derivations focused on the recurrent transport
mechanisms that generate the neuron-wise factors
$\Gamma^{(q)}_{t,\ell}$.
The generalized formulation used in the main text, which replaces the
global step size $\mu$ by the neuron-specific base rate
$\Lambda^{(q)}_{r,\ell}$, is introduced in
Section~\ref{sec:generalized_elr}.

Section~\ref{app:notation_conventions_elr} fixes notation.
Sections~\ref{app:lstm_effective} and~\ref{app:gru_effective} derive
the effective learning rates for the LSTM and GRU, respectively.
Section~\ref{app:diag_gated_rnns} treats the baseline gated RNN
architectures (DiagGate, SharedGate, ConstGate).
Section~\ref{app:multirate_discussion} concludes with a discussion of
how these effective learning rates reveal an implicit multi-rate
optimizer structure induced by gating.

\subsection{Notation and conventions}
\label{app:notation_conventions_elr}

Let $x_t\in\mathbb{R}^D$ be the input at step $t$,
$h_t\in\mathbb{R}^H$ the hidden state,
and, for the LSTM, $c_t\in\mathbb{R}^H$ the cell state.
For any $v\in\mathbb{R}^H$, let
$D(v)=\mathrm{diag}(v)\in\mathbb{R}^{H\times H}$.
Hadamard (elementwise) product is denoted by $\odot$.
The logistic and hyperbolic tangent functions are applied
elementwise:
$\sigma(\cdot)$ and $\tanh(\cdot)$.
We use the diagonal slope matrices
\begin{equation}
S^{\sigma}(u)=D(\sigma'(u)),
\qquad
S^{\tanh}(u)=D\!\bigl(1-\tanh^2(u)\bigr).
\end{equation}
When convenient, $D(h_{t-1})$ is abbreviated by $D_{h,t-1}$.
Weight matrices have shapes
$W_\bullet\in\mathbb{R}^{H\times D}$,
$U_\bullet\in\mathbb{R}^{H\times H}$,
and $b_\bullet\in\mathbb{R}^{H}$.

Throughout, one-step Jacobians are derivatives with respect to the previous recurrent state.
For the LSTM we stack
$s_t=[h_t;c_t]\in\mathbb{R}^{2H}$ and write
\begin{equation}
J_t=\frac{\partial s_t}{\partial s_{t-1}}
\in\mathbb{R}^{2H\times 2H},
\end{equation}
whereas for the GRU and the baseline gated RNNs we write
\begin{equation}
J_t=\frac{\partial h_t}{\partial h_{t-1}}
\in\mathbb{R}^{H\times H}.
\end{equation}

For a square matrix $A\in\mathbb{R}^{n\times n}$,
$\mathrm{diagvec}(A)\in\mathbb{R}^n$
denotes the vector formed by the diagonal of $A$.
We use interval-product notation throughout: for any matrix-valued sequence $\{A_j\}$,
\begin{equation}
A_{a:b}=\prod_{j=b+1}^{a} A_j,
\qquad
A_{b:b}=I.
\end{equation}
In the architecture-specific derivations below, matrix products are written using absolute time endpoints, consistently with Eq.~\eqref{eq:bptt_one_step}.
Thus,
\begin{equation}
\mathcal{M}_{t,\ell}
=
\prod_{j=\ell+1}^{t}J_j,
\end{equation}
where the second index $\ell$ denotes the absolute source time and the product transports from time $\ell$ to time~$t$. Under the lag convention used in the main text, the same transport at lag $d=t-\ell$ is denoted by $\mathcal{M}_{t,d}$. The rightmost factor corresponds to the earliest time step, consistently with the ordering convention stated in Appendix~\ref{app:first_order_expansion}.

The transport factors and effective learning rates derived below follow the same re-indexing: their second index denotes an absolute source time in this appendix and a temporal lag in the main text.

In the LSTM case, when $\mathcal{M}_{t,\ell}$ is written in
$2H\times 2H$ block form for $s=[h;c]$, the notation
$[\,\cdot\,]_{h,c}\in\mathbb{R}^{H\times H}$ refers to the
top-right block mapping $c_\ell\mapsto h_t$.

\subsection{LSTM: one-step Jacobian and effective learning rates}
\label{app:lstm_effective}

\paragraph{Dynamics.}
An LSTM maintains $(h_t,c_t)$ and computes
\begin{align}
a^i_t &= W_i x_t + U_i h_{t-1} + b_i,
& i_t &= \sigma(a^i_t),
& & \text{(input gate)}
\\[-1pt]
a^f_t &= W_f x_t + U_f h_{t-1} + b_f,
& f_t &= \sigma(a^f_t),
& & \text{(forget/retention)}
\\[-1pt]
a^o_t &= W_o x_t + U_o h_{t-1} + b_o,
& o_t &= \sigma(a^o_t),
& & \text{(output/expression)}
\\[-1pt]
a^g_t &= W_g x_t + U_g h_{t-1} + b_g,
& g_t &= \tanh(a^g_t),
& & \text{(cell candidate)}
\\[4pt]
c_t &= f_t \odot c_{t-1} + i_t \odot g_t,
\qquad
h_t \;=\; o_t \odot \tanh(c_t).
\end{align}
Define the diagonal gate matrices and slope matrices
\[
F_t=D(f_t),\quad I_t=D(i_t),\quad O_t=D(o_t),\quad G_t=D(g_t),
\]
\[
S^i_t=S^{\sigma}(a^i_t),\quad
S^f_t=S^{\sigma}(a^f_t),\quad
S^o_t=S^{\sigma}(a^o_t),\quad
S^g_t=S^{\tanh}(a^g_t),
\]
and the cell-expression factors
\[
S_t=S^{\tanh}(c_t)=D\!\bigl(1-\tanh^2(c_t)\bigr),\qquad
H_t=D\!\bigl(\tanh(c_t)\bigr),\qquad
E_t=D\!\bigl(o_t\odot(1-\tanh^2(c_t))\bigr)=O_tS_t.
\]

\paragraph{One-step Jacobian.}
With $s_t=[h_t;c_t]$, the Jacobian blocks are
\begin{align}
\frac{\partial c_t}{\partial c_{t-1}}
&=F_t,
&
\frac{\partial c_t}{\partial h_{t-1}}
&=
\underbrace{D(c_{t-1})\,S^f_tU_f}_{\text{via forget}}
+
\underbrace{I_tS^g_tU_g}_{\text{via candidate}}
+
\underbrace{G_tS^i_tU_i}_{\text{via input}}
=:C_t^{(h)},
\\[-1pt]
\frac{\partial h_t}{\partial c_{t-1}}
&=E_tF_t,
&
\frac{\partial h_t}{\partial h_{t-1}}
&=H_tS^o_tU_o + E_tC_t^{(h)}.
\end{align}
Collecting terms,
\begin{equation}
\label{eq:jacobian_lstm}
J_t=
\begin{bmatrix}
H_tS^o_tU_o + E_tC_t^{(h)} & E_tF_t \\
C_t^{(h)} & F_t
\end{bmatrix}
\in\mathbb{R}^{2H\times 2H}.
\end{equation}

\paragraph{First-order expansion of the transport.}
Decompose each one-step Jacobian in $\mathcal{M}_{t,\ell}$ as
\[
J_t=\mathcal{T}_t+\mathcal{R}_t,
\]
where
\[
\mathcal{T}_t=
\begin{bmatrix}
0 & E_tF_t \\
0 & F_t
\end{bmatrix},
\qquad
\mathcal{R}_t=J_t-\mathcal{T}_t.
\]
Thus $\mathcal{T}_t$ contains the diagonal transport terms associated
with cell-state retention ($F_t$) and final cell-to-hidden expression
($E_t$), while $\mathcal{R}_t$ collects the recurrently mixed
corrections.

Applying the first-order expansion from Appendix~\ref{app:first_order_expansion}, we obtain:
\begin{equation}
\label{eq:lstm_first_order_prod}
\mathcal{M}_{t,\ell}
\;\approx\;
\mathcal{T}_{t,\ell}
\;+\;
\sum_{p=\ell+1}^{t}
\mathcal{T}_{t,p}\,\mathcal{R}_p\,\mathcal{T}_{p-1,\ell},
\qquad
\mathcal{T}_{a,b}=\prod_{j=b+1}^{a}\mathcal{T}_j,
\quad
\mathcal{T}_{b,b}=I.
\end{equation}
Since the loss depends on $h_t$ (Section~\ref{sec:bptt}), only the top-right block of $\mathcal{M}_{t,\ell}$, mapping $c_\ell\mapsto h_t$, contributes directly to the gradient transport.

\paragraph{Zeroth-order contributions.}
Because $\mathcal{T}_t$ is block upper triangular with a zero $h\!\to\! h$ block, the $(h,c)$ block of the transport $\mathcal{T}_{t:\ell}$ reduces to a single surviving path in which the signal remains in the cell state until time $t$:
\[
[\,\mathcal{T}_{t,\ell}\,]_{h,c}
=
E_t\,\Phi_{t,\ell},
\qquad
\Phi_{a,b}=\prod_{j=b+1}^{a}F_j,
\quad
\Phi_{b,b}=I.
\]
In particular, the $c\!\to\!h$ transition can occur only at the final time $t$.
Extracting the diagonal gives the neuron-wise zeroth-order rates
\begin{equation}
\label{eq:lstm_gamma0}
\gamma^{(0)}_{t,\ell}
=
\mathrm{diagvec}\!\bigl(E_t\Phi_{t,\ell}\bigr)
\in\mathbb{R}^{H},
\qquad
\gamma^{(0,q)}_{t,\ell}
=
e_{t}^{(q)}\prod_{j=\ell+1}^{t}f_{j}^{(q)},
\qquad
e_{t}^{(q)}=o_{t}^{(q)}\left(1-\tanh^2(c_{t}^{(q)})\right).
\end{equation}

\paragraph{First-order diagonal correction.}
From the first-order expansion~\eqref{eq:lstm_first_order_prod}, the contribution of a single recurrent-mixing insertion to the top-right block of the transport is
\[
\gamma^{(1)}_{t,\ell}
=
\sum_{p=\ell+1}^{t}
\mathrm{diagvec}\!\Big(
[\,\mathcal{T}_{t,p}\,\mathcal{R}_p\,\mathcal{T}_{p-1,\ell}\,]_{h,c}
\Big)
\in\mathbb{R}^{H},
\]
where $[\,\cdot\,]_{h,c}$ denotes the $H\times H$ block mapping $c_\ell\mapsto h_t$.
Unlike the zeroth-order case, the presence of the recurrent correction $\mathcal{R}_p$ reintroduces $h\!\to\!h$ transport, allowing a single mixing event to occur at any intermediate time $p\in\{\ell+1,\dots,t\}$.
This term therefore collects the diagonal components of the first-order Fr\'echet expansion, corresponding to neuron-wise self-couplings induced by a single insertion of the recurrent mixing blocks.
Off-diagonal entries encode cross-neuron interactions and are intentionally discarded, since they do not admit a neuron-wise time-scale interpretation.

\paragraph{LSTM effective learning rates.}
With global learning rate $\mu>0$, we define the per-lag, per-neuron effective learning rates
\begin{equation}
\label{eq:lstm_mu_eff_vec}
\mu_{t,\ell}
=
\mu\bigl(\gamma^{(0)}_{t,\ell}+\gamma^{(1)}_{t,\ell}\bigr)
\in\mathbb{R}^{H},
\qquad
\mu^{(q)}_{t,\ell}
=
\mu\bigl(\gamma^{(0,q)}_{t,\ell}+\gamma^{(1,q)}_{t,\ell}\bigr).
\end{equation}
These quantities act as scalar multipliers modulating the contribution
of gradients originating at lag $(t-\ell)$ to parameter updates,
providing a neuron-wise characterization of the effective learning
dynamics induced by the LSTM gates.
The transport factor appearing in Section~\ref{sec:generalized_elr} is
\[
\Gamma^{(q)}_{t,\ell}
=
\gamma^{(0,q)}_{t,\ell}+\gamma^{(1,q)}_{t,\ell}.
\]

\subsection{GRU: one-step Jacobian and effective learning rates}
\label{app:gru_effective}

\paragraph{Dynamics.}
A GRU updates
\begin{align}
a^z_t &= W_z x_t + U_z h_{t-1} + b_z,
& z_t &= \sigma(a^z_t),
& & \text{(update/leak)}
\\[-1pt]
a^r_t &= W_r x_t + U_r h_{t-1} + b_r,
& r_t &= \sigma(a^r_t),
& & \text{(reset/filtering)}
\\[-1pt]
a^g_t &= W_h x_t + U_h(r_t\odot h_{t-1}) + b_h,
& g_t &= \tanh(a^g_t),
& & \text{(candidate)}
\\[4pt]
h_t &= (1-z_t)\odot h_{t-1} + z_t\odot g_t.
\end{align}
Let
\[
Z_t=D(z_t),\qquad
R_t=D(r_t),\qquad
G_t=D(g_t),
\]
\[
D_{h,t-1}=D(h_{t-1}),\qquad
S^z_t=S^{\sigma}(a^z_t),\qquad
S^r_t=S^{\sigma}(a^r_t),\qquad
S^g_t=S^{\tanh}(a^g_t).
\]

\paragraph{One-step Jacobian.}
Differentiating the leak and update paths gives
\begin{align}
\frac{\partial}{\partial h_{t-1}}
\bigl[(1-z_t)\odot h_{t-1}\bigr]
&=
(I-Z_t)-D_{h,t-1}S^z_tU_z,
\label{eq:gru_leak_path_block}
\\[-1pt]
\frac{\partial g_t}{\partial h_{t-1}}
&=
S^g_tU_h\bigl(R_t + D_{h,t-1}S^r_tU_r\bigr),
\label{eq:gru_g_block}
\\[-1pt]
\frac{\partial}{\partial h_{t-1}}
\bigl[z_t\odot g_t\bigr]
&=
G_tS^z_tU_z + Z_t\frac{\partial g_t}{\partial h_{t-1}}.
\label{eq:gru_update_block}
\end{align}
Summing contributions yields
\begin{equation}
\label{eq:jacobian_gru}
J_t
=
(I-Z_t)
+
(G_t-D_{h,t-1})S^z_tU_z
+
Z_tS^g_tU_hR_t
+
Z_tS^g_tU_hD_{h,t-1}S^r_tU_r.
\end{equation}

\paragraph{First-order expansion of the transport.}
Decompose the one-step Jacobian as
\[
J_t=\mathcal{T}_t+\mathcal{R}_t,
\]
where
\[
\mathcal{T}_t=I-Z_t
\]
is the diagonal retention operator induced by the update gate, and $\mathcal{R}_t$ collects all remaining recurrent mixing terms.
For $\mathcal{M}_{t,\ell}$, the first-order expansion gives
\begin{equation}
\label{eq:gru_first_order_prod}
\mathcal{M}_{t,\ell}
\;\approx\;
\mathcal{T}_{t,\ell}
\;+\;
\sum_{p=\ell+1}^{t}
\mathcal{T}_{t,p}\,\mathcal{R}_p\,\mathcal{T}_{p-1,\ell},
\qquad
\mathcal{T}_{a,b}=\prod_{j=b+1}^{a}\mathcal{T}_j,
\quad
\mathcal{T}_{b,b}=I.
\end{equation}
Since $\mathcal{T}_t=I-Z_t$ is diagonal, we may equivalently write
\[
\mathcal{T}_{t,\ell}
=
\Phi_{t,\ell},
\qquad
\Phi_{a,b}=\prod_{j=b+1}^{a}(I-Z_j),
\quad
\Phi_{b,b}=I.
\]

\paragraph{Zeroth-order contributions.}
The diagonal product $\Phi_{t,\ell}$ yields the primary update-gate retention envelope
\begin{equation}
\label{eq:gru_gamma0}
\gamma^{(0)}_{t,\ell}
=
\mathrm{diagvec}\!\bigl(\Phi_{t,\ell}\bigr)
\in\mathbb{R}^{H},
\qquad
\gamma^{(0,q)}_{t,\ell}
=
\prod_{j=\ell+1}^{t}(1-z_{j}^{(q)}),
\end{equation}
which governs the zeroth-order neuron-wise time scales associated with
the update gate.

In addition, the multiplicative action of the reset gate along the
candidate pathway induces further diagonal attenuations that are useful
to track explicitly.
We therefore define the auxiliary diagonal pathwise envelopes
\begin{equation}
\label{eq:gru_rho_eta}
\rho^{(0,q)}_{t,\ell}
=
\prod_{j=\ell+1}^{t}r_{j}^{(q)},
\qquad
\eta^{(0,q)}_{t,\ell}
=
\prod_{j=\ell+1}^{t}(1-z_{j}^{(q)})\,r_{j}^{(q)},
\end{equation}
which capture pure reset attenuation and mixed update-reset attenuation, respectively.
These quantities isolate the shrinkage associated with reset-controlled chains and are reported as complementary diagonal envelopes, rather than as additional terms in the strict $\mathcal{T}_t+\mathcal{R}_t$ product expansion.
They are included in the effective learning rates for interpretability, as they capture additional multiplicative attenuation along the candidate pathway.

\paragraph{First-order diagonal correction.}
The first-order diagonal correction associated with recurrent mixing is obtained by extracting the diagonal of the first-order term in~\eqref{eq:gru_first_order_prod}:
\[
\gamma^{(1)}_{t,\ell}
=
\sum_{p=\ell+1}^{t}
\mathrm{diagvec}\!\left(\mathcal{T}_{t,p}\,\mathcal{R}_p\,\mathcal{T}_{p-1,\ell}\right)
=
\sum_{p=\ell+1}^{t}
\mathrm{diagvec}\!\left(\Phi_{t,p}\,\mathcal{R}_p\,\Phi_{p-1,\ell}\right)\in\mathbb{R}^{H}.
\]
This term collects the diagonal components arising from a single
insertion of the recurrent mixing operator into an otherwise
gate-diagonal transport.
As in the LSTM case, off-diagonal components encode cross-neuron
interactions and are intentionally discarded in order to preserve a
neuron-wise time-scale interpretation.

\paragraph{GRU effective learning rates.}
With global learning rate $\mu>0$, we define the per-lag, per-neuron
effective learning rates as
\begin{equation}
\label{eq:gru_mu_eff_vec}
\mu^{(q)}_{t,\ell}
=
\mu\Bigl(
\gamma^{(0,q)}_{t,\ell}
+
\rho^{(0,q)}_{t,\ell}
+
\eta^{(0,q)}_{t,\ell}
+
\gamma^{(1,q)}_{t,\ell}
\Bigr),
\qquad q=1,\dots,H.
\end{equation}
These quantities provide a diagonal, neuron-wise summary of how update and reset gates jointly shape the effective learning dynamics across temporal lags.
The transport factor appearing in Section~\ref{sec:generalized_elr} is
\[
\Gamma^{(q)}_{t,\ell}
=
\gamma^{(0,q)}_{t,\ell}
+
\rho^{(0,q)}_{t,\ell}
+
\eta^{(0,q)}_{t,\ell}
+
\gamma^{(1,q)}_{t,\ell},
\]
so that the auxiliary envelopes $\rho^{(0)}_{t,\ell}$ and
$\eta^{(0)}_{t,\ell}$ are included in $\Gamma$ and hence in the
learnability envelope~$f(\ell)$.

\subsection{Baseline gated RNNs}
\label{app:diag_gated_rnns}

We also consider the baseline gated RNN models introduced
in~\cite{livi2025timescale}: a per-neuron (diagonal) gate, a shared
(global) scalar gate, and a constant scalar gate.
All follow the common update template
\[
h_t
=
(1-s_t)\odot h_{t-1}
+
s_t\odot \tilde{h}_t,
\qquad
\tilde{h}_t=\tanh(a^h_t),
\qquad
a^h_t=W_hx_t+U_hh_{t-1}+b_h,
\]
but differ in how the gate $s_t$ is produced.

For these architectures, the neuron-wise transport factor takes the
form
\[
\Gamma^{(q)}_{t,\ell}
=
\gamma^{(0,q)}_{t,\ell}
+
\gamma^{(1,q)}_{t,\ell},
\]
where $\gamma^{(0,q)}_{t,\ell}$ denotes the zeroth-order gate-product
envelope and $\gamma^{(1,q)}_{t,\ell}$ collects the diagonal
first-order corrections arising from recurrent mixing terms.

\subsubsection*{Per-neuron gate (DiagGate)}

The gate is computed coordinate-wise:
\begin{equation}
\label{eq:diaggate}
a^s_t=W_sx_t+U_sh_{t-1}+b_s,
\qquad
s_t=\sigma(a^s_t)\in(0,1)^H.
\end{equation}
Define
\[
S_t=D(s_t),
\qquad
S^s_t=S^{\sigma}(a^s_t),
\qquad
S^h_t=S^{\tanh}(a^h_t).
\]
The exact one-step Jacobian
$J_t=\partial h_t/\partial h_{t-1}\in\mathbb{R}^{H\times H}$ is
\begin{equation}
\label{eq:jacobian_diaggate}
J_t
=
\underbrace{(I-S_t)}_{\text{leak}}
+
\underbrace{\bigl(D(\tilde{h}_t)-D(h_{t-1})\bigr)S^s_tU_s}_{\text{gate sensitivity}}
+
\underbrace{S_tS^h_tU_h}_{\text{candidate path}}.
\end{equation}
The leak term $(I-S_t)$ is diagonal; the remaining two terms contain
$U_s$ and $U_h$ and are generally full rank, mixing information across
neurons.

\paragraph{Effective learning rates.}
Taking
\[
\mathcal{T}_t=I-S_t,
\qquad
\mathcal{R}_t=J_t-\mathcal{T}_t,
\]
the zeroth-order envelope is
\[
\gamma^{(0,q)}_{t,\ell}
=
\prod_{j=\ell+1}^{t}(1-s_{j,q}),
\]
yielding per-neuron effective learning rates
\[
\mu^{(q)}_{t,\ell}
=
\mu\bigl(\gamma^{(0,q)}_{t,\ell}
+\gamma^{(1,q)}_{t,\ell}\bigr),
\]
where $\gamma^{(1,q)}_{t,\ell}$ arises from the diagonal of the
first-order terms generated by the gate-sensitivity and candidate-path
corrections in Eq.~\eqref{eq:jacobian_diaggate}.

\subsubsection*{Shared global scalar gate (SharedGate)}

The gate is a single scalar at each time step:
\begin{equation}
\label{eq:sharedgate}
a^s_t=w_s^\top x_t+u_s^\top h_{t-1}+b_s,
\qquad
s_t=\sigma(a^s_t)\in(0,1).
\end{equation}
The update is
\[
h_t=(1-s_t)h_{t-1}+s_t\tilde{h}_t,
\qquad
\tilde{h}_t=\tanh(a^h_t).
\]
Using
\[
\frac{\partial s_t}{\partial h_{t-1}}=s_t(1-s_t)\,u_s,
\]
the Jacobian is
\begin{equation}
\label{eq:jacobian_sharedgate}
J_t
=
\underbrace{(1-s_t)I}_{\text{leak}}
+
\underbrace{s_tS^h_tU_h}_{\text{candidate path}}
+
\underbrace{s_t(1-s_t)(\tilde{h}_t-h_{t-1})u_s^\top}_{\text{rank-1 gate sensitivity}}.
\end{equation}
The last term is a rank-1 outer product and is the only source of
cross-neuron coupling other than $U_h$.

\paragraph{Effective learning rates.}
Taking
\[
\mathcal{T}_t=(1-s_t)I,
\qquad
\mathcal{R}_t=J_t-\mathcal{T}_t,
\]
the zeroth-order envelope is
\[
\gamma^{(0)}_{t,\ell}
=
\prod_{j=\ell+1}^{t}(1-s_j),
\]
which is identical for all neurons.
Hence
\[
\mu^{(q)}_{t,\ell}
=
\mu\bigl(\gamma^{(0)}_{t,\ell}
+\gamma^{(1,q)}_{t,\ell}\bigr),
\]
with neuron-dependent corrections $\gamma^{(1,q)}_{t,\ell}$ coming
from the diagonal of the first-order terms generated by the candidate
path and the rank-1 gate sensitivity in
Eq.~\eqref{eq:jacobian_sharedgate}.

\subsubsection*{Constant scalar gate (ConstGate)}

As a minimal baseline, the gate is fixed to a constant scalar
$s\in(0,1)$, independent of $t$, inputs, and hidden state.
The update is
\begin{equation}
\label{eq:constgate}
h_t
=
(1-s)h_{t-1}+s\tilde{h}_t,
\qquad
\tilde{h}_t=\tanh(a^h_t),
\qquad
a^h_t=W_hx_t+U_hh_{t-1}+b_h.
\end{equation}
The Jacobian is
\begin{equation}
\label{eq:jacobian_constgate}
J_t
=
\underbrace{(1-s)I}_{\text{fixed leak}}
+
\underbrace{sS^h_tU_h}_{\text{candidate path}}.
\end{equation}
Unlike DiagGate and SharedGate, there are no gate sensitivities,
because $s$ is constant and not learned.
The dynamics are therefore a rigid combination of identity and the
candidate path, with a fixed leakage rate $(1-s)$ applied to all
neurons.

\paragraph{Effective learning rates.}
Taking
\[
\mathcal{T}_t=(1-s)I,
\qquad
\mathcal{R}_t=J_t-\mathcal{T}_t,
\]
the zeroth-order envelope is
\[
\gamma^{(0)}_{t,\ell}=(1-s)^{t-\ell},
\]
identical for all neurons.
Thus
\[
\mu^{(q)}_{t,\ell}
=
\mu\bigl(\gamma^{(0)}_{t,\ell}
+\gamma^{(1,q)}_{t,\ell}\bigr),
\]
with $\gamma^{(1,q)}_{t,\ell}$ depending only on the diagonal of the
first-order candidate-path correction.

\subsection{Gates as implicit multi-rate optimizers}
\label{app:multirate_discussion}

The effective learning rates derived above quantify how gates reweight
BPTT contributions at lag $\ell$ on a per-neuron basis, even under a
fixed global learning rate $\mu$ in SGD-like optimizers.
Zeroth-order terms produce diagonal transport envelopes, while the
first-order terms introduce anisotropy: recurrent matrices modulated by
gate slopes mix coordinates and steer updates into privileged
subspaces.
As a result, gated RNNs act as implicit \emph{multi-rate optimizers}:
they induce heterogeneous, lag-dependent learning rates across neurons
and selectively bias update directions.
This interpretation of gates as multi-rate optimizers was originally
developed in~\cite{livi2025timescale} for the baseline gated models of
Section~\ref{app:diag_gated_rnns} and extends naturally to the LSTM
and GRU derivations above.

\section{Projected adaptive base rate via Rayleigh quotient}
\label{app:adaptive_base_rate}

This appendix provides additional details on the construction of the neuron--lag adaptive base rate $\Lambda^{(q)}_{r,\ell}$ used in Section~\ref{sec:generalized_elr}.
The key idea is to extract the effective optimizer learning rate along the parameter-space direction associated with neuron $q$ at lag $\ell$.

Under the lag re-indexing introduced at the end of Section~\ref{sec:bptt}, recall that
\begin{equation}
B_\ell(\theta)
=
\frac{\partial h_{t-\ell}}{\partial \theta}
\in
\mathbb{R}^{H \times P},
\end{equation}
where $H$ denotes the hidden-state dimension and $P=\dim(\theta)$ the number of trainable parameters. Its $q$-th row is
\begin{equation}
B_\ell^{(q)}
=
\frac{\partial [h_{t-\ell}]_q}{\partial \theta}
\in
\mathbb{R}^{1\times P}.
\end{equation}
Thus, $B_\ell^{(q)}$ identifies the parameter-space sensitivity of neuron~$q$ at the originating time $t-\ell$.
Denoting by
\begin{equation}
b_q=(B_\ell^{(q)})^\top\in\mathbb{R}^{P},
\end{equation}
this vector specifies the parameter direction associated with neuron $q$ at lag~$\ell$ in the BPTT decomposition. Equivalently, for a local parameter perturbation $\delta\theta$, the first-order change is
\begin{equation}
\delta[h_{t-\ell}]_q
=
\langle b_q,\delta\theta\rangle.
\end{equation}
Hence, $b_q$ is the sensitivity direction determining how parameter updates affect that neuron coordinate at the originating time, to first order; components orthogonal to $b_q$ do not contribute to this scalar coordinate at that order.

Under adaptive optimization, gradient updates at optimizer iteration~$r$ (cf.\ Eq.~\eqref{eq:sgd_update_adaptive}) are preconditioned by a diagonal matrix
\begin{equation}
\Lambda_r = \mathrm{diag}(\lambda_{1,r},\dots,\lambda_{P,r}),
\end{equation}
so that the optimizer rescales each parameter coordinate independently;
here $\lambda_{i,r}>0$ is the per-parameter adaptive learning rate of coordinate $i$ at training step~$r$, determined by the optimizer state.

Because $\Lambda_r$ acts anisotropically, the preconditioned sensitivity vector $\Lambda_r b_q$ is generally \emph{not} collinear with $b_q$.
Therefore, the optimizer does not simply rescale the neuron-specific direction but also distorts it in parameter space.
To isolate the component of the update that remains aligned with the original sensitivity direction, we decompose
\begin{equation}
\Lambda_r b_q
=
\Lambda^{(q)}_{r,\ell}\, b_q
+
r_q,
\qquad
\langle b_q,r_q\rangle = 0,
\end{equation}
where the residual $r_q$ is orthogonal to $b_q$.
The scalar coefficient $\Lambda^{(q)}_{r,\ell}$ therefore represents the magnitude of the orthogonal projection of the preconditioned vector $\Lambda_r b_q$ onto the one-dimensional subspace $\mathrm{span}(b_q)$.

This projection coefficient is uniquely given by the Rayleigh quotient~\cite{horn2012matrix}
\begin{equation}
\Lambda^{(q)}_{r,\ell}
=
\frac{
\langle b_q,\Lambda_r b_q\rangle
}{
\langle b_q,b_q\rangle
}
=
\frac{
B_\ell^{(q)}\,\Lambda_r\,(B_\ell^{(q)})^\top
}{
B_\ell^{(q)}(B_\ell^{(q)})^\top
}.
\end{equation}

Expanding the notation yields
\begin{equation}
\Lambda^{(q)}_{r,\ell} = \frac{\sum_{i=1}^{P}\lambda_{i,r}\,[B_\ell^{(q)}]_i^2}{\sum_{i=1}^{P}[B_\ell^{(q)}]_i^2} > 0.
\end{equation}
Thus, the effective base rate is a weighted average of the per-parameter adaptive learning rates, where the weights are determined by the squared parameter sensitivities of neuron $q$.

The construction has three structural properties that justify the Rayleigh quotient as the canonical per-neuron scalar reduction of $\Lambda_r$ along $b_q$.
First, it admits a variational characterization~\cite{horn2012matrix}:
$\Lambda^{(q)}_{r,\ell}$ is the unique scalar $c\in\mathbb{R}$
minimizing $\|\Lambda_r b_q - c\,b_q\|_2^2$, i.e.\ the best scalar
approximation of the optimizer's anisotropic action $\Lambda_r b_q$
within $\mathrm{span}(b_q)$ in the least-squares sense.
Second, it is scale-invariant in the sensitivity direction:
$\Lambda^{(q)}_{r,\ell}(\alpha\,b_q)=\Lambda^{(q)}_{r,\ell}(b_q)$ for any
$\alpha>0$, so the projected base rate depends only on the direction of
neuron $q$'s parameter sensitivity, not its magnitude, and is unaffected
by arbitrary normalization of the sensitivity vector.
Third, it reduces to plain SGD under isotropic preconditioning:
when $\Lambda_r = \mu\,I$, $\Lambda^{(q)}_{r,\ell}=\mu$ for every neuron
$q$, so the adaptive construction is a strict generalization of the effective learning rate derived with plain SGD $\mu^{(q)}_{t,\ell}=\mu\,\Gamma^{(q)}_{t,\ell}$.
Substituting $\Lambda_r=\mu I$ directly into the Rayleigh quotient gives
\begin{equation}
\label{eq:Lambda_q_SGD_reduces_to_mu}
\Lambda^{(q)}_{r,\ell}
=
\frac{\langle b_q, \mu I b_q \rangle}{\langle b_q, b_q \rangle}
=
\mu.
\end{equation}
Thus, the projected adaptive base rate collapses to the global learning rate $\mu$, and the generalized effective learning rates $\mu^{(q)}_{t,\ell} = \Lambda^{(q)}_{r,\ell}\Gamma^{(q)}_{t,\ell}$ in Eq.~\ref{eq:GELR} recover the original SGD formulation $\mu^{(q)}_{t,\ell} = \mu\,\Gamma^{(q)}_{t,\ell}$ in Eq.~\ref{eq:effective_learning_rate_SGD}.

In addition, because $\Lambda^{(q)}_{r,\ell}$ is a convex combination of positive adaptive rates $\lambda_{i,r}$, it satisfies
\begin{equation}
\label{eq:rayleigh_bound}
\min_i \lambda_{i,r}
\le
\Lambda^{(q)}_{r,\ell}
\le
\max_i \lambda_{i,r}.
\end{equation}

\section{Envelope validation}
\label{app:envelope_validation}

As discussed in Section~\ref{sec:envelope_GELR}, the envelope
$f(\ell)$ defined in~\eqref{eq:envelope_def} relies on transport
factors $\Gamma^{(q)}_{t,\ell}$ obtained from the first-order diagonal
expansion of the Jacobian product $\mathcal{M}_{t,\ell}$.  This
expansion is what makes it possible to assign each hidden unit a
lag-dependent rate $\mu^{(q)}_{t,\ell}$ with a transparent
decomposition into gating and mixing contributions---a structure that
cannot be read off the diagonal of $\mathcal{M}_{t,\ell}$ directly.
This appendix verifies that the interpretability gained by the
first-order expansion comes at no cost to the envelope's decay
profile: $f(\ell)$ preserves the decay trend of $f_{\mathcal{M}}(\ell)$.
We establish this with model-free linear statistics that assume no
parametric form for the decay.

Five architectures (ConstGate, SharedGate, DiagGate, GRU, LSTM) $\times$
three optimizers (SGD, AdamW, RMSProp) $\times$ five random seeds $= 75$ runs.
Hidden size $H{=}64$, sequence length $T{=}500$, batch size $B{=}256$, 200
training epochs on a sinusoidal delayed-regression task.  For each run, 800
$(t,\ell)$ samples per lag are drawn at 12 lags spanning 1--245.
At every sample we compute the exact envelope
$f_{\mathcal{M}}(\ell)$ and the first-order approximation $f(\ell)$ in Eq.~\ref{eq:envelope_def}.
We report two statistics: the Spearman rank correlation~$\rho$, which tests whether the two envelopes rank lags in the
same order, and the Pearson correlation~$r$ on the $\log_{10}$-transformed
envelopes, which tests whether the decay shapes are linearly related in
log-space.  Both are computed on only 12 points (one per lag), making them
deliberately strict: even a single outlier substantially reduces the
correlation.

\begin{table}[h]
\centering
\small
\begin{tabular}{l c c}
\toprule
Architecture & Spearman $\rho$ & Pearson $r$ \\
\midrule
ConstGate  & $1.000 \pm 0.000$ & $0.904 \pm 0.046$ \\
SharedGate & $1.000 \pm 0.000$ & $0.934 \pm 0.025$ \\
DiagGate   & $1.000 \pm 0.000$ & $0.957 \pm 0.047$ \\
GRU        & $1.000 \pm 0.000$ & $0.986 \pm 0.013$ \\
LSTM       & $0.992 \pm 0.009$ & $0.985 \pm 0.018$ \\
\midrule
\textit{All (75 runs)} & $\geq 0.972$ & $\geq 0.823$ \\
\bottomrule
\end{tabular}
\caption{Envelope validation across 5 architectures, 3 optimizers, and 5
seeds.  Each row aggregates 15 runs (3 optimizers $\times$ 5 seeds); values
are mean $\pm$ std.  The bottom row reports the worst case across all 75
runs.}
\label{tab:envelope_validation}
\end{table}

Spearman $\rho \geq 0.972$ across all 75 runs confirms that the first-order
expansion never disrupts the lag ordering of the envelope.  Pearson
$r \geq 0.823$ (mean $0.95$) on the $\log_{10}$-transformed envelopes
establishes that $\log f \approx a\,\log f_{\mathcal{M}} + b$, i.e.\ the
two envelopes are related by a power-law rescaling
$f \propto f_{\mathcal{M}}^{\,a}$.  Such a rescaling preserves the monotonic
and scaling structure on which the learnability analysis depends.

\section{Random projections: validation and multi-probe aggregation}
\label{app:alpha_chain_validation}

\subsection{Empirical validation of the one-dimensional matched statistic}
\label{app:projection_validation}

The learnability theory developed in this paper requires a scalar
matched statistic whose expectation factorizes into the envelope
$f(\ell)$ and an alignment coefficient
(Eq.~\eqref{eq:expectation_matched_statistic}).
In practice we obtain this statistic by projecting onto a fixed random
direction $w\in\mathbb{R}^P$.
This loses directional information, so the empirical question is
whether the resulting one-dimensional statistic is stable across tested
projection directions and has comparable tail behavior to the exact
lag-specific transport object.
To check robustness, throughout this appendix we estimate tail indices
with two complementary methods: the Koutrouvelis ECF regression
estimator~\cite{koutrouvelis1980regression} and a bootstrapped MCC
quantile estimator~\cite{mcculloch1986simple}.

To assess whether the one-dimensional matched statistic is empirically
justified, we compare three scalar objects on frozen trained
checkpoints, estimated on independent diagnostic sequences not used
during training:
\begin{enumerate}[label=(\alph*)]
\item \emph{Projected full-gradient noise.}
For a given projection direction $w$, define the scalar
\[
\xi^{(n)} \;=\; \langle g^{(n)}_{\mathrm{batch}}(\theta) - g_{\mathrm{full}}(\theta),\, w \rangle,
\]
where $g^{(n)}_{\mathrm{batch}}$ is a mini-batch gradient and
$g_{\mathrm{full}}$ is the full-dataset gradient.
This object captures the heavy-tailed structure of the projected
stochastic gradient noise motivated by~\cite{simsekli2019tail}, without
any lag decomposition.

\item \emph{Exact lag-projected sequence average.}
The exact lag-$\ell$ contribution to the parameter gradient is
$g_{t,\ell}(\theta) = \delta_t^\top \mathcal{M}_{t,\ell}\,B_\ell(\theta)$
(Eq.~\eqref{eq:lagl_gradient_contribution}).
Projecting this onto the same direction $w$ and averaging over
time positions within a sequence yields a scalar that reflects the
exact transport through the full Jacobian product, without the
first-order diagonal approximation.

\item \emph{First-order matched statistic, sequence average.}
This is the empirical matched statistic
$\widetilde{S}_N(\ell)$
defined in
Eq.~\eqref{eq:empirical_cross_sequence_average}, built from the
neuronwise alignment variables $\zeta^{(q)}_{t,\ell}$
(Eq.~\eqref{eq:neuronwise_alignment}) weighted by the generalized
effective learning rates $\mu^{(q)}_{t,\ell}$.
It is the theory-facing object whose expectation factorizes through
the envelope $f(\ell)$.
\end{enumerate}
Object~(a) provides an architecture-level reference for the projected gradient noise.
Object~(b) isolates the exact geometric transport at a specific lag, without the first-order envelope approximation or the optimizer weighting.
Object~(c) is the statistic actually used in the paper.
Comparing (b) and (c) reveals how much the first-order diagonal
approximation and the optimizer-rate weighting alter the tail behavior,
while comparing (a) and (b) shows how lag-specific decomposition
affects the projected gradient statistics.

Note that objects (b) and (c) differ in two respects:
the exact-versus-approximate Jacobian product and the
presence of the adaptive base rates $\Lambda^{(q)}_{r,\ell}$ in the
first-order construction.
The diagnostic therefore measures the combined effect of these two
factors, not each in isolation.

For each object, we estimate the tail index $\hat\alpha$ with both
estimators, namely the Koutrouvelis ECF regression estimator and the
bootstrapped MCC quantile estimator, applied to the
corresponding scalar samples.
Reporting both verifies that the qualitative conclusions do not depend
on the tail-index estimator.
We evaluate two architectures (DiagGate and GRU) across two training
seeds and multiple independent projection directions $w$.

Table~\ref{tab:alpha_chain_summary} reports the estimated tail indices.
The following patterns are consistent across seeds and projection
directions:
\begin{enumerate}[label=(\roman*)]
\item
\emph{Temporal averaging is a robust regularizer.}
In all cases, the sequence-averaged objects are substantially closer to
the Gaussian boundary ($\alpha=2$) than their instantaneous
counterparts (not shown in the table).
This confirms that averaging over the $T-\ell$ admissible time
positions within each sequence strongly attenuates heavy-tailed
behavior.

\item
\emph{The matched statistic has comparable tail behavior.}
Across both seeds and all tested projection directions, the
tail index of the first-order matched sequence average~(c) is
broadly consistent with that of the exact lag-projected sequence
average~(b), with the matched statistic often slightly closer to the
Gaussian boundary, especially for GRU.
This indicates that the first-order approximation and optimizer
weighting do not introduce qualitatively different tail behavior
relative to the exact transport.

\item
\emph{Projection variability is modest relative to seed variability.}
Within each training seed, different projection directions $w$ yield
tail-index ranges that are narrow compared to the shift observed
across training seeds.
The largest source of variation in the absolute $\hat\alpha$ values is
the training seed, not the choice of projection direction.
\end{enumerate}
These findings do not establish a seed-invariant quantitative
propagation law from the projected full-gradient noise to the
matched statistic.
The absolute $\hat\alpha$ levels shift materially across training seeds
-- most visibly for DiagGate, where the full-gradient probe moves from
near-Gaussian in one seed to clearly heavy-tailed in another.
Instead, the diagnostic supports a qualitative conclusion: the
one-dimensional matched statistic is a stable proxy for the underlying
lag-specific gradient transport, and temporal averaging is the dominant
regularizing mechanism at the sequence level.
This is the property required by the detection framework, which depends
on the tail behavior of the matched statistic itself
(Appendix~\ref{app:noise_floor}), not on precise quantitative
agreement with the full-gradient noise.

The modest projection variability documented above motivates a natural
extension: running the learnability pipeline for $K>1$ independent
projection directions and aggregating the resulting matched statistics.
The next subsection develops this multi-projection extension formally.

\begin{table}[tp]
\centering
\caption{Tail-index estimates ($\hat\alpha$) for the three diagnostic
objects, with ECF and MCC ranges reported in separate columns.
For the full-gradient probe, each projection direction yields one
estimate; the range is over directions $w$.
For the lag-specific objects, each direction is first averaged across
the four diagnostic lags ($\ell\in\{4,64,256,512\}$), then the range
is taken over directions.
All estimates are marked reliable by both estimators.}
\label{tab:alpha_chain_summary}
\smallskip
\renewcommand{\arraystretch}{1.15}
\begin{tabular}{c c c l c c}
\toprule
Seed & Model & \#\,$w$ & Diagnostic object & ECF range & MCC range \\
\midrule
\multirow{3}{*}{13} & \multirow{3}{*}{diag} & \multirow{3}{*}{4}
  & Full-gradient probe                 & $[1.995,\,2.000]$ & $[1.963,\,2.000]$ \\
  &                       &
  & Exact lag-projected seq.\ avg.      & $[1.904,\,1.935]$ & $[1.861,\,1.919]$ \\
  &                       &
  & Matched statistic seq.\ avg.        & $[1.864,\,1.951]$ & $[1.723,\,1.879]$ \\
\midrule
\multirow{3}{*}{13} & \multirow{3}{*}{gru} & \multirow{3}{*}{4}
  & Full-gradient probe                 & $[1.992,\,1.999]$ & $[1.931,\,2.000]$ \\
  &                       &
  & Exact lag-projected seq.\ avg.      & $[1.734,\,1.802]$ & $[1.607,\,1.707]$ \\
  &                       &
  & Matched statistic seq.\ avg.        & $[1.937,\,1.955]$ & $[1.875,\,1.884]$ \\
\midrule
\multirow{3}{*}{17} & \multirow{3}{*}{diag} & \multirow{3}{*}{8}
  & Full-gradient probe                 & $[1.620,\,1.773]$ & $[1.630,\,1.770]$ \\
  &                       &
  & Exact lag-projected seq.\ avg.      & $[1.357,\,1.484]$ & $[1.309,\,1.441]$ \\
  &                       &
  & Matched statistic seq.\ avg.        & $[1.787,\,1.951]$ & $[1.719,\,1.923]$ \\
\midrule
\multirow{3}{*}{17} & \multirow{3}{*}{gru} & \multirow{3}{*}{8}
  & Full-gradient probe                 & $[1.993,\,2.000]$ & $[1.957,\,2.000]$ \\
  &                       &
  & Exact lag-projected seq.\ avg.      & $[1.924,\,1.963]$ & $[1.876,\,1.905]$ \\
  &                       &
  & Matched statistic seq.\ avg.        & $[1.913,\,1.977]$ & $[1.869,\,1.955]$ \\
\bottomrule
\end{tabular}
\end{table}

\subsection{Multi-projection aggregation}
\label{app:multi_projection}

The single-projection matched statistic used in the main text can be
strengthened by averaging over $K$ independent projection directions.
This subsection shows that multi-projection aggregation does not
change the envelope decay class or any scaling exponent; its sole
effect is to replace the effective sample size $N$ with $NK$ (under
a cross-projection independence condition), thereby potentially enlarging the
learnability window.

\paragraph{Aggregated matched statistic.}
Let $w_1,\dots,w_K\in\mathbb{S}^{P-1}$ be independent random unit-norm directions, drawn independently of the data.
For each direction $w_k$, we compute a matched statistic $\widehat{S}_N^{(k)}(\ell)$ as defined in Eq.~\eqref{eq:theoretical_cross_sequence_average}.
The aggregated statistic is
\begin{equation}
\label{eq:aggregated_matched_statistic}
\widehat{S}_{N,K}(\ell)
=
\frac{1}{K}\sum_{k=1}^{K}\widehat{S}_N^{(k)}(\ell).
\end{equation}

The envelope $f(\ell)=\sum_q|\mu^{(q)}_{t,\ell}|$ is defined before
the projection (Eq.~\eqref{eq:empirical_envelope_definition}), so it
is unchanged by the choice or number of directions.
The expected signal of the aggregated statistic is
\begin{equation}
\mathbb{E}\left[\widehat{S}_{N,K}(\ell)\right]
= f(\ell)\,\overline{m}_{\mu,K}(\ell),
\qquad
\overline{m}_{\mu,K}(\ell)=\frac{1}{K}\sum_{k=1}^{K}\overline{m}_\mu^{(k)}(\ell),
\end{equation}
where $\overline{m}_\mu^{(k)}(\ell)$ is the alignment coefficient \eqref{eq:m_bar_mu_def} for direction $w_k$.

Under the same $\alpha$-stable modeling used for a single probe, if
the $K$ projected statistics are independent (or sufficiently weakly
dependent) at fixed $\ell$, averaging reduces the noise scale:
\[
\widehat{S}_{N,K}(\ell) - \mathbb{E}\left[\widehat{S}_{N,K}(\ell)\right]
\;\sim\;
\mathcal{S}\alpha\mathcal{S}\!\left(0,\ \sigma_\alpha(\ell)N^{-1/\kappa_\alpha}K^{-1/\kappa_\alpha}\right).
\]
The key factor is $K^{-1/\kappa_\alpha}$: averaging $K$ independent $\alpha$-stable variates reduces the scale by $K^{1/\kappa_\alpha}$.

\paragraph{Consequences for the scaling laws.}
Since $f(\ell)$ is unaffected, the three canonical decay classes (exponential, power-law, logarithmic) are preserved.
The detectability threshold therefore becomes
\begin{equation}
\label{eq:multi_proj_threshold}
f(\ell) \geq \frac{d_{\alpha,\epsilon}\sigma_\alpha(\ell)}{(NK)^{1/\kappa_\alpha}\,\overline{m}_{\mu,K}(\ell)}.
\end{equation}
The master proportionality of Eq.~\eqref{eq:master_prop_maintext} becomes
\[
N_K(\ell)
\;\asymp\;
\frac{1}{K}\,f(\ell)^{-\kappa_\alpha},
\]
The regime-specific formulas are:
\begin{enumerate}[label=(\roman*),itemsep=2pt]

\item \emph{Logarithmic decay}
  $f(\ell)\asymp c/[\log(1{+}\ell)]^{\vartheta}$ with $\vartheta>0$:
  \[
  \log\!\bigl(1+\mathcal H_{N,K}\bigr)
  \;\asymp\;
  (NK)^{1/(\kappa_\alpha\vartheta)},
  \]
  or equivalently
  \[
  \mathcal H_{N,K}
  =
  \exp\!\left(
  \Theta\!\left((NK)^{1/(\kappa_\alpha\vartheta)}\right)
  \right)-1,
  \]
  giving the strongest asymptotic benefit among the three canonical regimes.

\item \emph{Power-law decay}
  $f(\ell)\asymp c\,\ell^{-\beta}$:
  \;$\displaystyle\mathcal{H}_{N,K}
    \asymp(NK)^{1/(\kappa_\alpha\beta)}
    = K^{1/(\kappa_\alpha\beta)}\,\mathcal{H}_{N,1}$,
  giving a multiplicative gain.

\item \emph{Exponential decay}
  $f(\ell)\asymp c\,\lambda^\ell$:
  \;$\displaystyle\mathcal{H}_{N,K}
    \asymp\frac{\log(NK)}{\kappa_\alpha\log(1/\lambda)}$,
  giving an additive gain
  $\Delta\mathcal{H}\asymp\frac{\log K}{\kappa_\alpha\log(1/\lambda)}$.

\end{enumerate}

\paragraph{Effective number of independent projections.}
The clean $N\to NK$ improvement is a theoretical ceiling: it assumes that the
$K$ projected statistics are independent and sign-aligned at each lag.
Practical implementations realize a smaller benefit, for two reasons.
The projected statistics are computed from the same underlying sequence
gradients, so cross-projection dependence is generally present; and when the
projections are not sign-aligned, partial cancellations in the
matched-statistic computation attenuate the aggregated signal.
Writing $K_{\mathrm{eff}}(\ell)\le K$ for the effective number of independent,
non-cancelling projections, all formulas above hold with $K$ replaced by
$K_{\mathrm{eff}}(\ell)$: $K$ parallel probes behave like a
$K_{\mathrm{eff}}$-fold increase in independent data, with equality only in the ideal case.

\paragraph{Relation to Johnson--Lindenstrauss projections.}
This random-projection average should not be read as an application of
the Johnson--Lindenstrauss lemma~\cite{johnson1984extensions}.
The Johnson--Lindenstrauss result concerns a low-dimensional embedding
that preserves pairwise Euclidean distances in a finite point cloud,
whereas the present construction uses random directions to define
scalar matched-statistic probes and then averages those probes into one
sequence-level statistic.
We therefore do not require, or claim, a Johnson--Lindenstrauss
dimension bound for $K$.
The relevant property is instead the (theoretical) variance-reduction effect of
averaging independent scalar probes, with the $\alpha$-stable
noise-scale reduction captured above by the factor
$K^{-1/\kappa_\alpha}$, or by the corresponding
$K_{\mathrm{eff}}^{-1/\kappa_\alpha}$ under cross-projection
dependence.
The projection-wise diagnostic in
Appendix~\ref{app:projection_alpha_diagnostic} checks empirically that
the directional spread of the tail-index estimates is small in the
present runs.

In practice, multi-projection aggregation provides a second source of
improvement beyond the noise-scale reduction: averaging over directions
stabilizes the alignment coefficient $\overline{m}_{\mu,K}(\ell)$,
reducing projection-specific cancellation.
The empirical validation in
Sec.~\ref{app:projection_validation}, where projection variability is
found to be modest relative to seed variability, provides evidence
that even a small number of projections suffices to capture the
dominant signal.

\subsection{Bootstrap confidence intervals for the
projection-averaged tail index}
\label{app:bootstrap_ci}

For the ECF method, the tail index $\hat\alpha(\ell)$ reported in
Sec.~\ref{sec:exp_results} is a point estimate.
To quantify within-sample uncertainty around that estimate, we attach a
nonparametric bootstrap CI to each per-$(\text{seed},
\ell)$ ECF estimate.
For each $(\text{seed},\ell)$ pair, we bootstrap-resample, with
replacement, the projection-averaged sequence-level scalars over the
diagnostic sequences $B$ times, refit ECF on each resample, and report
the empirical $[2.5\%,\,97.5\%]$ quantile range of the resulting $B$
estimates as the $95\%$ CI on $\hat\alpha(\ell)$.
We use $B=200$ resamples and seed the bootstrap RNG per
$(\text{model},\,\text{seed})$ pair so that different architectures with
the same training seed receive independent resampling streams.
This complements the MCC bootstrap CIs already available from the
training pipeline: ECF point estimates now carry their own within-sample
uncertainty band on the same diagnostic grid.

The CI widths are tight across architectures:
the median 95\% CI width across $(\text{seed},\ell)$ pairs is
$0.0244$ (ConstGate),
$0.0341$ (SharedGate),
$0.0246$ (DiagGate),
$0.0402$ (GRU),
and $0.0174$ (LSTM).
LSTM is the narrowest and GRU the widest, mirroring the directional
spread reported in Sec.~\ref{app:projection_alpha_diagnostic}.
The minimum point estimate per architecture, with its bootstrap CI,
provides the most direct evidence of statistically resolved heavy tails:
ConstGate $1.79$ ($95\%$ CI $[1.77,\,1.81]$), SharedGate $1.19$ ($[1.15,
\,1.22]$), DiagGate $1.87$ ($[1.85,\,1.89]$), GRU $1.64$ ($[1.61,
\,1.67]$), and LSTM $1.77$ ($[1.75,\,1.80]$).
For all five architectures, at least one $(\text{seed},\ell)$ pair has a
$95\%$ CI strictly below $\alpha=2$, so the heavy-tailedness of the
matched statistic is statistically resolved relative to Gaussian for at
least some lags in every architecture.

\subsection{Directional variability of the tail index}
\label{app:projection_alpha_diagnostic}

The tail index $\hat\alpha(\ell)$ reported in
Sec.~\ref{sec:exp_results} is estimated from the projection-averaged
sequence-level matched statistic. The single scalar $\hat\alpha$ used
in the empirical SNR (Eq.~\eqref{eq:empirical_SNR}) is the median of
$\hat\alpha(\ell)$ over diagnostic lags.
To assess whether the underlying directional structure is approximately
isotropic or carries non-trivial direction-to-direction variability, we
additionally compute, for each $(\text{seed},\ell)$ pair, a separate ECF
tail-index estimate for each of the $K=50$ projection directions.
Each per-projection $\hat\alpha$ is itself estimated over the
diagnostic sequences for that fixed direction; the $K$ estimates
therefore form a distribution of tail indices over directions, and
constitute a diagnostic of directional variability rather than
additional independent samples for the projection-averaged tail index
used in the learnability
window.

Table~\ref{tab:projection_alpha} summarizes this distribution per
architecture.
For each $(\text{seed},\ell)$ pair we compute the median,
interquartile range (IQR), and lower decile of the $K$ projection-wise
$\hat\alpha$ values, together with the fractions of projection
directions for which $\hat\alpha<1.9$ (noticeable departure from the
Gaussian boundary) and for which $\hat\alpha<1.85$ (moderate
directional tail weight).
The table reports the cross-$(\text{seed},\ell)$ median of each of
these summaries.
We do not report the projection-wise minimum as a headline number,
since the minimum-of-$K$ statistic has high variance and does not
reflect typical directional structure.

\begin{table}[h]
\centering
\small
\caption{Directional variability of the tail index across $K=50$
projection directions, AdamW run.
For each $(\text{seed},\ell)$ pair, the $K$ per-projection ECF
estimates $\hat\alpha$ are summarized by their median, IQR, lower
decile, and the fractions $\Pr\{\hat\alpha<1.9\}$ and
$\Pr\{\hat\alpha<1.85\}$.
The table reports the cross-$(\text{seed},\ell)$ median of each of
these summaries.
The projection-wise minimum is intentionally not reported, as the
minimum-of-$K$ statistic has high variance and does not represent
typical directional structure.}
\label{tab:projection_alpha}
\begin{tabular}{l ccc cc}
\toprule
Model & median $\hat\alpha$ & IQR & lower decile & $\Pr\{\hat\alpha<1.9\}$ & $\Pr\{\hat\alpha<1.85\}$ \\
\midrule
const  & $1.962$ & $0.017$ & $1.941$ & $0.000$ & $0.000$ \\
shared & $1.934$ & $0.039$ & $1.863$ & $0.250$ & $0.080$ \\
diag   & $1.969$ & $0.058$ & $1.902$ & $0.100$ & $0.000$ \\
gru    & $1.915$ & $0.053$ & $1.849$ & $0.360$ & $0.120$ \\
lstm   & $1.990$ & $0.029$ & $1.948$ & $0.000$ & $0.000$ \\
\bottomrule
\end{tabular}
\end{table}

The IQR over directions is small for every architecture
($0.017$ -- $0.058$), so the $K=50$ projection-wise tail indices are
tightly concentrated for each $(\text{seed},\ell)$ pair: the directional
spread is much smaller than the cross-$(\text{seed},\ell)$ spread of the
median itself.
This pattern is consistent with approximately isotropic directional tail
behavior, in which the $K$-projection averaging that defines the matched
statistic acts mainly as variance reduction over directionally similar
realizations rather than as concentration of mass over an anisotropic
distribution.
The architectural ranking by directional tail weight matches the
projection-averaged ranking reported in
Sec.~\ref{sec:exp_results}:
GRU has the largest fraction of directions with
$\hat\alpha<1.9$ ($0.36$) and $\hat\alpha<1.85$ ($0.12$), then SharedGate
($0.25 / 0.08$), DiagGate ($0.10 / 0.00$), and finally ConstGate and
LSTM ($0.00 / 0.00$).

A useful empirical consistency check.
For the architectures that show genuine departures from the Gaussian
boundary in the projection-averaged analysis (GRU, LSTM, DiagGate), the
projection-averaged tail-index estimates reach values noticeably below
the per-projection medians reported in the table above: e.g.\ the
projection-averaged minima are $\hat\alpha\approx 1.64$ (GRU),
$\hat\alpha\approx 1.77$ (LSTM), and $\hat\alpha\approx 1.87$
(DiagGate), while the corresponding per-projection medians are
$1.915$, $1.990$, and $1.969$.
Averaging across the $K$ projection directions therefore does not push
the matched statistic toward the Gaussian boundary; in several
architectures its tail estimates are as heavy as, or heavier than, the
typical per-direction estimates.
This is consistent with shared within-sequence dependence across
projection directions: $K$-projection averaging provides variance
reduction without forcing the aggregate toward the Gaussian boundary,
and the projection-averaged matched statistic retains the joint tail
weight required by the theory.

\section{Noise floor of the averaged matched statistic}
\label{app:noise_floor}

This appendix states the assumptions behind the noise model of
Sec.~\ref{sec:stat_model} and explains the resulting $N^{-1/\kappa_\alpha}$
concentration rate of the averaged matched statistic.
We stress at the outset that the $\mathcal{S}\alpha\mathcal{S}$ description is
an \emph{effective modeling choice} for the matched-statistic fluctuations,
not a distributional identity derived from the data-generating process; the
detection framework uses only the tail index $\alpha$ and the scale
$\sigma_\alpha(\ell)$.
Throughout, we rely on standard properties of stable laws \cite{nolan2020stable}.

By construction, the averaged matched statistic is the cross-sequence average
$\widehat S_N(\ell)$ defined in Eq.~\eqref{eq:theoretical_cross_sequence_average},
built from the sequence-level averages $\bar S^{(n)}_{\ell}$ in
Eq.~\eqref{eq:theoretical_sequence_average}. Independence across sequences
implies that the sequence-level summands $\bar S^{(n)}_{\ell}$ are independent
for fixed~$\ell$. A raw-signed empirical proxy is obtained with the same
two-stage averaging structure by replacing the oracle sign-oriented summand by
its raw signed analogue, as discussed in the experiments.

Within each sequence, by contrast, the summands $S^{(n)}_{t,\ell}$ are not
independent: they share hidden states, parameters, and overlapping computation graphs.
The detection framework does not require within-sequence independence; it
requires only that the sequence-level averages $\bar{S}^{(n)}_\ell$, each
aggregating $T{-}\ell$ temporally dependent terms, be well approximated, across
sequences, by independent draws from a common heavy-tailed law with stable index~$\alpha$.
This is a modeling condition on the marginal behavior of the sequence-level statistic, and it is reasonable because each $\bar S^{(n)}_\ell$ aggregates many dependent contributions.
When this within-sequence dependence is sufficiently mixing, the aggregate is dominated by its heaviest contributions and is well-summarized by a single stable index, and the cross-sequence step then proceeds under the usual independence assumption.

The practical statistic supports this choice.
While the theoretical matched statistic in Eq.~\eqref{eq:matched_statistic}
involves nonlinear operations (the oracle sign factors
$\operatorname{sgn}(m_q(\ell))$), the empirical statistic
$\widetilde{S}_{t,\ell}$ in Eq.~\eqref{eq:empirical_raw_matched_statistic} is a
purely linear function of the gradient: a weighted sum over neurons with
weights given by the generalized effective learning rates $\mu^{(q)}_{t,\ell}$.
Since it aggregates gradient information linearly, and gradient fluctuations in
deep networks are empirically heavy-tailed (Sec.~\ref{sec:stat_model}), we
model its centered sequence-level version
$\bar S^{(n)}_{\ell}-\mathbb{E}[\bar S^{(n)}_{\ell}]$ as a heavy-tailed
$\mathcal{S}\alpha\mathcal{S}$ variate with lag-dependent scale parameter
$\sigma_\alpha(\ell)$.
The framework depends only on $\alpha$ and $\sigma_\alpha(\ell)$, not on the mechanism generating the heavy tails, and does not assume exact $\mathcal{S}\alpha\mathcal{S}$ behavior at finite~$N$; the stable family is used only as an analytically tractable summary that nests the Gaussian case at $\alpha=2$.

The resulting noise-floor model for the averaged matched statistic
$\widehat S_N(\ell)$ is
\begin{equation}
\label{eq:averaged_matched_statistic_noise_floor}
\widehat S_N(\ell)-\mathbb{E}[\widehat S_N(\ell)]
\overset{\mathrm{model}}{\sim}
\mathcal{S}\alpha\mathcal{S}\!\left(0,\sigma_\alpha(\ell) N^{-1/\kappa_\alpha}\right)
=
\mathcal{S}\alpha\mathcal{S}\!\left(0,\frac{\sigma_\alpha(\ell)}{N^{1/\kappa_\alpha}}\right),
\end{equation}
where $\kappa_\alpha$ is defined in Eq.~\eqref{eq:kappa_alpha_def}.
This display should be read as the canonical stable surrogate implied by the
modeling assumption above, not as an exact finite-sample identity.
The $N^{1/\alpha}$ scaling applies to sums of $\alpha$-stable variables,
whereas the empirical average introduces an additional factor $1/N$, yielding
the rate $N^{1/\alpha-1}=N^{-1/\kappa_\alpha}$ by
Eq.~\eqref{eq:kappa_alpha_def}.
Thus, the typical magnitude of stochastic fluctuations decays as
$N^{-1/\kappa_\alpha}$.
For $\alpha=2$ this recovers the classical Gaussian rate $N^{-1/2}$, while for
$\alpha<2$ concentration is slower due to heavy-tailed noise.
This noise floor determines the detectability thresholds and sample-complexity
bounds derived in the manuscript and underlies the compression of the
learnability window in the heavy-tailed regime.

\section{Proofs of Lemmas}
\label{app:lemma_proofs}

\paragraph{Proof of Lemma~\ref{lem:mu_l1_monotone}}
\begin{proof}

\textit{Part~(i): zeroth-order monotonicity.}\quad
For all architectures considered in this paper, the zeroth-order
transport terms are products of gate activations and bounded activation
derivatives, hence they have the generic form
\[
\gamma^{(0,q)}_{t,\ell}
=
\prod_{k=1}^{\ell} a^{(q)}_{t-k+1},
\qquad
a^{(q)}_{t-k+1}\in[0,1].
\]
Increasing the lag from $\ell$ to $\ell+1$ appends one additional
factor in $[0,1]$:
\[
|\gamma^{(0,q)}_{t,\ell+1}|
=
|\gamma^{(0,q)}_{t,\ell}|\cdot a^{(q)}_{t-\ell}
\le
|\gamma^{(0,q)}_{t,\ell}|.
\]
For the GRU, the auxiliary envelopes $\rho^{(0,q)}_{t,\ell}$ and
$\eta^{(0,q)}_{t,\ell}$ share this multiplicative structure and the
same argument applies.
Since each summand is nonincreasing, the zeroth-order proxy
$f_0(\ell)=\sum_q|\gamma^{(0,q)}_{t,\ell}|$ is nonincreasing.

\medskip
\textit{Part~(ii): first-order comparability.}\quad
The first-order diagonal correction $\gamma^{(1,q)}_{t,\ell}$ is a
finite sum of insertion terms, each obtained by placing a single
recurrent mixing contribution into an otherwise diagonal transport
chain (Appendix~\ref{app:first_order_expansion}).
Unlike the zeroth-order case, increasing the lag both
contracts existing summands (through additional gate factors in
$[0,1]$) and extends the summation range (through a new insertion
site).
The new summand involves recurrent-weight contributions whose
diagonal entries need not be sign-controlled.
Consequently, pointwise monotonicity of the absolute value
$|\gamma^{(1,q)}_{t,\ell}|$ cannot be deduced from gate
boundedness alone.

We therefore assume the dominance condition: there exists $c\in[0,1)$ such that
\begin{equation}
\label{eq:first_order_dominance}
|\gamma^{(1,q)}_{t,\ell}|\le c|\gamma^{(0,q)}_{t,\ell}| \qquad\text{for all relevant } q,\;\ell.
\end{equation}
Under~\eqref{eq:first_order_dominance},
\[
(1-c)\,|\gamma^{(0,q)}_{t,\ell}|
\;\le\;
|\gamma^{(0,q)}_{t,\ell}|-|\gamma^{(1,q)}_{t,\ell}|
\;\le\;
|\Gamma^{(q)}_{t,\ell}|
\;\le\;
|\gamma^{(0,q)}_{t,\ell}|+|\gamma^{(1,q)}_{t,\ell}|
\;\le\;
(1+c)\,|\gamma^{(0,q)}_{t,\ell}|,
\]
establishing that $|\Gamma^{(q)}_{t,\ell}|$ is comparable to the
nonincreasing zeroth-order term.

\medskip
\textit{Part~(iii): envelope comparability.}\quad
By Section~\ref{sec:generalized_elr}, the generalized effective
learning rates factorize as
$\mu^{(q)}_{t,\ell}=\Lambda^{(q)}_{r,\ell}\,\Gamma^{(q)}_{t,\ell}$,
with $\Lambda^{(q)}_{r,\ell}>0$ given by the Rayleigh quotient.
Let $\underline{\Lambda}=\min_i\lambda_{i,r}$ and
$\overline{\Lambda}=\max_i\lambda_{i,r}$ denote, respectively, the
lower and upper bound factors appearing in
Eq.~\eqref{eq:rayleigh_bound}, so that
$\underline{\Lambda}\le\Lambda^{(q)}_{r,\ell}\le\overline{\Lambda}$
for every $q$ and every $\ell$.
Since $\Lambda^{(q)}_{r,\ell}>0$, we have
$|\mu^{(q)}_{t,\ell}|=\Lambda^{(q)}_{r,\ell}\,|\Gamma^{(q)}_{t,\ell}|$;
combining this identity with the part~(ii) bound and summing over~$q$
yields
\[
\underline{\Lambda}\,(1-c)\,f_0(\ell)
\;\le\;
f(\ell)
\;\le\;
\overline{\Lambda}\,(1+c)\,f_0(\ell),
\]
where $f_0(\ell)=\sum_q|\gamma^{(0,q)}_{t,\ell}|$ is nonincreasing by
part~(i).
This establishes Eq.~\eqref{eq:envelope_comparability}.
\end{proof}

\paragraph{On the dominance condition.}
Condition~\eqref{eq:first_order_dominance} formalizes the requirement
that the first-order diagonal corrections remain perturbative.
By construction, $\gamma^{(1,q)}_{t,\ell}$ arises from the
single-insertion terms of the Fr\'{e}chet product
rule~\eqref{eq:first-order-expansion}, and its magnitude is controlled
by the norm of the off-diagonal (recurrent mixing) components of the
one-step Jacobian relative to the diagonal transport.
The companion numerical analysis in~\cite{livi2025timescale} confirms
that, across all architectures and training conditions considered, the
first-order approximation accounts for the dominant contribution to the
diagonal Jacobian-product entries.
The envelope validation in
Appendix~\ref{app:envelope_validation}, where the first-order envelope
$f(\ell)$ tracks the full-matrix envelope
$f_{\mathcal{M}}(\ell)$ with Spearman $\rho\ge0.972$, provides
further empirical support for the dominance regime.

\paragraph{Proof of Lemma~\ref{lem:horizon_scaling_alpha}}

\begin{proof}
Starting from the per-lag sample-complexity threshold in
Eq.~\eqref{eq:min_N_of_ell_for_detection},
\[
N(\ell)
=
d_{\alpha,\epsilon}^{\kappa_\alpha}\,
\left(
\frac{\sigma_\alpha(\ell)}
{|\overline{m}_\mu(\ell)|\,f(\ell)}
\right)^{\!\kappa_\alpha},
\]
and applying the boundedness assumptions
\(c_\sigma \le \sigma_\alpha(\ell)\le C_\sigma\) and
\(c_m\le |\overline{m}_\mu(\ell)|\le C_m\),
we obtain
\[
d_{\alpha,\epsilon}^{\kappa_\alpha}
\left(\frac{c_\sigma}{C_m}\right)^{\!\kappa_\alpha}\,
f(\ell)^{-\kappa_\alpha}
\ \le\
N(\ell)
\ \le\
d_{\alpha,\epsilon}^{\kappa_\alpha}
\left(\frac{C_\sigma}{c_m}\right)^{\!\kappa_\alpha}\,
f(\ell)^{-\kappa_\alpha},
\]
which yields Eq.~\eqref{eq:N_of_ell_sandwich_alpha} with
\(c_\star =
d_{\alpha,\epsilon}^{\kappa_\alpha}(c_\sigma/C_m)^{\kappa_\alpha}\)
and
\(C_\star =
d_{\alpha,\epsilon}^{\kappa_\alpha}(C_\sigma/c_m)^{\kappa_\alpha}\).

\medskip
\noindent
For the horizon bound, recall the detectability condition from
Eq.~\eqref{eq:theoretical_signal_threshold}:
\[
f(\ell)
\ge
\frac{
d_{\alpha,\epsilon}\,\sigma_\alpha(\ell)
}{
N^{1/\kappa_\alpha}\,|\overline{m}_\mu(\ell)|
}.
\]
The constant factor $d_{\alpha,\epsilon}$ is independent of $\ell$ and
$N$ and can be absorbed into the constants of the sandwich bound.

Bounding the right-hand side above and below using the assumed bounds on
$\sigma_\alpha(\ell)$ and $|\overline{m}_\mu(\ell)|$ gives
\[
d_{\alpha,\epsilon}\frac{c_\sigma}{C_m}\,N^{-1/\kappa_\alpha}
\ \le\
\frac{
d_{\alpha,\epsilon}\,\sigma_\alpha(\ell)
}{
N^{1/\kappa_\alpha}\,|\overline{m}_\mu(\ell)|
}
\ \le\
d_{\alpha,\epsilon}\frac{C_\sigma}{c_m}\,N^{-1/\kappa_\alpha}.
\]
The detectability condition is guaranteed whenever
\[
f(\ell)\ge
d_{\alpha,\epsilon}\tfrac{C_\sigma}{c_m}\,N^{-1/\kappa_\alpha},
\]
and cannot hold unless
\[
f(\ell)\ge
d_{\alpha,\epsilon}\tfrac{c_\sigma}{C_m}\,N^{-1/\kappa_\alpha}.
\]

By Lemma~\ref{lem:mu_l1_monotone}, $f(\ell)$ is comparable to the
nonincreasing proxy $f_0(\ell)$:
\[
  \underline{\Lambda}(1-c)\,f_0(\ell)
  \;\le\;
  f(\ell)
  \;\le\;
  \overline{\Lambda}(1+c)\,f_0(\ell).
\]
The comparability holds uniformly over the lag range under
consideration, as ensured by the dominance
condition~\eqref{eq:first_order_dominance} and the boundedness of the
adaptive base rates.
Since $f_0$ is nonincreasing,
$f_0^{\leftarrow}(y)=\sup\{\ell\ge1:f_0(\ell)\ge y\}$ is well defined.
From the comparability, the set
$\{\ell:f(\ell)\ge y\}$ is controlled by
$\{\ell:f_0(\ell)\ge y/[\overline{\Lambda}(1+c)]\}$ from above and
$\{\ell:f_0(\ell)\ge y/[\underline{\Lambda}(1-c)]\}$ from below,
each of which is an interval $[1,f_0^{\leftarrow}(\cdots)]$.
Applying this with
$y=d_{\alpha,\epsilon}\tfrac{C_\sigma}{c_m}N^{-1/\kappa_\alpha}$ and
$y=d_{\alpha,\epsilon}\tfrac{c_\sigma}{C_m}N^{-1/\kappa_\alpha}$
respectively and
absorbing all multiplicative factors into the constants
$\widetilde{C}_1,\widetilde{C}_2$ yields
\[
f_0^{\leftarrow}\!\Big(\widetilde{C}_1\,N^{-1/\kappa_\alpha}\Big)
\ \le\
\mathcal{H}_N
\ \le\
f_0^{\leftarrow}\!\Big(\widetilde{C}_2\,N^{-1/\kappa_\alpha}\Big),
\]
which establishes the claimed sandwich bound.

\paragraph{Generalized inverse.}
The generalized inverse is applied to the monotone proxy~$f_0$ rather than to $f$ itself.
Under the comparability bound~\eqref{eq:envelope_comparability}, $f$ and $f_0$ share the same decay regime (exponential, power-law, or logarithmic), so $f_0^{\leftarrow}$ and $f^{\leftarrow}$ yield the same asymptotic scaling of $\mathcal{H}_N$ up to multiplicative constants.
\end{proof}

The proof shows that the dependence on stochasticity and alignment enters only
through the constants
$(c_\sigma,C_\sigma,c_m,C_m)$, the comparability factors
$(\underline{\Lambda},\overline{\Lambda},c)$, and the tail index $\alpha$ (via
$\kappa_\alpha$); the functional dependence of $N(\ell)$ and $\mathcal{H}_N$ on
the envelope $f(\ell)$ is dictated entirely by the power
$f(\ell)^{-\kappa_\alpha}$.

\paragraph{On the boundedness assumptions.}
The conditions \(c_\sigma \le \sigma_\alpha(\ell)\le C_\sigma\) and
\(c_m\le |\overline{m}_\mu(\ell)|\le C_m\) should be understood as
mild regularity assumptions over the finite range of lags under
consideration.
Absent pathological degeneracies, such as vanishing alignment magnitude
or diverging fluctuation scale $\sigma_\alpha(\ell)\to\infty$, these
quantities remain finite and vary smoothly with~$\ell$.
Thus, the boundedness assumptions affect only constant factors in Eq.~\eqref{eq:N_of_ell_sandwich_alpha}, while the asymptotic dependence of $N(\ell)$ and $\mathcal{H}_N$ on the envelope $f(\ell)$ is entirely captured by the exponent $\kappa_\alpha$.

\section{Asymptotic Scaling of the Learnability Window under $\alpha$-Stable Noise}
\label{app:scaling_derivations}

This appendix provides the detailed derivations supporting the scaling relations summarized in Sec.~\ref{sec:scaling_laws}.
Throughout, we use the concentration exponent $\kappa_\alpha$ defined in Eq.~\eqref{eq:kappa_alpha_def}, which governs the statistical behavior of empirical averages under $\alpha$-stable noise.
Starting from the $\alpha$-stable model of Eq.~\eqref{eq:alpha_stable_model} and the finite-sample requirement of Eq.~\eqref{eq:min_N_of_ell_for_detection}, we show how the slow concentration rate characteristic of heavy-tailed averages determines the asymptotic growth of the learnability window~$\mathcal{H}_N$.

\paragraph{Per-lag sample-complexity threshold.}
From Eq.~\eqref{eq:min_N_of_ell_for_detection}, the Bayes-error
sample-complexity threshold for detecting a lag-$\ell$ dependency with
target Bayes error level~\(\epsilon\) is
\begin{equation}
\label{eq:app_N_of_ell_alpha}
N(\ell)
\;=\;
d_{\alpha,\epsilon}^{\kappa_\alpha}\,
\left(
\frac{\sigma_\alpha(\ell)}
{|\overline{m}_\mu(\ell)| f(\ell)}
\right)^{\!\kappa_\alpha}.
\end{equation}
The factor $d_{\alpha,\epsilon}^{\kappa_\alpha}$ depends only on the
tail index~$\alpha$ and the target detection error~$\epsilon$, and can
be absorbed into the asymptotic constants below.
We assume the boundedness conditions of
Lemma~\ref{lem:horizon_scaling_alpha},
\[
c_\sigma \le \sigma_\alpha(\ell) \le C_\sigma,
\qquad
c_m \le |\overline{m}_\mu(\ell)| \le C_m,
\]
and recall that $f(\ell)$ is comparable to the nonincreasing proxy
$f_0(\ell)$ (Lemma~\ref{lem:mu_l1_monotone}),
so that the generalized inverse is applied to $f_0$.

\paragraph{Two-sided sandwich bound.}
Substituting these bounds into~\eqref{eq:app_N_of_ell_alpha} yields constants
\[
c_\star =
d_{\alpha,\epsilon}^{\kappa_\alpha}
\Big(\tfrac{c_\sigma}{C_m}\Big)^{\!\kappa_\alpha},
\qquad
C_\star =
d_{\alpha,\epsilon}^{\kappa_\alpha}
\Big(\tfrac{C_\sigma}{c_m}\Big)^{\!\kappa_\alpha},
\]
such that
\begin{equation}
\label{eq:app_N_sandwich_alpha}
c_\star\,f(\ell)^{-\kappa_\alpha}
\ \le\
N(\ell)
\ \le\
C_\star\,f(\ell)^{-\kappa_\alpha}.
\end{equation}
Hence,
\begin{equation}
\label{eq:master_proportionality_law}
N(\ell)\asymp f(\ell)^{-\kappa_\alpha}
\end{equation}
gives the \emph{master proportionality relation} between the sample-complexity threshold and the envelope, formalizing that longer dependencies (smaller $f(\ell)$ values) require increasingly large sample sizes to detect under heavy-tailed noise.
We use the symbol $\asymp$ rather than $\propto$ because the preceding two-sided sandwich bound determines this dependence only up to multiplicative constants.

\paragraph{Interpreting the master proportionality empirically.}
Equation~\eqref{eq:master_proportionality_law} is a structural statement
about the scaling exponent relating sample complexity to envelope
decay, not a pointwise formula for predicting~$N(\ell)$ from~$f(\ell)$.
In practice, the mapping $f(\ell)\mapsto N(\ell)$ is mediated by
quantities that are themselves expensive to estimate from finite data:
the tail index~$\alpha(\ell)$, which varies with the lag and whose
estimation from heavy-tailed samples carries wide confidence intervals
even at moderate sample sizes; the scale parameter~$\sigma_\alpha(\ell)$;
and the lag-dependent alignment magnitude~$|\overline{m}_\mu(\ell)|$.
Small errors in~$\hat\alpha(\ell)$ are amplified through
$\kappa_\alpha=\alpha/(\alpha-1)$, which diverges as $\alpha\to 1$,
so that even a~$10\%$ relative error in the tail index can produce a
several-fold shift in the predicted sample complexity.
Moreover, $N(\ell)$ is determined empirically from a discrete grid of
candidate sample sizes via a detection threshold, introducing additional
quantization variability.
For these reasons, scatter plots of empirical $\hat{N}(\ell)$
versus~$\hat{f}(\ell)$ will typically exhibit substantial dispersion
around the predicted power law, particularly at large lags where both
the envelope and the tail-index estimates are least reliable.
The master proportionality should therefore be read as identifying the
correct functional form of the dependence and, through the
exponent~$\kappa_\alpha$, quantifying the penalty that heavy-tailed noise
imposes on sample requirements, rather than as a practical prediction
rule with low residual variance.

\paragraph{Learnability window as a level set.}
From the finite-sample detectability condition of
Eq.~\eqref{eq:theoretical_signal_threshold},
\[
f(\ell)
\ \ge\
\frac{
d_{\alpha,\epsilon}\,\sigma_\alpha(\ell)
}{
N^{1/\kappa_\alpha}\,|\overline{m}_\mu(\ell)|
},
\]
and using the same bounding arguments together with the envelope
comparability~\eqref{eq:envelope_comparability}, we can express the
learnability window $\mathcal{H}_N$ as a level set of the monotone
proxy~$f_0(\ell)$:
\begin{equation}
\label{eq:app_H_window_alpha}
f_0^{\leftarrow}\!\Big(\widetilde{C}_1\,N^{-1/\kappa_\alpha}\Big)
\ \le\
\mathcal{H}_N
\ \le\
f_0^{\leftarrow}\!\Big(\widetilde{C}_2\,N^{-1/\kappa_\alpha}\Big),
\end{equation}
where $f_0^{\leftarrow}(y)=\sup\{\ell\ge1: f_0(\ell)\ge y\}$ denotes the generalized
inverse of the nonincreasing proxy~$f_0$.
By Lemma~\ref{lem:mu_l1_monotone}, $f(\ell)$ is comparable to the
nonincreasing proxy $f_0(\ell)$, and the generalized inverse
$f_0^{\leftarrow}$ is well defined on $f_0$.
The comparability constants are absorbed into the level-set bounds
$\widetilde{C}_1,\widetilde{C}_2$.
This reproduces the sandwich bound in
Eq.~\eqref{eq:H_window_sandwich_alpha} of the main text.

\paragraph{Asymptotic regimes.}
For each canonical decay law of $f(\ell)$, the scaling analysis follows a two-step pipeline that links the envelope, the sample-complexity requirement, and the learnability window.
In the \emph{forward step}, we substitute the envelope form into the master proportionality~\eqref{eq:master_proportionality_law} to obtain \(N(\ell)\asymp f(\ell)^{-\kappa_\alpha}\) as an explicit function of~$\ell$.
In the \emph{inverse step}, we invoke the level-set bound~\eqref{eq:app_H_window_alpha} and solve for the largest $\ell$ compatible with a given budget~$N$, yielding $\mathcal{H}_N$ as an explicit function of~$N$.
Multiplicative factors from the sandwich bounds, the envelope comparability~\eqref{eq:envelope_comparability}, and the tail-index factor~$d_{\alpha,\epsilon}^{\kappa_\alpha}$ are absorbed into generic positive constants throughout.

\paragraph{(i) Logarithmic decay.}
Let $f(\ell)\asymp c/[\log(1+\ell)]^{\vartheta}$ with $c>0$ and $\vartheta>0$.
The forward step yields
\[
N(\ell)\;\asymp\; f(\ell)^{-\kappa_\alpha}
\;\asymp\;
c^{-\kappa_\alpha}[\log(1+\ell)]^{\kappa_\alpha\vartheta}
\;\asymp\;
[\log(1+\ell)]^{\kappa_\alpha\vartheta}.
\]
Solving for $\log(1+\ell)$ and invoking Eq.~\eqref{eq:app_H_window_alpha}, the inverse step gives
\begin{equation}
\label{eq:logarithmic_window_log_scale}
\log\!\bigl(1+\mathcal H_N\bigr)
\;\asymp\;
c^{1/\vartheta}\,
N^{1/(\kappa_\alpha\vartheta)}.
\end{equation}
Equivalently, there exist constants
$0<C_{\log}^{-}\le C_{\log}^{+}<\infty$ such that
\begin{equation}
\label{eq:logarithmic_window_bounds}
\exp\!\left(
C_{\log}^{-}c^{1/\vartheta}
N^{1/(\kappa_\alpha\vartheta)}
\right)-1
\;\le\;
\mathcal H_N
\;\le\;
\exp\!\left(
C_{\log}^{+}c^{1/\vartheta}
N^{1/(\kappa_\alpha\vartheta)}
\right)-1.
\end{equation}
The constants $C_{\log}^{-}$ and $C_{\log}^{+}$ absorb the comparability, alignment, noise-scale, and detection-threshold factors entering the sandwich bounds of Lemmas~\ref{lem:mu_l1_monotone}--\ref{lem:horizon_scaling_alpha}.
Thus, at the level of exponential order,
\begin{equation}
\label{eq:logarithmic_window_theta}
\mathcal H_N
=
\exp\!\left(
\Theta\!\left(
N^{1/(\kappa_\alpha\vartheta)}
\right)
\right)-1.
\end{equation}
Unlike multiplicative prefactors in the power-law and exponential regimes, the lower and upper constants in Eq.~\eqref{eq:logarithmic_window_bounds} appear inside the exponential and therefore cannot be collapsed into a single multiplicative $\asymp$ relation for $\mathcal H_N$.

When $\kappa_\alpha\vartheta>1$, the exponent $1/(\kappa_\alpha\vartheta)$ lies in $(0,1)$, and Eq.~\eqref{eq:logarithmic_window_theta} describes stretched-exponential growth in~$N$: the learnability window grows faster than any polynomial but slower than an ordinary exponential.
In particular, the basic case $\vartheta=1$ satisfies this condition for every $\alpha\in(1,2]$, since $\kappa_\alpha\ge2$.
If $\kappa_\alpha\vartheta=1$, the growth is exponential in~$N$; if $\kappa_\alpha\vartheta<1$, it is faster than exponential.
These distinctions refine the logarithmic-envelope regime without altering the level-set relation in Eq.~\eqref{eq:logarithmic_window_log_scale}.

Finally, the additive shift inside the logarithm ensures that $f(\ell)$ is finite and strictly positive at the smallest admissible lag $\ell=1$, where $\log\ell=0$ would make the envelope singular.
Replacing $\log(1+\ell)$ by $\log(c'+\ell)$ for any fixed $c'>0$ does not change the large-lag scaling, because $\log(c'+\ell)\sim\log\ell$ as $\ell\to\infty$.

\paragraph{(ii) Power-law decay.}
Let $f(\ell)\asymp c\,\ell^{-\beta}$ with $c>0$ and $\beta>0$.
The forward step yields
\[
N(\ell)\;\asymp\; f(\ell)^{-\kappa_\alpha}
\;\asymp\;
c^{-\kappa_\alpha}\ell^{\kappa_\alpha\beta}
\;\asymp\;
\ell^{\kappa_\alpha\beta}.
\]
Solving for $\ell$ and invoking~\eqref{eq:app_H_window_alpha},
the inverse step gives
\[
\ell\;\asymp\; N^{1/(\kappa_\alpha\beta)}
\quad\Longrightarrow\quad
\mathcal{H}_N\;\asymp\; N^{1/(\kappa_\alpha\beta)}.
\]

\paragraph{(iii) Exponential (geometric) decay.}
Let $f(\ell)\asymp c\,\lambda^{\ell}$ with $c>0$ and $\lambda\in(0,1)$.
The forward step yields
\[
N(\ell)\;\asymp\; f(\ell)^{-\kappa_\alpha}
\;\asymp\;
c^{-\kappa_\alpha}\lambda^{-\kappa_\alpha\ell}
\;\asymp\;
\lambda^{-\kappa_\alpha\ell}.
\]
Taking logarithms and invoking~\eqref{eq:app_H_window_alpha}, the inverse step gives
\[
\ell
\;\asymp\;
\frac{\log N+\kappa_\alpha\log c}{\kappa_\alpha\,\log(1/\lambda)}
\;\asymp\;
\frac{\log N}{\kappa_\alpha\,\log(1/\lambda)}
\quad\Longrightarrow\quad
\mathcal{H}_N\;\asymp\;\frac{\log N}{\kappa_\alpha\,\log(1/\lambda)}.
\]

The constants inherited from the envelope prefactor $c$, the bounds $c_\sigma$, $C_\sigma$, $c_m$, $C_m$, the detection factor $d_{\alpha,\epsilon}^{\kappa_\alpha}$, and the first-order dominance condition affect the three inverse scaling laws differently.
For the power-law case, they produce multiplicative constants in $\mathcal{H}_N\asymp N^{1/(\kappa_\alpha\beta)}$.
For the exponential case, they produce additive shifts after taking logarithms and therefore do not change the leading $\log N$ behavior.
For the logarithmic-envelope case, they determine the distinct lower and upper coefficients inside the exponential bounds in Eq.~\eqref{eq:logarithmic_window_bounds}.
This is why the corresponding order is stated through $\log(1+\mathcal H_N)\asymp N^{1/(\kappa_\alpha\vartheta)}$ rather than through a multiplicative $\asymp$ relation for $\mathcal{H}_N$.

\clearpage
\section{Vanishing and exploding gradients as extreme envelope regimes}
\label{app:vanishing_exploding}

This appendix derives the envelope bounds summarized in
paragraph~(v) of Sec.~\ref{sec:learnability_theoretical_implications},
showing that the vanishing-gradient regime is the fast-exponential
corner of the envelope taxonomy and discussing the complementary
exploding-gradient case.

\paragraph{Setup: strong Jacobian contraction.}
The pathological vanishing-gradient regime corresponds to recurrent
dynamics that contract strongly and uniformly: each one-step state
Jacobian $J_j$ satisfies
\begin{equation}
\label{eq:vanish_jacobian_contraction}
  \|J_j\|_2 \;\le\; \lambda, \qquad \lambda\in(0,1),
\end{equation}
with $\lambda$ bounded appreciably away from~$1$; smaller~$\lambda$
means faster contraction and a more severe pathology.
Here $\|\cdot\|_2$ is the spectral norm (largest singular value).
By submultiplicativity, the exact BPTT transport over a lag of $\ell$
steps inherits the contraction,
\begin{equation}
  \|\mathcal{M}_{t,\ell}\|_2
  \;\le\;
  \prod_{j=t-\ell+1}^{t}\|J_j\|_2
  \;\le\;
  \lambda^{\ell},
\end{equation}
and, since $|e_a^\top M e_b|\le\|M\|_2$ for any matrix $M$ and unit
vectors $e_a,e_b$ (Cauchy--Schwarz), every exact entry of
$\mathcal{M}_{t,\ell}$ decays at least as fast,
\begin{equation}
\label{eq:vanish_entry_bound}
  |e_a^\top \mathcal{M}_{t,\ell}\,e_b|
  \;\le\;
  \|\mathcal{M}_{t,\ell}\|_2
  \;\le\;
  \lambda^{\ell}.
\end{equation}

\paragraph{The gap: the envelope is built from the retained transport.}
The transport factor $\Gamma^{(q)}_{t,\ell}$ of
Eq.~\eqref{eq:transport_factor} is \emph{not} an exact entry of
$\mathcal{M}_{t,\ell}$.
It is assembled from the first-order expansion of the Jacobian product
(Appendix~\ref{app:RNNs_jacobians_and_elr}), which decomposes each
one-step Jacobian as
\begin{equation}
  J_j = \mathcal{T}_j + \mathcal{R}_j,
\end{equation}
a zeroth-order (gate-diagonal) transport operator $\mathcal{T}_j$ plus a recurrent-mixing correction $\mathcal{R}_j$, and retains the diagonal of the resulting product.
The zeroth-order transport factor is therefore an entry of the retained product $\mathcal{T}_{t,\ell}=\prod_{j=t-\ell+1}^{t}\mathcal{T}_j$, not of $\mathcal{M}_{t,\ell}$.
Consequently, the entry bound~\eqref{eq:vanish_entry_bound} does not by itself control $\gamma^{(0,q)}_{t,\ell}$. The contraction of $\mathcal{M}_{t,\ell}$ could in principle arise from cancellation between $\mathcal{T}_j$ and $\mathcal{R}_j$, leaving $\mathcal{T}_j$ itself non-contractive.
In general,
\begin{equation}
  \|J_j\|_2\le\lambda
  \quad\not\Longrightarrow\quad
  \|\mathcal{T}_j\|_2\le\lambda .
\end{equation}

\paragraph{Compatibility condition (no contraction by cancellation).}
The effective learning rates and hence the envelope are derived in the perturbative regime in which the recurrent mixing is a genuine correction to the retained transport, i.e.\ contraction is visible in $\mathcal{T}_j$ rather than produced by cancellation with $\mathcal{R}_j$.
We state this explicitly as the following condition
\begin{equation}
\label{eq:vanish_compat}
  \|\mathcal{R}_j\|_2 \;\le\; \eta\,\|\mathcal{T}_j\|_2,
  \qquad \eta\in[0,1).
\end{equation}
We further assume the contraction is strong enough (i.e. $\lambda$ is sensibly below 1) that the induced exponential rate $\lambda_{\mathrm{env}}$ of the envelope decay satisfies
\begin{equation}
  \lambda_{\mathrm{env}} = \frac{\lambda}{1-\eta} \;<\; 1 .
\end{equation}

\paragraph{Bridge: contraction transfers to the retained transport.}
Under~\eqref{eq:vanish_jacobian_contraction} and~\eqref{eq:vanish_compat}, the triangle inequality gives
\begin{equation}
  \|\mathcal{T}_j\|_2
  =
  \|J_j-\mathcal{R}_j\|_2
  \;\le\;
  \|J_j\|_2 + \|\mathcal{R}_j\|_2
  \;\le\;
  \lambda + \eta\,\|\mathcal{T}_j\|_2,
\end{equation}
and therefore
\[
  (1-\eta)\,\|\mathcal{T}_j\|_2 \;\le\; \lambda .
\]
Since $\eta<1$, division by $1-\eta$ gives
\[
  \|\mathcal{T}_j\|_2
  \;\le\;
  \frac{\lambda}{1-\eta}
  =
  \lambda_{\mathrm{env}} .
\]
By sub-multiplicativity,
\begin{equation}
\label{eq:vanish_retained_product}
  \|\mathcal{T}_{t,\ell}\|_2
  \;\le\;
  \prod_{j=t-\ell+1}^{t}\|\mathcal{T}_j\|_2
  \;\le\;
  \lambda_{\mathrm{env}}^{\ell}.
\end{equation}

\paragraph{Zeroth-order envelope bound (retained paths).}
Recall that $\Gamma^{(q)}_{t,\ell}$ denotes the neuronwise transport factor entering the generalized effective learning rate $\mu^{(q)}_{t,\ell}=\Lambda^{(q)}_{r,\ell}\Gamma^{(q)}_{t,\ell}$ (Eq.~\eqref{eq:transport_factor}). Depending on the architecture, the zeroth-order part of $\Gamma^{(q)}_{t,\ell}$ may consist of one or several retained diagonal transport paths; a single path for the LSTM and the baseline gated RNNs, and the update, reset, and update--reset paths for the GRU.
To make this path decomposition explicit only in this appendix, let $\mathcal{P}_q$ denote the finite set of retained zeroth-order paths associated with neuron $q$, and let $\gamma^{(0,p,q)}_{t,\ell},\ p\in\mathcal{P}_q$, denote the zeroth-order scalar contribution of path $p$ for neuron $q$.
Thus, the notation $\gamma^{(0,p,q)}_{t,\ell}$ refines the architecture-level zeroth-order term $\gamma^{(0,q)}_{t,\ell}$ used in Appendix~\ref{app:RNNs_jacobians_and_elr}: when there is only one retained path, $\gamma^{(0,p,q)}_{t,\ell}=\gamma^{(0,q)}_{t,\ell}$; for the GRU, the paths correspond to $\gamma^{(0,q)}_{t,\ell}$, $\rho^{(0,q)}_{t,\ell}$, and $\eta^{(0,q)}_{t,\ell}$.

Each path $p\in\mathcal{P}_q$ is a bounded endpoint projection of a retained product,
\begin{equation}
  \gamma^{(0,p,q)}_{t,\ell}
  =
  e_a^\top B^{(p)}_t\,\mathcal{T}^{(p)}_{t,\ell}\,e_b,
  \qquad
  \mathcal{T}^{(p)}_{t,\ell}
  =
  \prod_{j=t-\ell+1}^{t}\mathcal{T}^{(p)}_j,
\end{equation}
with a bounded endpoint factor $\|B^{(p)}_t\|_2\le C_0$ (the identity for
the baseline gated RNNs and the GRU, the cell-readout factor $E_t$ for
the LSTM). Under the compatibility
condition~\eqref{eq:vanish_compat} applied to each path, where
$\mathcal{R}^{(p)}_j$ denotes the residual obtained after isolating
the retained one-step operator $\mathcal{T}^{(p)}_j$ from $J_j$,
$\|\mathcal{R}^{(p)}_j\|_2\le\eta\|\mathcal{T}^{(p)}_j\|_2$ with
$\lambda_{\mathrm{env}}=\lambda/(1-\eta)<1$, the bridge above gives
$\|\mathcal{T}^{(p)}_{t,\ell}\|_2\le\lambda_{\mathrm{env}}^{\ell}$, and by
$|e_a^\top M e_b|\le\|M\|_2$,
\begin{equation}
\label{eq:vanish_gamma0_bound}
  |\gamma^{(0,p,q)}_{t,\ell}|
  \;\le\;
  C_0\,\lambda_{\mathrm{env}}^{\ell}.
\end{equation}
Since $|\mathcal{P}_q|$ is finite and architecture-dependent, the total
retained zeroth-order contribution obeys
\begin{equation}
  \sum_{p\in\mathcal{P}_q}|\gamma^{(0,p,q)}_{t,\ell}|
  \;\le\;
  C_0'\,\lambda_{\mathrm{env}}^{\ell},
  \qquad
  C_0'=|\mathcal{P}_q|\,C_0 .
\end{equation}
For the LSTM cell-retention path one can take the retained-path rate to
be $\lambda$ itself: the cell-to-cell diagonal block of the full Jacobian
is $F_j=\mathrm{diag}(f_j)$, so
$|f^{(q)}_j| = |[J_j]_{H+q,\,H+q}| \le \|J_j\|_2 \le \lambda$, and since
$0\le e^{(q)}_t\le 1$,
$|\gamma^{(0,q)}_{t,\ell}| = |e^{(q)}_t|\prod_{j=t-\ell+1}^{t}|f^{(q)}_j|
\le \lambda^{\ell}$.

\paragraph{From the zeroth order to the envelope.}
The first-order correction is controlled relative to the retained paths
by the dominance condition of Lemma~\ref{lem:mu_l1_monotone}(ii),
\begin{equation}
  |\gamma^{(1,q)}_{t,\ell}|
  \;\le\;
  c\sum_{p\in\mathcal{P}_q}|\gamma^{(0,p,q)}_{t,\ell}|,
  \qquad c<1,
\end{equation}
so that
\begin{equation}
  |\Gamma^{(q)}_{t,\ell}|
  \;\le\;
  \sum_{p\in\mathcal{P}_q}|\gamma^{(0,p,q)}_{t,\ell}|
  + |\gamma^{(1,q)}_{t,\ell}|
  \;\le\;
  (1+c)\,C_0'\,\lambda_{\mathrm{env}}^{\ell}
  \;=\;
  C_\Gamma\,\lambda_{\mathrm{env}}^{\ell}.
\end{equation}
Finally, the Rayleigh-quotient bound Eq.~\eqref{eq:rayleigh_bound} gives
$0<\Lambda^{(q)}_{r,\ell}\le\overline{\Lambda}_r=\max_i\lambda_{i,r}$,
where $\overline{\Lambda}_r$ is the largest adaptive base rate across
coordinates at the fixed optimizer state~$r$, so the envelope satisfies
\begin{equation}
\label{eq:vanish_envelope_bound}
  f(\ell)
  =
  \sum_{q=1}^{H}\bigl|\Lambda^{(q)}_{r,\ell}\,\Gamma^{(q)}_{t,\ell}\bigr|
  \;\le\;
  H\,\overline{\Lambda}_r\,C_\Gamma\,\lambda_{\mathrm{env}}^{\ell}.
\end{equation}
Optimizer adaptation enters only through $\overline{\Lambda}_r$ and it can rescale the envelope but cannot change its exponential decay rate $\lambda_{\mathrm{env}}$.
Plain SGD is the canonical instance: under isotropic preconditioning $\Lambda_r=\mu I$ the projected base rate reduces to $\Lambda^{(q)}_{r,\ell}=\mu$ (Eq.~\eqref{eq:Lambda_q_SGD_reduces_to_mu}), giving $f(\ell)\le H\mu\,C_\Gamma\,\lambda_{\mathrm{env}}^{\ell}$.

\paragraph{Vanishing gradient as a data-complexity wall.}
When the state-space dynamics is very contractive, the envelope lies in the fast-exponential decay class, $f(\ell)\lesssim\lambda_{\mathrm{env}}^{\ell}$.
Under the bounded-alignment and bounded-noise assumptions of Sec.~\ref{sec:scaling_laws}, the finite-sample requirement~\eqref{eq:min_N_of_ell_for_detection} grows exponentially in the lag, $N(\ell) \;\ge\; C\,\lambda_{\mathrm{env}}^{-\kappa_\alpha\ell}$, with $C>0$ absorbing the lag-independent constants, so the learnability window is capped by a logarithmic ceiling.
Stronger contraction (smaller $\lambda_{\mathrm{env}}$) narrows the
detectable temporal range, and heavier tails (smaller~$\alpha$, larger
$\kappa_\alpha$) compress the ceiling further. This is the precise sense
in which vanishing gradients impose a \emph{data-complexity wall}: even
when the mean gradient is nonzero, the exponential sample cost renders
long-range dependencies practically undetectable.

\paragraph{Necessity of the compatibility condition.}
Condition~\eqref{eq:vanish_compat} cannot be dropped. Consider the
scalar ($H=1$) example in which, for some \(a\in(\lambda,1]\),
\begin{equation}
  \mathcal{T}_j=a,
  \qquad
  \mathcal{R}_j=-(a-\lambda),
  \qquad
  J_j=\mathcal{T}_j+\mathcal{R}_j=\lambda .
\end{equation}
Then $\|J_j\|_2=\lambda<1$, so the exact dynamics are strongly
contractive, but the retained zeroth-order product is $a^\ell$, which
decays more slowly than $\lambda^\ell$ and does not decay at all when
$a=1$.
The full contraction is produced purely by cancellation between the
retained transport $\mathcal{T}_j$ and the residual
$\mathcal{R}_j$, precisely the situation excluded by
Eq.~\eqref{eq:vanish_compat} when
$\lambda_{\mathrm{env}}=\lambda/(1-\eta)<1$.
Indeed, in this example
$\|\mathcal{R}_j\|_2=a-\lambda$, so
Eq.~\eqref{eq:vanish_compat} would require
$a-\lambda\le\eta a$, or equivalently
$\eta\ge 1-\lambda/a$; for \(a\) close to one this forces
\(\eta\) large enough that \(\lambda/(1-\eta)\ge a\), and hence the
retained path is not guaranteed to be contractive.
Intuitively, this corresponds to a pathological retained path whose
gate-like transport is almost fully open (\(a\approx1\)) while an
opposing residual term cancels it in the full Jacobian.
Such cancellation can make the exact Jacobian product vanish while the
retained envelope path remains long-lived, so it is not the
perturbative regime in which the envelope analysis is intended to
operate.

\paragraph{Exploding gradients.}
In the complementary regime of exploding gradients, a closed-form scaling
law is not available within the present framework. Although the signal
amplitude may grow with lag, gradient clipping, optimizer rescaling, and
the compounding of multiplicative noise can inflate $\sigma_\alpha(\ell)$
and drive the effective tail behavior toward heavier tails. The net
effect on detectability is then no longer determined by the envelope
alone, and the learnability window need not expand even when transport
magnitudes grow. Qualitatively, exploding gradients impair learnability
from the opposite direction: vanishing gradients starve the signal,
whereas exploding gradients destabilize it through noise and clipping.

\section{Envelope fitting details and optimizer-dependent regimes}
\label{app:fitting_details_and_optimizer_regimes}

This appendix collects two sets of results that complement the main text.
First, we report the detailed envelope-fit parameters for the simulations with AdamW, including AIC (Akaike Information Criterion) and BIC (Bayesian Information Criterion) model-comparison diagnostics and per-seed variability (Section~\ref{app:fits_adamw}). The numerical values in this subsection support the regime classification reported in Section~\ref{sec:exp_results}.
Then, we report the corresponding results for the simulations performed with plain SGD
under the same experimental conditions (Section~\ref{app:results_sgd}),
which probe how the realized envelope regime and the matched-statistic noise depend on the optimizer. We stress that no hyper-parameter tuning has been performed.

\subsection{Envelope fitting details with AdamW}
\label{app:fits_adamw}

For each architecture, we fit three families to the seed-averaged envelope $\hat f(\ell)$: an exponential $\hat f(\ell)\approx c\,\exp(-\lambda\ell)$, a power-law $\hat f(\ell)\approx c\,\ell^{-\beta}$, and a tempered power-law $\hat f(\ell)\approx c\,\ell^{-\beta}\exp(-(\ell/\ell_c)^k)$.

\paragraph{Use of AIC and BIC.}
The tempered power-law family has more degrees of freedom than either
the exponential or the pure power-law family, and is therefore more
adaptable by construction. It can represent an intermediate power-law-like
range together with an eventual finite-size cutoff, which is precisely the
behavior expected in realistic systems. Consequently,
the fact that AIC and BIC often favor the tempered family should not be
interpreted as the basis for the regime classification.

The regime claim in the main text instead rests on the simpler and more useful contrast between exponential and power-law fits over the statistically accessible lag range.
We report the tempered fits only as descriptive finite-size diagnostics: they quantify how the observed envelopes interpolate between an accessible slow-decay range and an eventual cutoff, but they are not used to decide whether an architecture belongs to the exponential or power-law regime.

\paragraph{Results.}

Table~\ref{tab:fits_adamw_combined} shows the fitted parameters.
ConstGate and SharedGate are firmly in the exponential regime: the simple exponential fit reaches $r^2\approx 1$ with decay rates $\lambda_{\mathrm{const}}\approx 0.050$ and $\lambda_{\mathrm{shared}}\approx 0.071$, while the simple power-law fit drops to $r^2\approx 0.85$.
The tempered family recovers $k=1$ for both, confirming that the residual structure favored by the tempered fit is also exponential.
DiagGate and GRU show the opposite ordering: the simple power-law fit ($r^2\geq 0.98$) clearly dominates the simple exponential fit ($r^2\leq 0.93$), with $\beta_{\mathrm{diag}}\approx 1.51$ and $\beta_{\mathrm{gru}}\approx 0.95$.
In both cases, the tempered family selects $k=0.25$ and a finite $\ell_c$, indicating a strong stretched-exponential cutoff multiplying the power-law trend; the regime we describe in the main text as power-law-with-cutoff.

LSTM requires a careful reading.
The simple exponential fit yields $r^2\approx 0.994$, higher than the simple power-law fit ($r^2\approx 0.89$); a literal application of the $r^2$ comparison would therefore label LSTM exponential.
The fitted rate, however, is $\lambda_{\mathrm{lstm}}\approx 0.004$, more than an order of magnitude smaller than the rates of ConstGate ($0.050$) and SharedGate ($0.071$), so LSTM is not behaving like a short-memory exponential model.
The tempered fit places LSTM at $k\approx 0.70$ and $\ell_c\approx 301$, between the gated regime ($k\approx 0.25$ for DiagGate and GRU) and the pure exponential limit ($k=1$ for ConstGate and SharedGate).
This is consistent with the finite-window ambiguity discussed in the main text. Over the explored horizon, a very slowly decaying exponential and a strongly tempered power-law are nearly indistinguishable, and the tempered family provides the cleanest structural description.

The cross-seed standard deviations in parentheses show that the constrained architectures are highly stable, while DiagGate, GRU, and LSTM exhibit larger variability in the fitted decay parameters.
This variability does not change the qualitative regime assignment. DiagGate and GRU consistently favor the power-law family over the simple exponential fit, whereas LSTM remains the ambiguous intermediate case with a very small exponential rate and a tempered fit between the gated and exponential regimes.

\begin{table}[h]
\centering
\small
\caption{Envelope fits on the seed-averaged envelope $\hat f(\ell)$ under AdamW.
Fit window: $\ell\in[16,756]$ ($n=186$ lag points after trimming).
Parentheses report the cross-seed standard deviation (s.d.) of the corresponding
simple-family parameter obtained by fitting each seed separately.}
\label{tab:fits_adamw_combined}
\begin{tabular}{l rr rr rrr}
\toprule
       & \multicolumn{2}{c}{Exponential}
       & \multicolumn{2}{c}{Power-law}
       & \multicolumn{3}{c}{Tempered power-law} \\
\cmidrule(lr){2-3} \cmidrule(lr){4-5} \cmidrule(lr){6-8}
Model
& $r^2$ & $\lambda$ (s.d.)
& $r^2$ & $\beta$ (s.d.)
& $r^2$ & $k$ & $\ell_c$ \\
\midrule
const
& $1.000$ & $0.050\;(0.000)$
& $0.850$ & $11.97\;(0.02)$
& $1.000$ & $1.00$ & $153.9$ \\

shared
& $1.000$ & $0.071\;(0.018)$
& $0.849$ & $16.91\;(4.04)$
& $1.000$ & $1.00$ & $134.5$ \\

diag
& $0.927$ & $0.006\;(0.001)$
& $0.982$ & $\phantom{0}1.51\;(0.10)$
& $1.000$ & $0.25$ & $153.9$ \\

gru
& $0.901$ & $0.004\;(0.002)$
& $0.992$ & $\phantom{0}0.95\;(0.42)$
& $0.999$ & $0.25$ & $\phantom{0}89.9$ \\

lstm
& $0.994$ & $0.004\;(0.005)$
& $0.890$ & $\phantom{0}1.05\;(1.07)$
& $0.999$ & $0.70$ & $301.4$ \\
\bottomrule
\end{tabular}
\end{table}

\begin{table}[hp!]
\centering
\small
\caption{AIC and BIC for the envelope fits in Table~\ref{tab:fits_adamw_combined}.}
\label{tab:fits_adamw_ic}
\begin{tabular}{l rrr rrr}
\toprule
       & \multicolumn{3}{c}{AIC} & \multicolumn{3}{c}{BIC} \\
       \cmidrule(lr){2-4} \cmidrule(lr){5-7}
Model  & Exp & Power & Tempered & Exp & Power & Tempered \\
\midrule
const  & $-931.8$ & $\phantom{-1}536.5$ & $-1369.4$ & $-925.3$ & $\phantom{-1}542.9$ & $-1353.3$ \\
shared & $-897.4$ & $\phantom{-1}666.6$ & $-1481.1$ & $-890.9$ & $\phantom{-1}673.0$ & $-1465.0$ \\
diag   & $-394.8$ & $-653.6$ & $-1819.6$ & $-388.4$ & $-647.1$ & $-1803.4$ \\
gru    & $-512.9$ & $-975.4$ & $-1740.8$ & $-506.5$ & $-968.9$ & $-1724.7$ \\
lstm   & $-989.6$ & $-434.4$ & $-1433.9$ & $-983.1$ & $-427.9$ & $-1417.8$ \\
\bottomrule
\end{tabular}
\end{table}

\subsection{Results with plain SGD}
\label{app:results_sgd}

The plain-SGD sweep in Table~\ref{tab:fits_sgd} contrasts sharply with the AdamW results in Section.~\ref{app:fits_adamw}.
Once the adaptive preconditioner is removed, DiagGate and GRU no longer exhibit the power-law-with-cutoff envelopes observed under AdamW.
Instead, ConstGate, SharedGate, DiagGate, and GRU collapse onto nearly the same exponential envelope, with $\lambda\approx 0.051$ ($\tau_{\rm env}\approx 19.5$--$19.7$).
LSTM also falls in an exponential regime, but with a much faster decay rate $\lambda\approx 0.690$ ($\tau_{\rm env}\approx 1.45$).

The empirical learnability windows behave accordingly.
At the largest sample budget considered, $N=25600$, the ECF-based window remains zero for ConstGate and DiagGate across all seeds, is nearly zero for LSTM, and opens only modestly for SharedGate and GRU.
The cross-seed mean windows are approximately $0$, $13$, $0$, $126$, and $2$ lags for ConstGate, SharedGate, DiagGate, GRU, and LSTM, respectively.
The MCC estimates give the same qualitative picture, with a few additional small nonzero windows for ConstGate and SharedGate.
Thus, as predicted by the scaling law for exponential envelopes, the learnability window grows very slowly under plain SGD and often cannot leave zero within the explored data budget.

Although ConstGate, SharedGate, DiagGate, and GRU share nearly the same exponential rate $\lambda\approx 0.051$, their windows differ substantially (from $0$ to about $126$ lags). This spread at fixed envelope decay does not reflect the decay class itself, but the architecture-dependent constants entering the detectability bound (i.e. the alignment $|\overline m_\mu(\ell)|$, the noise scale $\sigma_\alpha(\ell)$, and the tail index) consistent with the discussion in the main text.
Tightening these bound constants would refine the predicted window levels without changing the regime classification.

\begin{table}[h]
\centering
\small
\caption{Envelope fits under plain SGD.
Fit window: $\ell\in[16,756]$.
Under SGD, the adaptive base rate is the scalar learning rate $\mu$, so
the envelope reduces to the transport contribution up to a constant factor.}
\label{tab:fits_sgd}
\begin{tabular}{lrrrr}
\toprule
Model & Exp. $r^2$ & $\lambda$ & Power $r^2$ & $\beta$ \\
\midrule
const  & $0.999$ & $0.051$ & $0.848$ & $12.04$ \\
shared & $0.999$ & $0.051$ & $0.847$ & $12.17$ \\
diag   & $0.999$ & $0.051$ & $0.847$ & $12.04$ \\
gru    & $1.000$ & $0.051$ & $0.846$ & $12.09$ \\
lstm   & $1.000$ & $0.690$ & $0.844$ & $163.50$ \\
\bottomrule
\end{tabular}
\end{table}


\clearpage
\section{Code availability}
\label{app:code}

The code used to reproduce the experiments reported in this paper is available at
\begin{center}
\url{https://github.com/lorenzolivi/learnability}
\end{center}
The repository includes scripts and instructions for reproducing the results.

\end{document}